\documentclass[5p]{elsarticle}

\usepackage{amsmath,amssymb}
\usepackage{mathpazo}
\usepackage{enumitem}
\usepackage{times}
\usepackage{graphicx}
\usepackage{subfigure}
\usepackage{epsfig}
\usepackage{epstopdf}
\usepackage{multirow}
\usepackage{esvect}
\usepackage{pifont}
\usepackage{units}
\usepackage{url}
\usepackage{xcolor}
\usepackage{tikz}
\usepackage{arydshln}
\usepackage[flushleft]{threeparttable}
\usepackage{widetext}
\usepackage{amsmath}
\usepackage{amssymb}        
\usepackage{colortbl}
\usepackage{mathtools}
\usepackage{pifont}
\usepackage[linesnumbered,ruled]{algorithm2e}
\usepackage{rotating} 
\usepackage[colorlinks=true]{hyperref}
\usepackage{todonotes}
\usepackage{fancyhdr}
\usepackage{amsmath}
\usepackage{amssymb}
\usepackage{latexsym}
\usepackage{framed}
\usepackage{makecell}
\usepackage{booktabs}
\usepackage{url}
\usepackage{xcolor}
\usepackage{hyperref}
\usepackage{booktabs,array,varwidth}
\usepackage{amsmath,amssymb,amsthm}

\usepackage{cleveref}

\usepackage[pagewise]{lineno}
\modulolinenumbers[250]

\def\checkmark{\tikz\fill[scale=0.3](0,.35) -- (.25,0) -- (1,.7) -- (.25,.15) -- cycle;}

\emergencystretch=\maxdimen
\usetikzlibrary{trees}
\usetikzlibrary{shadings}
\tikzstyle{every node}=[draw=black,thin,anchor=west, minimum height=2.5em]

\biboptions{authoryear}

\begin{document}

\begin{frontmatter}

\title{A comprehensive survey on deep active learning in medical image analysis}%

\author[1,2]{Haoran Wang}
\author[3]{Qiuye Jin}
\author[1,2]{Shiman Li}
\author[1,2]{Siyu Liu}
\author[1,2]{Manning Wang \corref{cor1}}
\author[1,2]{Zhijian Song \corref{cor1}}

\cortext[cor1]{Corresponding authors: Manning Wang and Zhijian Song (Emails: hrwang20@fudan.edu.cn, mnwang@fudan.edu.cn, zjsong@fudan.edu.cn)}

\address[1]{Digital Medical Research Center, School of Basic Medical Sciences, Fudan University, Shanghai 200032, China}
\address[2]{Shanghai Key Laboratory of Medical Image Computing and Computer Assisted Intervention, Shanghai 200032, China}
\address[3]{Computational Bioscience Research Center (CBRC), King Abdullah University of Science and Technology (KAUST), Thuwal 23955, Saudi Arabia}

\begin{abstract}
Deep learning has achieved widespread success in medical image analysis, leading to an increasing demand for large-scale expert-annotated medical image datasets. 
Yet, the high cost of annotating medical images severely hampers the development of deep learning in this field. 
To reduce annotation costs, active learning aims to select the most informative samples for annotation and train high-performance models with as few labeled samples as possible. 
In this survey, we review the core methods of active learning, including the evaluation of informativeness and sampling strategy. 
For the first time, we provide a detailed summary of the integration of active learning with other label-efficient techniques, such as semi-supervised, self-supervised learning, and so on. 
We also summarize active learning works that are specifically tailored to medical image analysis. 
{Additionally, we conduct a thorough comparative analysis of the performance of different AL methods in medical image analysis with experiments.}
In the end, we offer our perspectives on the future trends and challenges of active learning and its applications in medical image analysis.
\end{abstract}

 
\end{frontmatter}

\section{Introduction} 
\label{intro}
Medical imaging visualizes anatomical structures and pathological processes.
It also offers crucial information in lesion detection, diagnosis, treatment planning, and surgical intervention. 
In recent years, the rise of artificial intelligence (AI) has led to significant success in medical image analysis. 
The AI-powered systems for medical image analysis have {approached} the performance of human experts in certain clinical tasks.
Notable examples include skin cancer classification \citep{esteva2017dermatologist}, lung cancer screening with CT \citep{ardila2019end}, polyp detection during colonoscopy \citep{wang2018development}, and prostate cancer detection in whole-slide images \citep{tolkach2020high}.
Therefore, these AI-powered systems can be integrated into existing clinical workflows, which helps to improve diagnostic accuracy for clinical experts \citep{sim2020deep} and support less-experienced clinicians \citep{tschandl2020human}.

Deep learning (DL) models serve as the core of these AI-powered systems for learning complex patterns from raw images and generalizing them to more unseen cases.
Leveraging their robust feature extraction and generalization capabilities, DL models have also achieved remarkable success in the field of medical image analysis \citep{zhou2021review}.
The success of DL often relies on large-scale human-annotated datasets. 
For example, the ImageNet dataset \citep{deng2009imagenet} contains tens of millions of labeled images, and it's widely used in developing DL models for computer vision (CV). 
The size of medical image datasets keeps expanding, but it is still relatively smaller than that of natural image datasets. 
For example, the brain tumor segmentation dataset BraTS consists of multi-sequence 3D MRI scans.
The BraTS dataset expanded from 65 patients in 2013 \citep{menze2014multimodal} to over 1,200 in 2021 \citep{baid2021rsna}.
The latter is equivalent to more than 700,000 annotated 2D images
\footnote{{BraTS 2021 dataset has scans with annotations available for 1,251 patients which all have the same spatial shape. Each case features four MRI sequences, and each sequence contains 155 2D axial slices. Thus, it contains a total of $1,251 \times 4 \times 155 = 775,620$ slices.}}. 
However, the high annotation cost limits the construction of large-scale medical image datasets, mainly reflected in the following two aspects:

\begin{figure*}[!ht]
  \centering
  \includegraphics[width=0.9\textwidth]{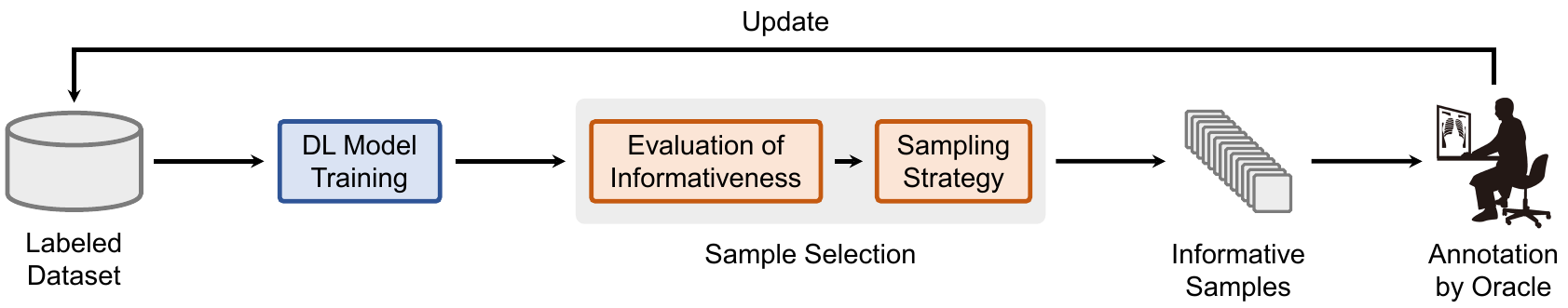}
  {\caption{Illustration of the process of active learning.}}
  \label{fig_al}
\end{figure*}

\textbf{1. Fine-grained annotation of medical images is labor-intensive and time-consuming.}
In clinical practice, automatic segmentation helps clinicians outline different anatomical structures and lesions more accurately.
However, training such a segmentation model requires pixel-wise annotation, which is extremely tedious \citep{rajpurkar2022ai}. 
Another case is in digital pathology.
Pathologists usually require detailed examinations and interpretations of pathological tissue slices under high-magnification microscopes. 
Due to the complex tissue structures, pathologists must continuously adjust the microscope's magnification. 
As a result, it usually takes 15 to 30 minutes to examine a single slide \citep{qu2022towards}.
Making accurate annotations is even more challenging for pathologists.
In conclusion, the annotation process in medical image analysis demands a considerable investment of time and labor.

\textbf{2. The high bar for medical image annotation leads to high costs.}
In CV, tasks like object detection and segmentation also require fine-grained annotations.
However, the widespread use of crowdsourcing platforms has significantly reduced the cost of obtaining high-quality annotations in these tasks \citep{kovashka2016crowdsourcing}. 
However, crowdsourcing platforms have certain limitations in medical image annotation. 
Firstly, annotating medical images demands both medical knowledge and clinical expertise. 
Complex cases even require discussions among multiple senior experts. 
Secondly, even in relatively simple tasks, crowdsourcing workers tend to provide annotations of poorer quality than professional annotators in medical image analysis.
For example, results in \cite{radsch2023labelling} supported the conclusion above in annotating the segmentation mask of surgical instruments.  
Crowdsourcing platforms could also raise privacy concerns \citep{rajpurkar2022ai}. 
{
Nevertheless, we will face new challenges when the annotators change from crowdsourcing workers to clinical experts. 
First of all, recruiting doctors for annotation is very expensive. 
For instance, a radiologist usually takes about 60 minutes to manually segment brain tumors per patient in its multi-sequence MRI volumes\citep{menze2014multimodal}.
And the median hourly rate of a radiologist is \$219 in the US\footnote{{https://www.salary.com/tools/salary-calculator/radiologist-hourly}}. 
Besides, to minimize individual bias for certain scenarios, it is common to have a doctor annotate the same case multiple times or have multiple doctors annotate it.
Multiple annotation rounds and annotators introduce intra- and inter-annotator variability and handling such variabilities leads to additional annotation costs \citep{karimi2020deep}.}
In summary, high-quality annotations often require the involvement of experienced doctors, which inherently increases the annotation cost of medical images.

The high annotation cost is one of the major bottlenecks of DL in medical image analysis. 
Active learning (AL) is considered one of the most effective solutions for reducing annotation costs. 
The main idea of AL is to select the most informative samples for annotation and then train a model with these samples in a supervised way.
In the general practice of AL, annotating a part of the dataset could reach comparable performance of annotating all samples. 
As a result, AL saves the annotation costs by querying as few informative samples for annotation as possible.
{The process of AL is illustrated in Fig.\ref{fig_al}, which we will detail in §\ref{settings}.}
Specifically, we refer to the AL works focusing on training a deep model as deep active learning.

Reviewing AL works in medical image analysis is essential for reducing annotation costs. 
\cite{budd2021survey} investigated the role of humans in developing and deploying DL in medical image analysis, where AL is considered a vital part of this process. 
In \cite{tajbakhsh2020embracing}, AL was one of the solutions for training high-performance medical image segmentation models with imperfect annotation.
As one of the methods in label-efficient deep learning for medical image analysis, \cite{jin2023label} summarized AL methods from model and data uncertainty.
There are also several surveys on AL in machine learning or CV. 
\cite{settles2009active} provided a general introduction and comprehensive review of AL works in the machine learning era. 
After the advent of DL, \cite{ren2021survey} reviewed the development of deep active learning and its applications in CV and natural language processing. 
\cite{liu2022survey} summarized the model-driven and data-driven sample selectors in deep active learning.
\cite{zhan2022comparative} reimplemented high-impact works in deep active learning with fair comparisons. 
\cite{takezoe2023deep} reviewed recent developments of deep active learning in CV and its industrial applications. 

However, the surveys mentioned above have certain limitations. 
Firstly, new ideas and methods are constantly emerging with the rapid development of deep active learning. 
Thus, a more comprehensive survey of AL is needed to cover the latest advancements. 
Secondly, a recent trend is combining AL with other label-efficient techniques, which is also highlighted as a future direction by related surveys \citep{takezoe2023deep, budd2021survey}. 
However, existing surveys still lack summaries and discussions on this topic.
{
Thirdly, limited surveys have evaluated the performance of different AL methods on the medical imaging dataset, indicating a near absence of such efforts. 
}
Finally, the high annotation cost emphasizes the increased significance of AL in medical image analysis, yet related reviews still lack comprehensiveness in this regard.

This survey comprehensively reviews AL for medical image analysis, including core methods, integration with other label-efficient techniques, and AL works tailored to medical image analysis.
We first searched relevant papers on Google Scholar and arXiv using the keyword ``Active Learning" and expanded the search scope through citations. 
{
The included papers in this survey mainly belong to the field of medical image analysis. 
It should be noted that some important works of AL in the general CV field are also included since the development of AL in medical image analysis is influenced by the advance of AL in CV. 
Ignoring these works would flaw the logic and taxonomy of this survey. 
To balance the AL works of different fields, we first present the seminal works in each subsection, which may include works in the general CV field, and then provide a detailed review of the AL papers related to medical image analysis within this category.
}
Additionally, most works in this survey are published in top-tier journals (e.g., TPAMI, TMI, MedIA, TBME, JBHI, etc.) and conferences (e.g., CVPR, ICCV, ECCV, ICML, ICLR, NeurIPS, MICCAI, ISBI, MIDL, etc.). 
As a result, this survey involves nearly 164 relevant AL works with 234 references. 
The contributions of this paper are summarized as follows: 
\begin{itemize}
    \item Through an exhaustive literature search, we provide a comprehensive survey and a novel taxonomy for AL works, especially those focusing on medical image analysis.
    
    \item While previous surveys mainly focus on evaluating informativeness, we further summarize different sampling strategies in deep active learning, such as diversity and class-balance strategies, aiming to provide references for future method improvement.
    
    \item In line with current trends, this survey is the first to provide a detailed review of the integration of AL with other label-efficient techniques, including semi-supervised learning, self-supervised learning, domain adaptation, region-based active learning, and generative models.
    
    {\item For the sake of promoting research and contributing to the community, this survey evaluated the performance of several popular AL methods on multiple medical imaging datasets. The codes are also made public for better reproducibility.}
\end{itemize}

\begin{figure*}[!ht]
  \centering
  \includegraphics[width=\textwidth]{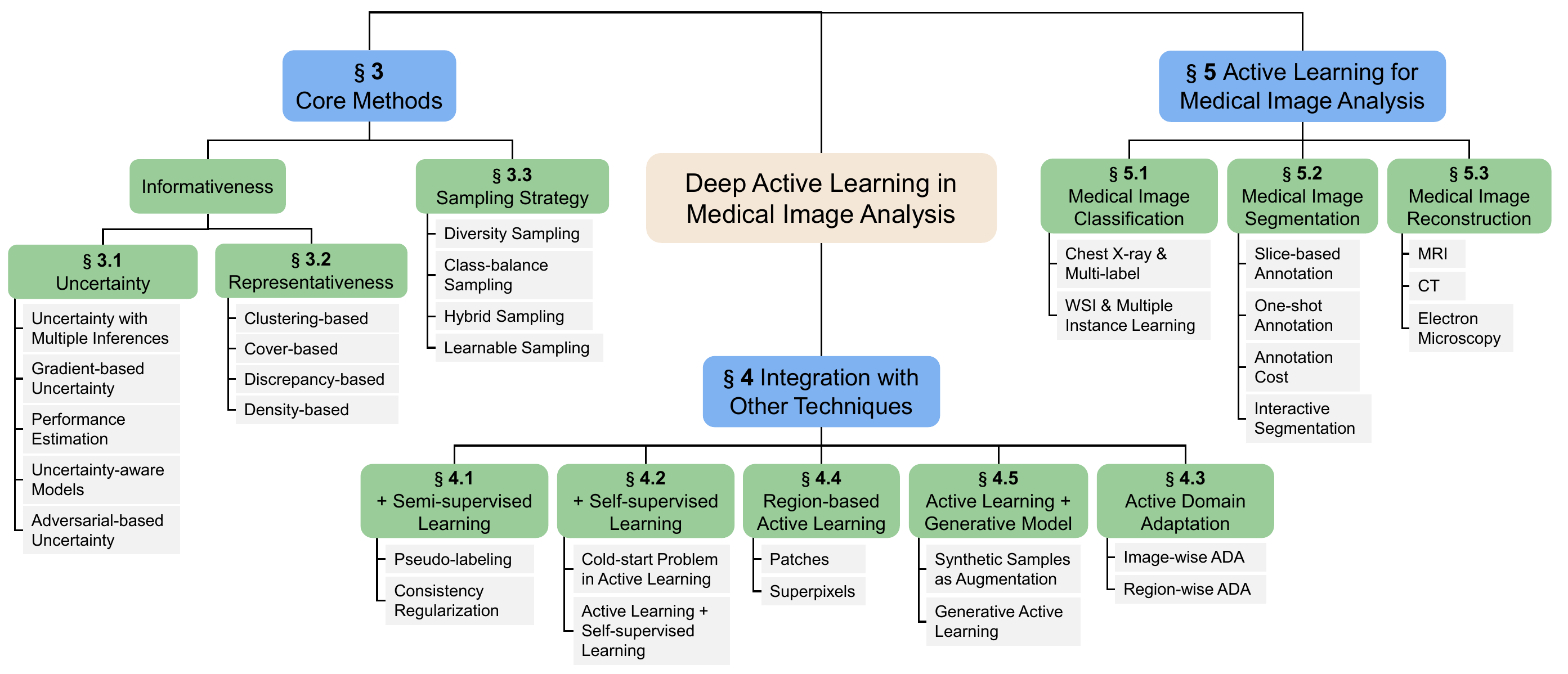}
  {\caption{Overall framework of this survey.}}
  \label{fig_overview}
\end{figure*}

The rest of this survey is organized as follows: §\ref{settings} introduces problem settings and mathematical formulation of AL, §\ref{core_methods} discusses the core methods of AL, including evaluation of informativeness (§\ref{uncertainty} \& §\ref{representativeness}) and sampling strategies (§\ref{sampling_strategy}), §\ref{integration} reviews the integration of AL with other label-efficient techniques, §\ref{al4mia} summarizes AL works tailored to medical image analysis. {The experimental settings, results, and analysis are in §\ref{exps}}. We discuss existing challenges and future directions of AL in §\ref{challenge} and conclude the whole paper in §\ref{conclusion}. The overall framework of this survey is shown in Fig. \ref{fig_overview}. 

Due to the rapid development of AL, many related works are not covered in this survey. 
We refer readers to our constantly updated website\footnote{https://github.com/LightersWang/Awesome-Active-Learning-for-Medical-Image-Analysis} for the latest progress of AL and its application in medical image analysis.

\section{Problem Settings and Formulations of Active Learning} 
\label{settings}
AL generally involves three problem settings: membership query synthesis, stream-based selective sampling, and pool-based active learning \citep{settles2009active}. 
In the case of membership query synthesis, we can continuously query any samples in the input space for annotation, including synthetic samples produced by generative models \citep{angluin1988queries, angluin2004queries}. 
We also refer to this setting as generative active learning in this survey.
Membership query synthesis is typically suitable for low-dimensional input spaces.
However, when expanded to high-dimensional spaces (e.g., images), the queried samples produced by generative models could be unidentifiable for human labelers.
The recent advances of deep generative models have shown great promise in synthesizing realistic medical images, and we further discuss its combination with AL in §\ref{generative}. 
Stream-based selective sampling assumes that samples arrive one by one in a continuous stream, and we need to decide whether or not to request annotation for incoming samples \citep{cohn1994improving}. 
This setting is suitable for scenarios with limited memory, such as edge computing, but it neglects sample correlations.

Most AL works follow pool-based active learning, which draw samples from a large pool of unlabeled data and requests oracle (e.g., doctors) for annotations. 
Moreover, if multiple samples are selected for labeling at once, we can further call this setting ``batch-mode".
Deep active learning is in batch-mode by default since retraining the model every time a sample is labeled is impractical.
Also, one labeled sample does not necessarily result in significant performance improvement.
Therefore, unless otherwise specified, all works in this survey follow the setting of batch-mode pool-based active learning.

The flowchart of active learning is illustrated in Fig. \ref{fig_al}. 
Assuming a total of $T$ annotation rounds, active learning primarily consists of the following steps:

\textbf{(1) Sample Selection:} 
In the $t$-th round of annotation, $1 \le t \le T$, an {informativeness function $I$} is used to evaluate the informativeness of each sample in the unlabeled pool $D_t^u$. 
Then, a batch of samples is selected with a certain sampling strategy $S$.
{
In medical image analysis, active learning selects a batch of images most of the time (i.e., image-wise selection).
In this survey, the sampling unit of AL selection is an image (could be 2D or 3D) unless specifically stated.
However, with the development of AL, region-wise (§\ref{region}, §\ref{regionADA}) or slice-wise annotations (§\ref{slice}) are adopted in AL. 
Please refer to these sections for more details.
}

Specifically, the queried dataset of $t$-th round $D_t^q$ is constructed as follow:

\begin{equation}
D_t^q = \underset{D_t^q \subset D_t^u}{S} \left(\underset{x \in D_t^u}{I} \left(x, f_{\theta_{t-1}}\right), b\right)
\end{equation}

\noindent where $x$ represents sample in the dataset, $D_t^u$ and $D_t^q$ are unlabeled and queried dataset in round $t$, respectively. 
$f_{\theta_{t-1}}$ and $\theta_{t-1}$ represent the deep model and its parameters from the previous round, respectively. 
The annotation budget $b$ is the number of queried samples for each round, far less than the total count of unlabeled samples, i.e., $b = |D_t^q| \ll |D_t^u|$.

\textbf{(2) Annotation by Oracle:} 
After sample selection, the queried set $D_t^q$ is sent to oracle (e.g., doctors) for annotation, and newly labeled samples are added into the labeled dataset $D_t^l$. 
The update of $D_t^l$ is as follow:

\begin{equation}
   D_t^l = D_{t-1}^l \cup \{(x, y) | x \in D_t^q\}
\end{equation}

\noindent where $y$ represents the label of $x$, and $D_t^l$ and $D_{t-1}^l$ denote the labeled sets for round $t$ and the previous round, respectively. Besides, the queried samples should be removed from the unlabeled set $D_t^u$:

\begin{equation}
   D_t^u = D_{t-1}^u \backslash \{x | x \in D_t^q\}
\end{equation}

{
It is worth noting that some current works combine active learning with interactive segmentation. 
In interactive segmentation, the model assists experts in annotation, thereby reducing the difficulty of the annotation process.
For more details, please refer to section §\ref{interactive}.
}

\textbf{(3) DL Model Training:} 
After oracle annotation, we train the deep model using the labeled set of this round $D_t^l$ in a fully supervised manner. 
The deep model $f_{\theta_t}$ is trained on $D_t^l$ to obtain the optimal parameters $\theta_t$ for round $t$. 
The mathematical formulation is as follows:

\begin{small}
\begin{equation}
   \theta_t = \underset{\theta}{\arg\min} \underset{(x, y) \in D_t^l}{\mathbb{E}} \left[\mathcal{L}(f_{\theta}(x), y)\right] = \underset{\theta}{\arg\min} \underset{(x, y) \in D_t^l}{\mathbb{E}} \left[\mathcal{L}(x, y; \theta)\right] 
\end{equation}
\end{small}

\noindent where $\mathcal{L}(f_{\theta}(x), y)$ represents the loss function and it could be rewritten as $\mathcal{L}(x, y; \theta)$ for simplicity.

\textbf{(4) Repeat steps 1 to 3 until the annotation budget limit or the expected performance is reached.} 
{Recently, some works adopted the one-shot fashion in active learning, which performed sample selection without multiple rounds.
Please refer to §\ref{oneshot}.
}

It is worth noting that the model needs proper initialization to start the AL process. 
If the initial model $f_{\theta_0}$ is randomly initialized, it could only produce meaningless informativeness. 
To address this issue, most AL works randomly choose a set of samples as initially labeled dataset $D_0^l$ and train $f_{\theta_0}$ upon $D_0^l$. 
For more details on better initialization of AL using pre-trained models, please refer to §\ref{self}.

\section{Core Methods of Active Learning} 
\label{core_methods}
In this survey, we consider the evaluation of informativeness and sampling strategy as the core methods of AL.
Informativeness represents the value of annotating each sample. 
Higher informativeness often indicates a higher priority to request these samples for labeling.
Typical metrics of informativeness include uncertainty and representativeness.
{Based on the informativeness scores, a certain sampling strategy is used to select a small number of unlabeled samples for annotation. 
Most AL works simply ranked these samples by their informativeness metrics and selected the highest ones according to the annotation budget (i.e., top-k selection). 
However, current informativeness scores are more or less flawed and they may cause issues like redundancy or class imbalance among the queried samples. 
Therefore, we need more advanced sampling strategies to mitigate these issues arising from imperfect informativeness metrics.}

{
In this section, we reviewed two major informativeness metrics, including uncertainty (§\ref{uncertainty}) and representativeness (§\ref{representativeness}), and sampling strategy (§\ref{sampling_strategy}).
As a unique contribution of this survey, we, for the first time, explicitly define sampling strategies as core methods of AL and review how to design a better sampling strategy in AL. 
}
Additionally, we provide a summarization of all the cited AL works in this survey. 
Methods and basic metrics of uncertainty or representativeness and sampling strategies are detailed in Table \ref{tab2}.

\subsection{Evaluation of Informativeness: Uncertainty}
\label{uncertainty}

\begin{table}
    \centering
    \caption{Formulations of uncertainty metrics based on prediction probability in active learning. In the equations column, $x$ stands for sample, $f$ is the deep model, while $C$ is the number of classes. In the direction column, $\uparrow$ means higher values indicating higher uncertainty, while $\downarrow$ means lower values indicating higher uncertainty.}
    \scalebox{0.75}{
    \begin{tabular}{ccc}
        \toprule
        Names & Equations & Direction  \\
        \midrule
        \makecell*[c]{Prediction Probability} & \makecell*[c]{$\mathbf{p}=\text{Softmax}\left(f_{\theta}\left(x\right)\right)\in\mathbb{R}^C,$\\ $\mathbf{p}=\left[p_1,p_2,\cdots,p_C\right]$} & - \\ 
        \midrule
        \makecell*[c]{Least Confidence\\ \citep{lewis1994heterogeneous}} & $\underset{i}{\max} {p_i}$ & $\downarrow$ \\ 
        \makecell*[c]{Entropy \\ \citep{joshi2009multiclass}} & $-\sum_{i=1}^{C}{p_i \log{p_i}}$ & $\uparrow$ \\ 
        \makecell*[c]{Margin \\ \citep{roth2006marginbased}} & $\ \underset{i}{\max}{p_i}-\underset{j,j\neq k}{\max}{p_j},k=\underset{i}{\arg\max} {p_i}$ & $\downarrow$ \\ 
        \makecell*[c]{Mean Variance \\ \citep{gal2017deep}} & $-\frac{1}{C}\sum_{i=1}^{C}\left(p_i-\bar{p}\right)^2,\bar{p}=-\frac{1}{C}\sum_{i=1}^{C}p_i$ & $\uparrow$ \\
        \bottomrule
    \end{tabular}}
    \label{tab_metrics}
\end{table}

{Despite great progress has been made in medical image analysis, safety and interpretability are still unsolved problems for deploying DL models in real-world clinical practice. 
Due to the high variability of medical images and the limited training data, the predictions of DL models are not reliable and trusted.
Correctly assessing and quantifying uncertainty in medical image analysis would allow models to alert ambiguities, artifacts, and unseen patterns in the data \citep{ghesu2021quantifying, linmans2023predictive}.
Such properties of uncertainty are helpful in AL since novel patterns in the unlabeled samples can be identified by the uncertainty. 
Therefore, uncertainty is frequently used in active learning as an informative metric.
In the AL query, samples with higher uncertainty are considered hard and more likely to be misclassified by the current model. 
Annotating and training on these samples helps the model learn new patterns and improve performance.} 


{Tracing back to the cause of uncertain predictions, the uncertainty can be mainly separated into two types: aleatoric uncertainty (AU) and epistemic uncertainty (EU) \citep{kendall2017uncertainties}.
AU (i.e., data uncertainty) captures the noisy observations in data, such as motion artifacts of MRI or metal artifacts of CT in medical image analysis. 
AU cannot be reduced by acquiring more data.
A high EU (i.e., model uncertainty) indicates that the samples contain knowledge that has not yet been mastered by the model.
Therefore, the EU can be reduced by involving more data.
However, most of the AL works did not consider the separation of AU and EU. 
So, the terminology of uncertainty in AL mainly refers to prediction uncertainty, which is the composition of AU and EU. 
This is because explicitly separating AU and EU is usually very difficult and may not bring much benefit in the practice of AL \citep{kahl2024values}.
In this survey, unless explicitly stated otherwise, all uncertainties refer to predictive uncertainties, meaning that no separation between AU and EU has been performed.
}

The most straightforward uncertainty metrics in deep AL are based on prediction probabilities with a single forward pass. 
These metrics have been widely used in AL since the machine learning era, and their formulations are detailed in Table \ref{tab_metrics}. 
However, directly transferring them to deep AL would be challenging due to the notorious issue of over-confidence in deep neural networks \citep{mehrtash2020confidence, guo2017calibration}.
Over-confidence refers to the model having excessively high confidence in its predictions, even though they might not be accurate.
It could result in high confidence (e.g., 0.99) of the wrong class for misclassified samples. 
For uncertain samples, it leads to extreme confidence (e.g., 0.99 or 0.01) instead of normal one (e.g., 0.6 or 0.4) as it should.
As a result, over-confidence distorts uncertainty estimation since it affects the predicted probabilities for all classes.

\begin{figure}[!ht]
  \centering
  \includegraphics[width=\columnwidth]{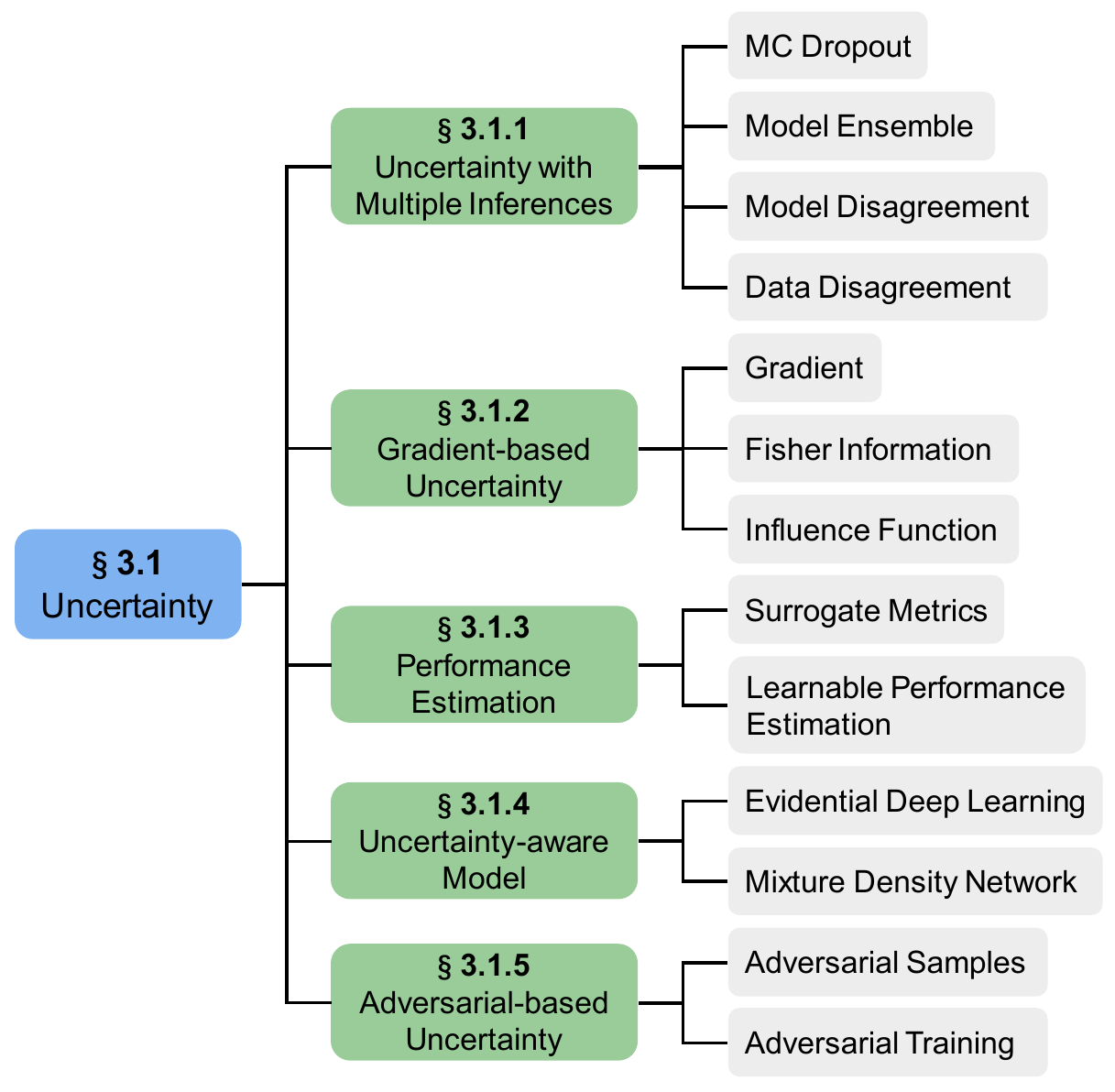}
  \caption{The taxonomy of uncertainty-based active learning.}
  \label{fig_uncer}
\end{figure}

This section divides the uncertainty-based AL into multiple inference, gradient-based uncertainty,  performance estimation, uncertainty-aware models, and adversarial-based uncertainty. 
The taxonomy of uncertainty-based AL is shown in Fig. \ref{fig_uncer}.

\subsubsection{Uncertainty with Multiple Inferences}
\label{mul_infer}

{
To mitigate over-confidence, a common strategy for uncertainty-based AL is to run the model multiple times under perturbations.
The main idea is to reduce the bias introduced by network architectures or training data.
These biases often contribute to the over-confidence issue. 
Two approaches are often used to utilize multiple inference results for AL.
The first is to calculate the classic uncertainty metrics with the average probability of multiple inferences.
Averaging the prediction probabilities of multiple inferences helps to reduce individual bias that causes over-confidence. 
The other approach takes the disagreement between different prediction results as uncertainty quantification.  
Samples with higher disagreement indicate higher uncertainty and are suitable for annotation in AL.
}

{
In this section, we will introduce four types of methods for AL with multiple inferences: Monte Carlo dropout (MC dropout), model ensemble, model disagreement, and data disagreement.
The first two used the average probability of results of multiple inferences to calculate the uncertainty metrics, like entropy and margin.
The last two are based on disagreement.
For the source of perturbation, the first three perturb the model parameters while the last perturb the input data.
}

\noindent \textbf{MC dropout} randomly discards certain neurons in the deep model during each inference \citep{gal2016dropout}.
With MC dropout enabled, the model runs multiple times to get different predictions. 
\cite{gal2017deep} was the pioneering work of deep AL.
They were the first to use MC dropout in computing uncertainty metrics like entropy, standard deviation, and Bayesian active learning disagreement (BALD) \citep{houlsby2011bayesian}. 
Results showed that MC Dropout could significantly improve the performance of uncertainty-based deep AL.
{Besides, they were also among the first to apply deep AL in medical image analysis. 
In the skin lesion analysis dataset ISIC 2016, they found that BALD consistently outperformed the random baseline. 
In brain cell type classification, \cite{yuan2020few} calculated entropy using the average probability of multiple MC dropout runs.
\cite{gu2018reliable} adopted the variance of multiple MC dropout runs as the uncertainty metric in the classification of confocal endomicroscopy and gastrointestinal endoscopy.
}

\noindent \textbf{Model ensemble} trains multiple models to get numerous predictions during inference.
\cite{beluch2018power} conducted a detailed comparison of models ensemble and MC dropout in uncertainty-based AL.
{
Results in the standard datasets demonstrated that the model ensemble performs better. 
For AL in diagnosing diabetic retinopathy, the proposed method achieved significant improvement compared to the random baseline. 
}
However, model ensemble requires significant training overhead in DL. 
To reduce the computational costs, snapshot ensemble \citep{huang2017snapshot} obtained multiple models in a single run with cyclic learning rate decay.
An early attempt in \cite{beluch2018power} showed that snapshot ensemble leads to worse performance than model ensemble.
\cite{jung2023simple} improved the snapshot ensemble by maintaining the same optimization trajectory in different AL rounds, along with parameter regularization.
Results showed that the improved snapshot ensemble outperforms the model ensemble. 
{\cite{nath2021diminishing} employed stein variational gradient descent to train an ensemble of models, aiming to ensure diversity.
Their proposed method showed advantages to other competitors in segmenting the pancreas and tumor on CT and hippocampus on MRI.
}



\noindent \textbf{Model disagreement:}
We can utilize the disagreement between the outputs of different models, which can be also referred to as Query-by-Committee (QBC) \citep{seung1992query}. 
{This type of method was widely used in AL for medical image analysis.}
Suggestive annotation (SA) is the pioneering work of AL for medical image analysis \citep{yang2017suggestive}.
They trained multiple segmentation networks with bootstrapping.
Variance among these models is used as the disagreement metric.
{
SA demonstrated superior performance in segmenting glands on pathological images and lymph nodes on ultrasound images.}
{
In abdominal multi-organ segmentation, \cite{qu2023annotating} trained three different segmentation models and adopted the variance between their predictions.}
{
In carotid intima-media segmentation for ultrasound images, \cite{tang2023pld} selected samples with the highest Kullback-Leibler (KL) divergence between the predictions of teacher and student models for annotation.
}
In polyp segmentation of capsule colonoscopy, \cite{bai2022discrepancybased} trained multiple decoders using class activation maps (CAMs) \citep{zhou2016learning} generated by a classification network.
They further proposed model disagreement and CAM disagreement for sample selection. 
Model disagreement included entropy of prediction probabilities and Dice between outputs of different decoders, while CAM disagreement measured the Dice between CAMs and outputs of all decoders. 
This method selected samples with high model disagreement and CAM disagreement for annotation.
However, samples with low model disagreement but high CAM disagreement were treated as pseudo-labels for semi-supervised training. 
In rib fracture detection, \cite{huang2020rectifying} adopted Hausdorff distance to measure the disagreements between different CAMs.
Besides, \cite{mackowiak2018cereals} adopted vote entropy between different MC dropout inferences as the disagreement metric.

\noindent \textbf{Data disagreement:}
Since training multiple models can be computationally expensive, measuring the disagreements between different perturbations of input data is also helpful in AL.
Kullback-Leibler (KL) divergence is a commonly used metric for quantifying disagreement.
{In COVID diagnosis, \cite{wu2021covidal} computed KL divergence between different versions of augmentations as the disagreement measure to select informative CT scans for annotation. }
\cite{siddiqui2020viewal} calculated the KL divergence between predictions of different viewpoints in 3D scenes to select informative regions for AL. 
Additionally, recent works have adopted alternative metrics to calculate disagreement. 
\cite{lyu2023boxlevel} proposed input-end committee, which randomly augmented the input data to get multiple predictions. 
They further measured the classification and localization disagreements between different predictions with cross-entropy and variance, respectively. 
\cite{parvaneh2022active} interpolated the unlabeled samples and labeled prototypes in the feature space.
If the interpolated sample's prediction disagrees with the corresponding prototype's label, it indicates that the unlabeled samples introduce new features.
Thus, these unlabeled samples should be sent for annotation.
Results showed advancements across various datasets and settings.

\subsubsection{Gradient-based Uncertainty}
\label{gradient}

Gradient-based optimization is the cornerstone of DL-based medical image analysis. 
The gradient of each sample reflects its contribution to the change of model parameters. 
A larger gradient length indicates a tremendous change of parameters by the sample, thus implying high uncertainty. 
Furthermore, gradients are independent of predictive probabilities, which makes them less susceptible to over-confidence. 
Three metrics that are frequently used as gradient-based uncertainty: gradients, Fisher information, and influence functions. 
{
It should be noted that the gradient computation in this section did not use the ground truth labels which are unavailable for unlabeled samples. 
Instead, the corresponding methods either used supervised loss with pseudo-labels (e.g. cross-entropy loss with pseudo-labels) or unsupervised loss (e.g. entropy loss), thereby making the gradient computation independent of the true labels.
}

\noindent \textbf{Gradient:}
A larger gradient norm (i.e., gradient length) denotes a greater influence on model parameters, indicating higher uncertainty in AL. 
{
As an early attempt, \cite{otalora2017training} adopted the classic expected gradient length \citep{settles2007multiple} to select valuable samples for annotation in exudate classification of eye fundus images.
}
As a popular and pioneering work in the DL era, \cite{ash2020deep} proposed batch active learning by diverse gradient embeddings (BADGE).
They calculated the gradients only for the parameters of the network's final layer, with the most confident classes as pseudo labels in gradient computation.
Then, k-Means++ is performed on gradient embeddings for sample selection.
Results showed competitive performances of BADGE across diverse datasets, network architectures, and hyperparameter settings. 
{Gradient has been widely used in active learning of medical image analysis.
\cite{aklilu2022alges} extended the BADGE framework into semantic segmentation of laparoscopic surgical images.
}
\cite{wang2022boosting} proved mathematically that a larger gradient norm corresponds to a lower upper bound of test loss.
Thus, they employed expected empirical loss and entropy loss for gradient computation, which both obviate the necessity for labels.
{
The former is the weighted sum of the losses of each class and the class probabilities, which are as follows:
\begin{equation}
    \mathcal{L}_{exp}(x) = \sum\limits_{i=1}^{C} \left[ p_{i} \cdot \mathcal{L}(x, y_i; \theta) \right] 
    \label{expected_loss}
\end{equation}
\noindent where $y_i$ is the label of class $i$. 
The entropy loss is solely based on the probabilities of all classes, which are as follows:
\begin{equation}
    \mathcal{L}_{ent}(x) = -\sum\limits_{i=1}^{C} p_{i} \log p_{i} 
    \label{entropy_loss}
\end{equation}
The proposed method outperformed other comparative methods in cryo-electron tomography (cyro-ET) subtomogram classification.
}
\noindent Besides, \cite{dai2020suggestive} proposed a new gradients-based active learning method in MRI brain tumor segmentation. 
They first trained a variational autoencoder (VAE) \citep{kingma2013auto} to learn the data manifold.
Then, they trained a segmentation model and calculated gradients of Dice loss using available labeled data.
The sample selection was guided by the gradient projected onto the data manifold.
Their extended work \citep{dai2022suggestive} further demonstrated superior performance in MRI whole brain segmentation.

\noindent {\textbf{Fisher information} is effective in AL of machine learning models \citep{chaudhuri2015convergence, sourati2017asymptotic}. 
Fisher information (FI) reflects the overall uncertainty of model parameters according to data distribution. 
FI is defined as the expectation of the squared gradients with respect to the model parameter, the formulation is as follows: 
\begin{equation}
    \mathcal{I}_{Fisher}(x; \theta) = \mathbb{E}_{y} \left[ \nabla_{\theta}^{2} \mathcal{L}(x, y; \theta)  \right] 
    \label{fisher}
\end{equation}
\noindent where $\mathcal{I}$ is the notation of Fisher information. The trace of the inverse of FI often serves as the objective for AL:
\begin{small} 
\begin{equation}
    \underset{D^q \subset D^u}{\arg \min} \operatorname{Tr}\left[\left( \sum_{x \in D^q} \mathcal{I}_{Fisher}(x ; \theta)\right)^{-1} \left( \sum_{x \in D^u} \mathcal{I}_{Fisher}(x ; \theta) \right) \right]
    \label{fisher_argmin}
\end{equation}
\end{small} 
By solving Eq. \ref{fisher_argmin}, the selected samples could help the model converge faster toward optimal parameters. 
}
However, the computation cost of FI-based methods grows quadratically with the increase of model parameters, which is unacceptable for deep active learning. 
{\cite{sourati2018active} and their extended work \citep{sourati2019intelligent} were the first to incorporate FI into deep active learning of medical image analysis.}
They used the average gradients of each layer to calculate the FI matrix, thus reducing the computation cost.
{Due to the absence of ground truth labels, they adopt the expected empirical loss (i.e., Eq. \ref{expected_loss}) for gradient computation.}
This method outperformed competitors in brain extraction across different age groups and pathological conditions. 
Additionally, \cite{ash2021gone} only computed the FI matrix for the network's last layer.
{The gradient computation of this work is the same as that of BADGE \citep{ash2020deep}.}

\noindent \textbf{Influence function:}
\cite{liu2021influence} employed influence function \citep{koh2017understanding} to select samples that bring the most positive impact on model performance. 
{
The influence function of an unlabeled sample is defined as follows \citep{weisberg1982residuals}:
\begin{small} 
\begin{equation}
    \mathcal{I}_{Influence}(x; D^{l}) = 
    - \left( \sum_{(x, y) \in D^{l}} \nabla_{\theta}  \mathcal{L}(x, y; \theta) \right) H_{\theta}^{-1} \nabla_{\theta} \mathcal{L}(x, y; \theta)
    \label{influence}
\end{equation}
\end{small} 
\noindent where $H_{\theta}^{-1}$ is the Hessian matrix of the labeled set, $H_{\theta} = \sum_{(x, y) \in D^{l}} \nabla_{\theta}^{2}  \mathcal{L}(x, y; \theta)$. 
In Eq. \ref{influence}, the gradients of the first (i.e., the sum of gradients of labeled samples) and the second term (i.e., the Hessian matrix) can be derived with the ground truth label.
For the third term, they replaced the true gradients with the gradients of the expected empirical loss due to the unavailability of the ground truth label.
}

\subsubsection{Performance Estimation}
\label{per_est}

In this section, uncertainty metrics are estimations of the current task's performance. 
There are two types of such metrics: test loss or task-specific evaluation metrics. 
These metrics reflect the level of prediction error.
{For instance, a low Dice score in tumor segmentation for a patient suggests the model failed to produce accurate segmentation.}
Request annotations for these samples would be beneficial for improving the model's performance.
However, due to the unavailability of ground truth labels, we can only estimate these metrics instead of calculating them precisely.
There are primarily two methods for estimating performance: surrogate metrics and learnable performance estimation.

\noindent \textbf{Surrogate metrics} are widely used in active learning of medical image analysis.
For example, these metrics could be upper or lower bounds for loss or task-specific evaluation metrics.
{In breast cancer segmentation on immunohistochemistry images,} \cite{shen2020deep} calculated the intersection over union (IoU) of all predictions by MC dropout.
They found a strong linear correlation between this IoU and the real Dice coefficient. 
In skin lesion and X-ray hand bone segmentation, \cite{zhao2021dsal} calculated the average Dice coefficient between the predictions of the intermediate layers and final layer through deep supervision.
They found a linear correlation between this average Dice and the real Dice coefficient.
Besides, \cite{huang2021semisupervised} found that within limited training iterations, the loss of a sample is bounded by the norm of the difference between the initial and final network outputs. 
Inspired by this, they proposed cyclic output discrepancy (COD) as the difference in model output between two consecutive annotation rounds. 
Results indicated that a higher COD is associated with higher loss.
Therefore, they opted for samples with high COD.
They also demonstrate a linear correlation with the evaluation metrics with post-hoc validation.

\noindent \textbf{Learnable performance estimation:} 
We can train auxiliary neural network modules to predict the performance metrics. 
As one of the most representative works in this line of research, learning loss for active learning (LLAL) \citep{yoo2019learning} trained an additional module to predict the loss value of a sample without its label. 
Since loss indicates the quality of network predictions, the predicted loss is a natural uncertainty metric for sample selection.
Results showed that predicted and actual losses are strongly correlated.
The proposed method also outperformed several AL baselines.
In lung nodule detection with CT scans, \cite{liu2020deep} built upon LLAL to predict the loss of each sample and bounding box. 
In COVID diagnosis, \cite{wu2021covidal} adopted both the predicted loss and the disagreements between different predictions for sample selection. 
{
\cite{wu2022federated} further combined the loss prediction and sample diversity in the federated active learning of COVID diagnosis and colonoscopy polyp analysis. 
}
Since AL focuses only on uncertainty ranking of the unlabeled samples, \cite{kim2021taskaware} relaxed the loss regression to loss ranking prediction.
Thus, they replaced the loss regressor in LLAL with the ranker in RankCGAN \citep{saquil2018ranking}. 
Results showed that loss ranking prediction outperforms the actual loss regression in LLAL.
{
\cite{zhou2021qualityaware} and their subsequent work \citep{zhou2022volumetric} introduced a quality assessment module to provide a predicted average IoU score for each slice.
They interactively selected slices with the lowest scores in each volume for annotation.
}

\subsubsection{Uncertainty-aware Model}
\label{uncer_model}
{
In the above sections, uncertainty is derived based on the commonly used deterministic model in DL. 
However, some models can inherently capture uncertainty, such as VAE or probabilistic U-Net for medical image analysis \citep{kohl2018probabilistic}.
In this way, they no longer output a point estimate but instead a distribution of possible predictions, thus mitigating over-confidence. 
We refer to them as uncertainty-aware models in this survey.
They only require a single forward pass of the deep model, thus significantly reducing computational and time costs during inference. 
Evidential deep learning (EDL) and mixture density networks (MDN) are often used for uncertainty-aware models in AL.
}

\noindent \textbf{Evidential deep learning} replaces the Softmax distribution with a Dirichlet distribution \citep{sensoy2018evidential}.
The network's output is interpreted as the parameters of a Dirichlet distribution, so the predictions followed the Dirichlet distribution.
The Dirichlet distribution will be sharp if the model is confident about the predictions. 
Otherwise, it will be flat. 
Another advantage brought by EDL is that AU and EU are easy to obtain with a Dirichlet distribution.
{
In chest X-ray classification, \cite{balaram2022consistencybased} modified the EDL-based AL to accommodate the multi-label setting. 
Specifically, they transformed the Dirichlet distribution in EDL into multiple Beta distributions, each corresponding to one class label. 
They then calculated the entropy of the Beta distributions as the AU for annotation.
\cite{chen2023think} proposed a federated AL method based on EDL for medical image analysis.
Following the federated AL setting, they kept a global model across all clients and local models for each client.
AUs of the global and local models and the EUs of the global models were used for sample selection.
}
\cite{park2023active} introduced a model evidence head to scale the parameters of the Dirichlet distribution adaptively in object detection, which enhanced training stability. 
They first calculated the EU for each detection box.
Then, the sample-level uncertainty was obtained through hierarchical uncertainty aggregation. 
{Besides, \cite{xie2022dirichletbased} introduced EDL into active domain adaptation.
Samples with high distribution and data uncertainties are selected for annotation, which are both based on EDL.}

\noindent \textbf{Mixture density networks:} 
\cite{choi2021active} transformed the classification and localization heads in object detection networks to the architecture of MDN \citep{bishop1994mixture}. 
Besides the coordinates and class predictions of each bounding box, the MDN heads produced the variance of classification and localization. 
They used the variances as uncertainty metrics for sample selection.
Results showed that this method is competitive with MC dropout and model ensemble while significantly reducing the inference time and model size.

\subsubsection{Adversarial-based Uncertainty}
\label{adversarial}

Uncertainty in AL can also be estimated adversarially, including adversarial samples and adversarial training. 

\noindent \textbf{Adversarial samples} {help measure the sample's distance to the decision boundary implicitly, while a higher distance indicates higher uncertainty.
By attacking the deep models, adding carefully designed perturbations to original samples results in adversarial samples \citep{goodfellow2014explaining}.}
The differences between adversarial and original samples are nearly indiscernible to the human eye.
However, deep models would produce extremely confident but wrong predictions for adversarial samples.
The reason is that adversarial attacks push the original samples to the other side of the decision boundary with minimal cost, resulting in visually negligible changes but significantly different predictions. 
From this perspective, the strength of adversarial attacks reflects the sample's distance to the decision boundary \citep{heo2019knowledge}.
A small perturbation indicates that the sample is closer to the decision boundary and, thus, is considered more uncertain. 
\cite{ducoffe2018adversarial} adopted the DeepFool algorithm \citep{moosavi2016deepfool} for adversarial attacks.
Samples with small adversarial perturbations are requested for labeling. 
\cite{rangwani2021s3vaada} attacked the deep model by maximizing the KL divergence between predictions of adversarial and original samples while the strength of perturbation is limited.

\noindent \textbf{Adversarial training} alternates between training feature extractors and classifiers with conflicting objectives, aiming to expose uncertain samples by increasing classifier disagreements. 
\cite{yuan2021multiple} and their extended work \citep{wan2023multiple} implemented this with two classifiers on labeled and unlabeled datasets, first tuning classifiers while fixing the feature extractor to reveal more uncertain samples, then adjusting the feature extractor against fixed classifiers to minimize the gap between labeled and unlabeled samples. 
After several rounds, samples with the greatest disagreements are annotated.

\subsection{Evaluation of Informativeness: Representativeness}
\label{representativeness}

While uncertainty-based methods play a crucial role in deep AL, they still face certain challenges:
\textbf{1. Outlier selection:} 
The goal of using uncertainty in AL is to improve performance by querying hard samples of the current model. 
However, these methods could also select outliers that harm the model training \citep{karamcheti2021mind}.
This happens mainly because uncertainty-based methods often ignore the intrinsic characteristics of the sample itself. 
\textbf{2. Distribution misalignment:} 
In the feature space, uncertain samples are often located near the decision boundary \citep{settles2009active}.
Therefore, the distribution of samples selected by uncertainty-based methods is usually different from the overall data distribution. 
This discrepancy introduces dataset bias and leads to a performance drop. 
{
This challenge could be alleviated if the relationship between different samples is carefully considered during the AL query.
In summary, uncertainty-based AL lacks exploration of the visual information carried in each sample and the relationship between different samples.
These challenges above call for a new informativeness metric in AL.
}

\begin{figure}[!ht]
  \centering
  \includegraphics[width=\columnwidth]{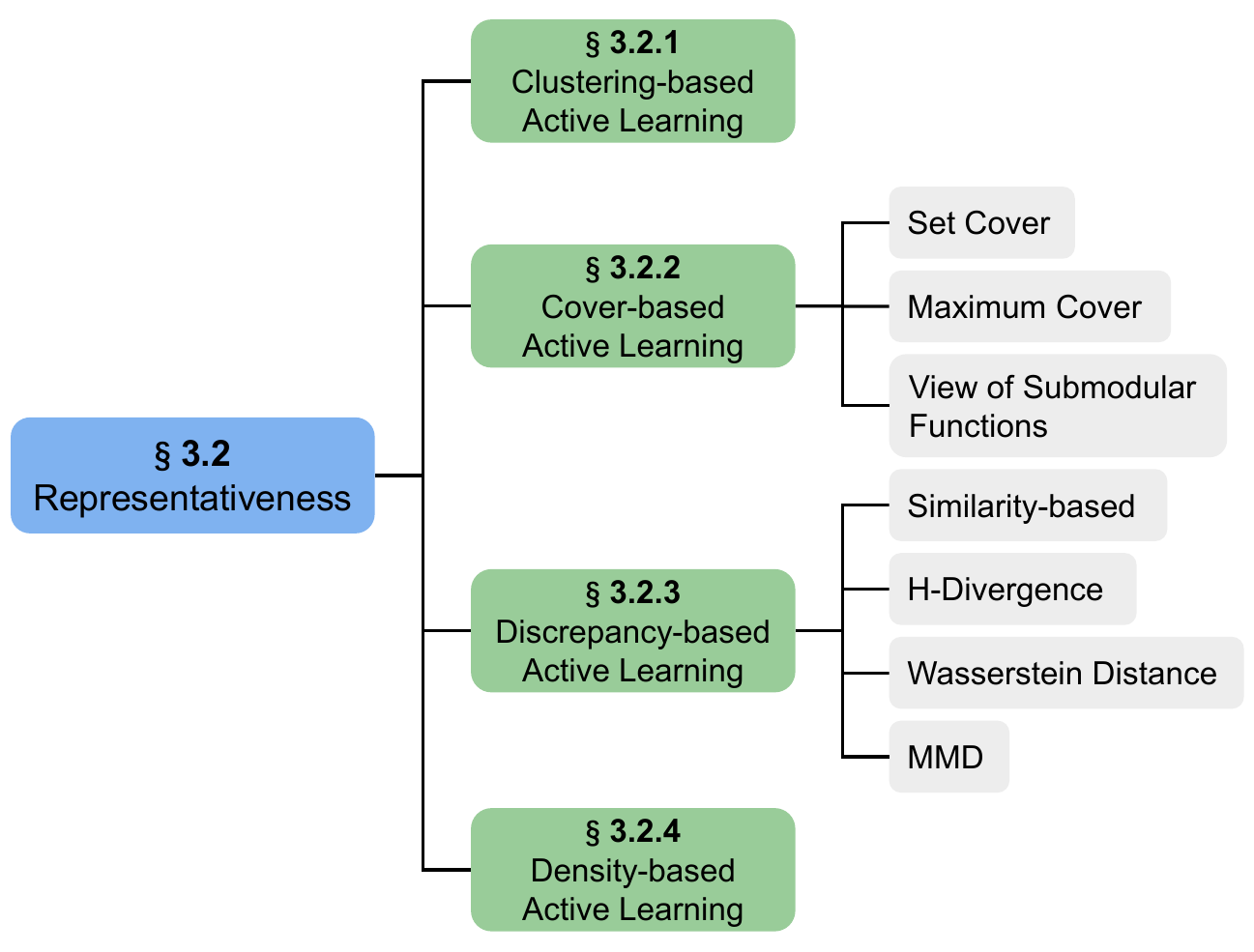}
  \caption{The taxonomy of representativeness-based active learning.}
  \label{fig_repre}
\end{figure}

{
Representativeness is adopted in AL to overcome challenges brought by uncertainty.}
Representativeness-based AL aims to select a subset of samples that can represent the entire dataset. 
{
Specifically, representative samples should be visually distinctive in properties like imaging style or visual content.
In medical image analysis, images are often high-dimensional and thus computation-intensive for the DL model. 
Besides, important information like lesions or tissues is not always directly visible or easily distinguishable.
A good feature representation greatly reduces the image dimensionality and also extracts anatomical, histological, pathological, or even functional information in medical images. 
Therefore, the query process of representativeness-based AL is often conducted in the feature space.
Besides, representative samples should also be widely distributed across the data distribution rather than concentrated in a specific region.
In other words, these samples should be diverse.
This is to minimize redundancy in the query result and try to keep the original data distribution as much as possible. 
Therefore, proper metrics of sample-wise or distribution-wise distance and sample density are needed to assess the landscape of a dataset. 
Besides, the uniqueness of medical images may require different distance metrics than that of natural images. 
}
This section introduces four types of representativeness-based AL: clustering-based, cover-based, discrepancy-based, and density-based representativeness AL.
The taxonomy of these methods is shown in Fig. \ref{fig_repre}.

\subsubsection{{Clustering-based Active Learning}}
\label{cluster}
{
With the advancement of feature extraction in medical image analysis, images with similar appearances tend to group together in the feature space \citep{zheng2019biomedical}.
Therefore, a straightforward approach involves clustering on the data embeddings to select representative samples.
This type of method grouped the data into several clusters and then selected the centroid samples of each cluster. 
It leveraged the inherent structure within the data for insightful groupings and was also very easy to implement.
K-Means was the most popular choice in clustering-based active learning. 
\cite{pourahmadi2021simple} performed k-Means on the off-the-shelf self-supervised features, then selected cluster centers for annotation. 
Based on the self-supervised feature, \cite{jin2022oneshot} adopted the k-Means++ for clustering and the silhouette coefficient to determine the optimal number of clusters. 
Their proposed method achieved commendable performance in lung segmentation of chest X-rays and lesion segmentation of dermoscopic images.
In nuclei segmentation, \cite{lou2023which} performed coarse-level and fine-level clustering using K-Means, aiming to select informative patches from pathological images.
In connectomics, \cite{lin2020two} proposed two-stream clustering for active selection. 
They first predicted the semantic mask for each unlabeled sample and simplified the AL task to judge the correctness of each predicted ROI.
Besides, they trained two feature extractors of VAE with segmentation masks and unlabeled images, respectively. 
For two-stream clustering, they first applied mask-level clustering with mask features to group ROIs with similar appearances.
Within each mask cluster, image-level clustering is further performed. 
This method achieved excellent performance in synapse detection and mitochondria segmentation.
Results also showed that two-stream clustering outperforms clustering with concatenated mask and image features by preventing the image feature from dominating the results.
}

\subsubsection{Cover-based Active Learning}
\label{cover}

We can formulate representativeness-based AL as a covering problem.
A classic example of the covering problem is facility location, such as covering all the city's streets with billboards \citep{farahani2009facility}.  
Likewise, cover-based AL uses a few samples to cover the entire dataset{, which is analogous to using several spheres to cover all the samples in the feature space, with the center of each sphere being the selected sample}.
Ideally, these samples should be representative and contain information on other samples.
These methods usually involve two settings: set cover and maximum cover. 
Both settings are NP-hard, meaning they cannot be optimally solved in polynomial time.
However, near-optimal solutions could be achieved in linear time using greedy algorithms, which iteratively select samples that cover most of the other samples for annotation \citep{feige1998threshold}.
{
These two variants are slightly different in the problem setting. 
The set cover is constrained by complete coverage, which means that it cannot omit any sample in the dataset. 
To achieve this goal, the radius of its covering spheres may be very large, and when there are very few samples selected, outliers might be chosen as the centers of the spheres \citep{yehuda2022active}. 
The goal of max cover is to cover as much of the entire dataset as possible, breaking the constraint of set cover, and thus avoiding the issue of outlier selection.
}

\noindent \textbf{Set cover:} 
Core-Set \citep{sener2018active} followed the setting of k-Center location \citep{hochbaum1985best}, which is also a variant of the set cover problem. 
They employed farthest-first traversal to solve the k-Center problem for selecting representative samples. 
The L2 distance of deep features is used to measure the similarity between different samples.
\cite{agarwal2020contextual} introduced contextual diversity for AL, a metric that fused uncertainty and diversity of samples spatially and semantically. 
They replaced the L2 distance with contextual diversity and adopted the same farthest-first traversal for sample selection as \cite{sener2018active}.
\cite{caramalau2021sequential} adopted graph convolutional networks (GCN) to model the relationships between labeled and unlabeled samples. 
GCNs improved the feature representation of unlabeled samples with the labeled dataset.
Enhanced feature representation was further used for Core-Set sampling.

\noindent \textbf{Maximum cover:} 
{
As a pioneering work of AL in medical image analysis, SA \citep{yang2017suggestive} stands out as one of the initial endeavors to introduce the concept of representativeness into AL.  
}
SA first selected highly uncertain samples and then chose representative samples for annotation.
{
The formulation of the representativeness part in SA followed the setting of maximum cover.
}
The representativeness metric was based on the cosine similarity of deep features.
Specifically, sample $x$ is represented by the most similar sample from queried dataset $D_{t}^q$

\begin{equation}
    r\left(D_{t}^q, x\right) = \underset{x^\prime \in D_{t}^q}{\max}{sim\left(x^\prime, x\right)}
    \label{sa1}
\end{equation}

\noindent where $r$ is the representativeness of sample $x$ with respect to $ D_{t}^q$ and $sim(\cdot, \cdot)$ represents cosine similarity.
Besides, representativeness $R$ between $D_{t}^q$ and the unlabeled set $D_t^u$ is as follow:

\begin{equation}
    R\left(D_{t}^q, D_t^u\right) = \underset{x\in D_t^u}{\sum} r\left(D_{t}^q, x\right)
    \label{sa2}
\end{equation}

\noindent where a larger $R\left(D_{t}^q,D_t^u\right)$ indicates that $D_{t}^q$ better represents $D_t^u$. 
It should be noted that SA is a generalization of the maximum cover problem since the cosine similarity ranges from 0 to 1. 
But they still employed a greedy algorithm to find sample $x$ that maximizing $R\left(D_{t}^q \cup x, D_t^u\right) - R\left(D_{t}^q, D_t^u\right)$. 
{Many subsequent AL works built their framework of cover-based AL on SA, especially in the field of medical image analysis. }
\cite{xu2018quantization} quantized the segmentation networks in SA and found that it improved the accuracy of gland segmentation while significantly reducing memory usage. 
\cite{zheng2019biomedical} proposed representative annotation (RA), which omits the uncertainty query in SA.
RA trained a VAE for feature extraction and partitioned the feature space using hierarchical clustering.
They selected representative samples in each cluster using a similar strategy to SA. 
{RA achieved superior performance in gland segmentation on histological images, fungus segmentation on electron microscopy images, and whole heart segmentation on MRI.}
{In breast cancer segmentation on immunohistochemistry images,} \cite{shen2020deep} changed the similarity measure in SA from $sim(\cdot, \cdot)$ to $ 1-sim(\cdot, \cdot)$, which enhanced the diversity of the selected samples. 
{Additionally, some works follow different formulations than those of SA in maximum cover.}
In keypoint detection of medical images, \cite{quan2022which} proposed a representative method to select template images for few-shot learning. 
First, they trained a feature extractor using self-supervised learning and applied the {scale-invariant feature transform descriptor} for initial keypoint detection. 
Next, they calculated the average cosine similarity between template images and the entire dataset. 
Finally, they picked the template combination with the highest similarity for annotation.
\cite{yehuda2022active} found that Core-Set \citep{sener2018active}, which followed the setting of the set cover, tends to select outliers, especially when the annotation budget is low.
To address this issue, they proposed ProbCover, which changed the setting from set cover to maximum cover.
With the help of self-supervised deep features and a graph-based greedy algorithm, ProbCover effectively avoided outlier selection in cover-based AL.

\noindent \textbf{View of submodular functions:}
Both set cover and maximum cover can be formulated from the perspective of submodular set functions \citep{fujishige2005submodular}. 
{
These functions show diminishing returns.
Specifically, given two sets $A$ and $B$, $A \subset B$, for every element $z$ that not in $B$, a submodular set function $g$ has that $g\left( A \cup {z} \right) - g\left( A \right) \geq g\left( B \cup {z} \right) - g\left( B \right)$.
This property makes submodular set functions suitable for AL. 
Suppose the informativeness function $I$ is submodular. 
It means that each newly queried sample brings less informativeness gain than the previous one, which indicates that highly informative samples should be queried first.
Besides, if we can formulate optimization problems in terms of monotonic and submodular functions, we can use a greedy algorithm to get near-optimal solutions in linear time.
For AL, if $I$ is submodular and monotonic, it means that we could greedily select the samples that maximize $I$.
}
In cover-based AL, methods like SA and RA followed the setting of submodular functions, but the authors didn't present their methods from this perspective.
Introducing submodular functions would extend the formulation of AL and ensure the selected samples are both representative and diverse.
Typical steps for this type of method involve calculating sample similarities, constructing a submodular optimization problem, and solving it using a greedy algorithm \citep{wei2015submodularity}. 
\cite{kothawade2021similar} introduced an AL framework based on submodular information measures, effectively addressing issues such as scarcity of rare class, redundancy, and out-of-distribution data. 
In object detection, \cite{kothawade2022talisman} focused on samples of minority classes. 
{
They first constructed a reference dataset containing samples of certain classes of interest.
Then, unlabeled samples similar to the reference set for annotation through submodular mutual information (SMI).
SMI is used to measure the similarity between two sets. Suppose two sets $A$, $B$ and a submodular function $g$, the SMI is defined as $\mathcal{I}_{SMI} = g(A) + g(B) - g(A \cup B)$. 
Please refer to \cite{kothawade2022prism} for more detailed definitions of SMI. }

\subsubsection{Discrepancy-based Active Learning}
\label{discrepancy}

In discrepancy-based AL, unlabeled samples farthest from the labeled set are considered the most representative. 
The main idea is that if we queried such samples for multiple rounds, the discrepancy between the distributions of labeled and unlabeled sets would be significantly reduced. 
Therefore, a small set of samples could well represent the entire dataset.
The key to these methods is measuring the discrepancy (i.e., distance) between two high-dimensional distributions. 
{
In this section, we present four discrepancies between probability distributions: similarity-based discrepancy, H-divergence, Wasserstein distance, and maximum mean discrepancy (MMD).
}

\noindent \textbf{Similarity-based discrepancy:}
As a practical and easy-to-implement metric, we can approximate the distance between distributions based on sample similarity. 
{In gland and MRI infant brain segmentation, \cite{li2020attention} adopted the average cosine similarity as the distance between two datasets.
They selected samples far from the labeled set and close to the unlabeled set.}
\cite{caramalau2021sequential} proposed UncertainGCN, which employed GCN to model the relationship between labeled and unlabeled samples.
They selected the unlabeled samples with the lowest similarity to the labeled set. 
In object detection, \cite{wu2022entropy} constructed prototypes with sample features and prediction entropy. 
They selected unlabeled samples that were far from the labeled prototype.

\noindent \textbf{H-divergence} estimates the distance of distribution with the help of the discriminator from generative adversarial networks (GAN) \citep{goodfellow2014generative}. 
More specifically, the discriminator tries to distinguish between labeled and unlabeled samples, and there is a close relationship between H-divergence and the discriminator's output \citep{gissin2019discriminative}. 
Variational adversarial active learning (VAAL) \citep{sinha2019variational} combined VAE with a discriminator for discrepancy-based AL. 
In VAAL, the VAE mapped samples to a latent space while the discriminator distinguished whether samples were labeled. 
These two are mutually influenced by adversarial training.
VAE tried to fool the discriminator into judging all samples as labeled while the discriminator attempted to differentiate between labeled and unlabeled samples correctly. 
After multiple rounds of adversarial training, VAAL selected samples that the discriminator deemed most likely to be unlabeled for annotation.
VAAL inspired many subsequent works.  
{
\cite{khanal2023m} adopted multimodal information to improve VAAL.
For multimodal medical images, they modified the VAE to reconstruct images of both modalities using the latent code of only one modality. 
The proposed method was evaluated on brain tumor segmentation, classification, and chest X-ray classification.
}
\cite{gissin2019discriminative} trained the discriminator without adversarial training. 
\cite{zhang2020staterelabeling} replaced the discriminator's binary label with sample uncertainty.
They also combined features of VAE with features from the supervised model.
\cite{wang2020dual} adopted a neural network module for sample selection.
To train such a module, they added another discriminator on top of VAAL, which aimed to differentiate between the real and VAE-reconstructed features for unlabeled samples. 
After adversarial training of both discriminators, the module selected uncertain and representative samples.   
\cite{kim2021taskaware} combined learning loss for active learning with VAAL, feeding both loss ranking predictions and VAE features into the discriminator.

\noindent \textbf{Wasserstein distance} is widely used for computing distribution distances. 
\cite{shui2020deep} indicated that H-divergence compromises the diversity of sample selection, while Wasserstein distance ensures the queried samples are representative and diverse. 
They further proposed Wasserstein adversarial active learning (WAAL), which built upon VAAL and adopted an additional module for sample selection. 
They trained this module by minimizing the Wasserstein distance between labeled and unlabeled sets. 
WAAL selected samples that are highly uncertain and most likely to be unlabeled for annotation.
\cite{mahmood2022lowbudget} formulated AL as an optimal transport problem.
They aimed at minimizing the Wasserstein distance between the labeled and unlabeled sets with self-supervised features.
They further adopted mixed-integer programming that guarantees global convergence for diverse sample selection. 
Moreover, \cite{xie2023active} considered the candidates as continuously optimizable variables based on self-supervised features. 
They randomly initialized the candidate samples at first.
Then, they maximized the similarity between candidates and their nearest neighbors while minimizing the similarity between candidates and labeled samples. 
Finally, they selected the nearest neighbors of the final candidates for annotation. 
They proved the objective is equivalent to minimizing the Wasserstein distance between the labeled and unlabeled samples.

\noindent {\textbf{Maximum mean discrepancy} measures the distance of two distributions as the distances between their mean features with kernel trick \citep{gretton2012kernel}.
In active domain adaptation (will be detailed in §\ref{ada}), \cite{hwang2022combating} adopted MMD to measure the distance between the source and target domain.
Then, MMD was used to select representative and diverse samples in the target domain.
It should be noted that the Wasserstein distance belongs to the family of integral probability metrics (IPM), while MMD simultaneously falls into the range of IPM and previously mentioned H-divergence. 
Please refer to \cite{zhao2022comparing} for a more detailed taxonomy of the discrepancy between probability distributions.
}

\subsubsection{Density-based Active Learning}
\label{density}
Density-based active learning tends to select samples from the most densely populated area of the data distribution. 
It employs density estimation to characterize the data distribution in a high-dimensional feature space.
The likelihood is the estimated density of the data distribution, and a more densely populated area indicates a higher likelihood. 
In this case, representative samples are samples with high likelihood.
However, such methods can easily cause redundancy in sample selection. 
As a result, techniques like clustering are frequently used to improve diversity in sample selection. 
Density-based AL directly estimates the data distribution, which prevents the need to solve complex optimization problems. 
{
In shoulder MRI musculoskeletal segmentation, \cite{ozdemir2021active} adopted infoVAE \citep{zhao2017infovae} to estimate the density of each sample in the labeled dataset and unlabeled pool.
Specifically, MMD replaced the KL divergence as the regularization term in the training of infoVAE.
The posterior probability by the encoder was used as the density metric.
Samples with higher density regarding the unlabeled pool and lower density regarding the labeled dataset were selected for annotation. 
}
TypiClust \citep{hacohen2022active} projected samples to a high-dimensional feature space via a self-supervised encoder.
The density of a sample was defined as the reciprocal of the L2 distances to its k-nearest neighbors. 
Additionally, TypiClust performed clustering beforehand to ensure the diversity of selected samples. 
\cite{wang2022unsupervised} proposed two variants of density-based AL.
The first variant fixed the feature representation.
The process was similar to TypiClust, but they maximized the distances between selected samples to ensure diversity. 
The other variant was in an end-to-end fashion.
Feature representation and sample selection were trained simultaneously.
This variant used a learnable k-Means clustering to jointly optimize cluster assignment and feature representation with a local smoothness constraint.

{
It is worth noting that cover-based and density-based AL differ in both concept and methodology.
In concept, samples in cover-based AL tend to cover the entire dataset.
However, they do not have to lie in the densest area of the data distribution.
For example, \cite{yehuda2022active} showed that Core-Set \citep{sener2018active}, a popular cover-based method, tends to select outliers in the low-budget regime. 
In this case, cover-based AL is opposite to density-based AL, which also indicates that density-based AL may be a better choice in the low-budget regime.
From the perspective of methodology, cover-based AL needs to solve an NP-hard problem in linear time with a greedy algorithm.
Although this algorithm results in acceptable solutions, it’s almost impossible to know how the AL performance would be if the optimal solutions could be achieved. 
For density-based AL, the NP-hard problem is replaced with density estimation, which is more computation-efficient. 
}


\subsection{Sampling Strategy}
\label{sampling_strategy}

\begin{figure}[!ht]
  \centering
  \includegraphics[width=0.9\columnwidth]{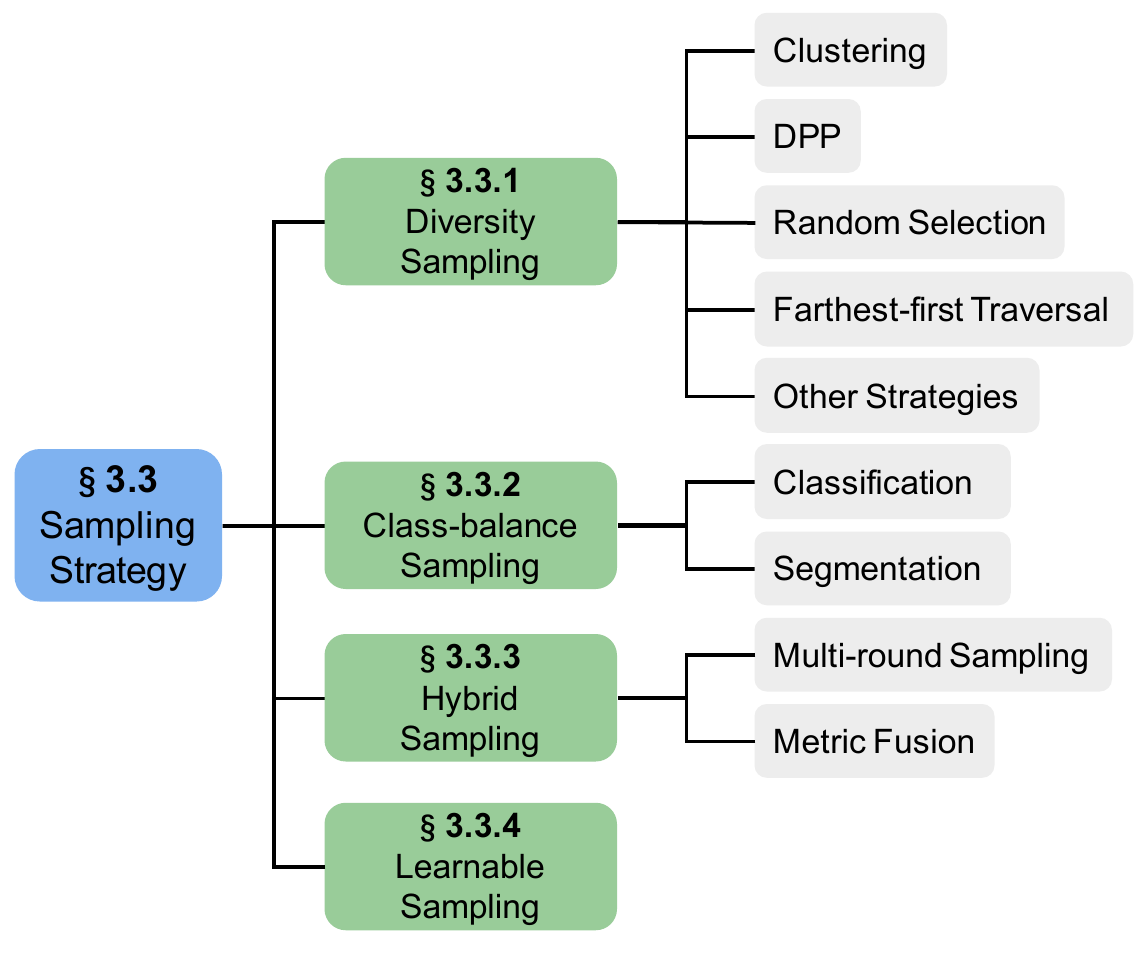}
  \caption{The taxonomy of different sampling strategies in active learning.}
  \label{fig_sample}
\end{figure}

With a well-developed informativeness metric, most deep AL works simply adopted top-k to select samples with the highest informativeness for annotation.
However, existing informativeness metrics face several issues, such as redundancy and class imbalance.
{
These issues are exacerbated due to the unique characteristics of medical images.
Despite the high variability, medical images of the same region-of-interest (ROI) could be classified into several groups, and images within each group share a high similarity \citep{zheng2019biomedical}. 
Also, class imbalance is notorious in medical image analysis since the healthy objects often outnumber the diseased ones.
}
Instead of proposing a better informativeness metric, we can improve the sampling strategy upon the top-k selection to effectively resolve these issues above.
Besides, specific sampling strategies can also be used for combining multiple informativeness metrics. 
Furthermore, with the recent development of deep AL, more and more studies directly employ neural networks for sample selection. 
In this context, we no longer evaluate informativeness but directly choose informative samples with neural networks. 
Regrettably, despite the importance of sampling strategies in AL, prior works or surveys have seldom discussed their specific attributes.
As one of the contributions of this survey, we systematically summarize different sampling strategies in AL, including diversity sampling, class-balanced sampling, hybrid sampling, and learnable sampling.
The taxonomy of different sampling strategies in AL is shown in Fig. \ref{fig_sample}.

\subsubsection{Diversity Sampling}
\label{diversity}

Diversity strategies aim to reduce sampling redundancy in active learning,  meaning certain selected samples are highly similar to each other.
The lack of diversity leads to the waste of the annotation budget.
Besides, redundancy in the training set causes the deep models to overfit to limited training samples, thus leading to a performance drop. 
Therefore, many AL methods employ diversity sampling to mitigate the redundancy in selected samples. 
In this section, we discuss four strategies of diversity sampling, including clustering, farthest-first traversal, determinantal point process (DPP), and specific strategies tailored to certain informativeness metrics.

\noindent \textbf{Clustering} is one of the most commonly used strategies of diversity sampling. 
This strategy improves the coverage of the entire feature space, thereby easily boosting diversity. 
\cite{ash2020deep} employed k-Means++ clustering on gradient embeddings to select diverse uncertain samples. 
Besides, \cite{citovsky2021batch} boosted margin-based uncertainty sampling with hierarchical clustering.
They selected samples with the smallest margins within each cluster. 
When the number of queries exceeded the number of clusters, samples from smaller clusters were prioritized. 
This method can scale to a huge annotation budget (e.g., one million). 
{\cite{zheng2019biomedical} incorporated hierarchical clustering with cover-based AL.
In their experiments, clustering showed consistent performance improvement in multiple medical imaging datasets, which demonstrated that clustering does improve the sampling diversity. }

{
It is important to highlight that clustering in this section is different from that of in §\ref{cluster}. 
To ensure that the selected samples are sufficiently representative, clustering-based AL generally chooses samples that are closest to the cluster centers. 
However, when clustering is used to enhance diversity, we can not only select samples closest to the cluster centers, but also select samples with the highest uncertainty, or even randomly select within each cluster. 
Therefore, clustering can serve as a plug-and-play technique to conveniently enhance the sampling diversity in AL.}

\noindent \textbf{Determinantal point process} is a stochastic probability model for selecting subsets from a larger set.
DPP reduces the probability of sampling similar elements to ensure diversity in the results. 
\cite{biyik2019batch} employed two DPPs for sample selection: Uncertainty DPP is based on uncertainty scores, while Exploration DPP aims to find samples near decision boundaries. 
Then, sampling results from both DPPs were sent for expert annotation. 
However, DPP is more computationally intensive compared to clustering. 
\cite{ash2020deep} compared the performance and time cost of using k-Means++ and k-DPP.
Results showed that their performance is similar, but the time cost for k-Means++ is significantly lower than that for k-DPP.
{
Besides, \cite{mi2020learning} adopted DPP in AL for medical image reconstruction, please refer to §\ref{med_reconstruction} for details. }

\noindent {
\textbf{Random Selection} could also be used for better diversity.
In prostate segmentation of MRI, \cite{gaillochet2023active} randomly partitioned the entire dataset into different batches, which were referred to as `stochastic batches'. 
Batches with the highest uncertainty scores were selected for annotation.
Experimental results showed that the stochastic batches consistently improved the performance of various uncertainty-based AL methods under an extremely low budget.
Their extended works \citep{GAILLOCHET2023102958} further illustrated the effectiveness of stochastic batches on the anterior and posterior hippocampus segmentation.
}

\noindent \textbf{Farthest-first Traversal} is also a widely used strategy for diverse queries, which was first adopted by \cite{sener2018active}.
This strategy requires the distance between sampling points to be as large as possible, which leads to a more uniform distribution of selected samples in the feature space.
{
\cite{li2023hal} adopted a farthest-first traversal strategy with cosine distance for a diverse initial labeled dataset.
Experiments on breast ultrasound, liver CT, and chest X-ray segmentation showed significant effectiveness of the farthest-first traversal.
}
Besides, \cite{agarwal2020contextual} and \cite{caramalau2021sequential} improved the diversity with farthest-first traversal with their proposed contextual diversity and GNN-augmented features, respectively.

\noindent \textbf{Other strategies:} 
In uncertainty-based AL, BatchBALD \citep{kirsch2019batchbald} extended BALD-based uncertainty AL to batch mode.
Results showed that BatchBALD improved the sampling diversity compared to \citep{gal2017deep}. 
FI-based methods formulated AL as a semi-definite programming (SDP) problem to improve sampling diversity and various methods were employed for solving SDP.
\cite{sourati2019intelligent} used a commercial solver to solve SDP, while \cite{ash2021gone} proposed a greedy algorithm to adapt to high-dimensional feature space. 
{
In skin lesion analysis, \cite{shi2019active} introduced image hashing for diversity sampling.
In their proposed method, the first principal component of each image was used for feature representation.
Then they mapped similar images into the same buckets using local sensitivity hashing.
Samples were uniformly selected from each bucket for human annotation.
}

\subsubsection{Class-balance Sampling}
\label{class-balance}

Class imbalance is a common issue for DL in medical image analysis, where a small set of classes have many samples while the others only contain a few samples \citep{zhang2023deep}. 
{For example, long tail distribution of classes existed in almost all tasks of medical image classification, such as skin lesion classification and whole-slide image classification. }
Training on imbalanced datasets can lead to the overfitting of the majority classes and underfitting of the minority classes. 
Apart from dealing with class imbalance during training, AL mitigates class imbalance by avoiding over-annotation of the majority classes and enhancing the annotation of the minority classes during dataset construction.

\noindent \textbf{Classification:}
{
In a class-imbalanced COVID-19 dataset, \cite{chong2021evaluation} evaluated multiple informativeness scores and sampling strategies.
Results showed that diversity sampling is more favored for class-imbalance.
}
\cite{jin2022deep} assumed that samples closer to the tail of the distribution are more likely to belong to the minority classes. 
Thus, the tail probability is equivalent to the likelihood of minority classes.
Specifically, they trained a VAE for feature extraction and adopted copula to estimate the tail probabilities upon VAE features. 
Finally, informative samples were selected with clustering and unequal probability sampling. 
The proposed method was validated on the ISIC 2020 dataset, which has a long-tailed distribution. 
\cite{kothawade2022clinical} used submodular mutual information to focus more on samples of minority classes.
They achieved excellent results on medical classification datasets in five different modalities, including X-rays, pathology, and dermoscopy. 
In blood cell detection under microscopy, \cite{sadafi2019multiclass} requested expert annotation of a sample whenever its classification probability of the minority class exceeded 0.2.
Besides, \cite{choi2021vabal} directly estimated the probability of a classifier making a mistake for a given sample and decomposed it into three terms using Bayesian rules. 
First, they trained a VAE to estimate the likelihood of the data given a predicted class.
Then, an additional classifier was trained upon VAE features to estimate class prior probabilities and the probability of mislabeling a specific class. 
By considering all three probabilities, they successfully mitigated class imbalance in AL.
The proposed method achieved good performance on stepwise class-imbalanced CIFAR-10 and CIFAR-100 datasets. 
For uncertainty-based methods, \cite{bengar2022classbalanced} introduced an optimization framework to maintain class balance. 
They compensated the query of minority classes with the most confident samples of that class, leading to a more balanced class distribution in the queried dataset.

\noindent \textbf{Segmentation:}
Due to certain AL methods selecting regions instead of the entire image for annotation, there is a need to ensure that the selected regions contain rare or small objects {(e.g., optic chiasma or optic nerve in head and neck multi-organ segmentation)}. 
\cite{cai2021revisiting} and \cite{wu2022d2ada} both proposed class-balanced sampling strategies for such scenarios, as detailed in §\ref{region}.

\subsubsection{Hybrid Sampling}
\label{hybrid}

In AL, more and more works use multiple informativeness metrics simultaneously.
However, how to effectively integrate multiple metrics remains a critical issue.
This issue is addressed by the hybrid sampling discussed in this section. 
Two approaches to hybrid sampling are often used, including multi-round sampling and metric fusion.

\noindent \textbf{Multi-round sampling} first selects a subset of samples based on one particular informativeness metric and continues sample selection within this subset based on another informativeness metric.
{Multi-round sampling is widely used in AL for medical image analysis for its convenience \citep{shen2020deep, li2020attention, wang2021annotationefficient}.}
For example, SA \citep{yang2017suggestive} performed representativeness sampling based on uncertainty to reduce redundancy in the sampled set. 
Besides, \cite{wu2022d2ada} employed an adaptive strategy that sets dynamic weights to adjust the budget of representativeness and uncertainty sampling.
The weight of representativeness sampling is larger initially, while the situation is reversed in the latter phase. 
This is because representativeness methods can quickly spot typical data, while uncertainty methods continuously improve the model by querying samples with erroneous predictions.

\noindent \textbf{Metrics fusion} is another widely used approach of hybrid sampling.
It directly combines different informativeness metrics. 
For example, one could directly sum up all metrics and select the samples with the highest values for annotation. 
{
Metrics fusion is also widely used in AL of the medical domain \citep{li2024hybrid, zhou2021active, wu2021covidal}.
}
Besides, ranked batch-mode \citep{cardoso2017ranked} can adaptively fuse multiple metrics in AL.

\subsubsection{Learnable Sampling}
\label{learnable}

Previously mentioned AL methods typically follow a ``two-step" paradigm, which first involves the evaluation of informativeness and then selects samples based on specific heuristics (i.e., sampling strategy). 
However, learnable sampling skips the informativeness evaluation and directly uses neural networks for sample selection. 
In this context, the neural network is known as a ``neural selector".

One of the most common methods of learnable sampling is to formulate sample selection as a reinforcement learning (RL) problem, where the learner and the dataset are considered the environment, and the neural selector serves as the agent. 
The agent interacts with the environment by selecting a limited number of samples for annotation, and the environment returns a reward to train the neural selector.
{
In medical image classification, \cite{wang2020deep_RL} employed an actor-critic framework where the critic network is used to evaluate the quality of the samples selected by the neural selector. 
This method has performed excellently in lung CT disease classification and diabetic retinopathy classification of fundus images.
}
Besides, \cite{haussmann2019deep} adopted a probabilistic policy network as the neural selector. 
The rewards returned by the environment encouraged the neural selector to choose diverse and representative samples.
The neural selector is trained using the REINFORCE algorithm \citep{williams1992simple}. 
\cite{agarwal2020contextual} utilized contextual diversity as RL rewards and trains a bidirectional long short-term memory network as the neural selector. 

For more works on learnable sampling in AL, such as formulating AL as few-shot learning or training neural selectors by meta-learning, please refer to the survey of \cite{liu2022survey}.

\section{Integration of Active Learning and Other Label-Efficient Techniques} 
\label{integration}
As discussed in §\ref{intro}, the high annotation cost has severely dragged down the development of DL in medical image analysis.  
Despite the wide use of AL in medical image analysis, various methods have been proposed to reduce the large amount of labeled data required for training deep models, such as semi-supervised and self-supervised learning, etc. 
These methods, including active learning, are collectively called label-efficient deep learning \citep{jin2023label}.
Label-efficient learning is a broad concept that includes all related technologies designed to improve annotation efficiency. 
In the real-world practice of AL in medical image analysis, there is still room for higher label efficiency by integrating AL with other label-efficient techniques.
{
For the example of AL in medical image segmentation, since many samples were left unlabeled in the cycle of AL, we could further include them to achieve better performance by integrating AL with semi-supervised learning.
The rapid development of self-supervised learning in medical image analysis introduced many powerful pre-trained models \citep{taleb20203d}. 
These models are also valuable in AL of medical image analysis for their superior ability of feature extraction.
For another circumstance, since the ROI in medical imaging is usually small, we could select and annotate the informative regions that contain the ROI in AL instead of annotating the whole image.
As a result, integrating active learning with other label-efficient techniques holds significant potential to increase annotation efficiency.
}
However, existing surveys have not yet systematically organized and categorized this line of research. 
Hence, as one of the main contributions of this survey, we comprehensively reviewed the integration of AL with other label-efficient techniques, including semi-supervised learning, self-supervised learning, domain adaptation, region-based annotation, and generative models. 
Additionally, how each surveyed work integrated with other label-efficient techniques is summarized in Table \ref{tab2}.

\begin{table*}[!p]
\centering
\rotatebox[origin=c]{90}{%
\tabcolsep=3pt
\begin{varwidth}{\textheight}
\centering
\caption{Methodology summarization of surveyed active learning works.}
\label{tab2}\footnotesize
\resizebox{\textwidth}{!}{%
\begin{tabular}{ccccccccccccc}
\toprule
&\multirow{2}[2]{*}{Year}&\multirow{2}[2]{*}{Venues}&\multicolumn{2}{c}{Uncertainty}&\multicolumn{2}{c}{Representativeness}&\multirow{2}[2]{*}{Sampling Strategy}&\multirow{2}[2]{*}{SemiSL}&\multirow{2}[2]{*}{SelfSL}&\multirow{2}[2]{*}{ADA}&\multirow{2}[2]{*}{Region}&\multirow{2}[2]{*}{Generative}\\
\cmidrule[.06em](l{.3em}r{.3em}){4-5}
\cmidrule[.06em](l{.3em}r{.3em}){6-7}
& & &Method&Basic Metrics&Method&Basic Metrics& & & & & &\\
\midrule

\cite{zhu2017generative} & 2017 & arXiv & Single Model & Distance to Decision Boundary & - & - & Top-k & & & & & \checkmark \\ \midrule

\cite{zhou2017finetuning} & 2017 & CVPR & \makecell[c]{Single Model\\Multiple Inferences - Data Disagreement} & \makecell[c]{Entropy\\KL Divergence}& - & - & Hybrid - Fusion& & & & &\\ \midrule

\cite{gal2017deep} & 2017 & ICML & Multiple Inferences - MC Dropout& \makecell[c]{Entropy, BALD,\\Least Confidence, Variance} & - & - & Top-k & & & & &\\ \midrule

\cite{yang2017suggestive} & 2017 & MICCAI & Multiple Inferences - Model Disagreement & Variance & Cover-based& Cosine Similarity & Hybrid - Multi-round& & & & &\\ \midrule

\cite{wang2017costeffective} & 2017 & TCSVT & Single Model & Least Confidence, Margin, Entropy & - & - & Top-k & Pseudo-label & & & &\\ \midrule

\cite{ducoffe2018adversarial} & 2018 & arXiv & Adversarial Samples & Distance to Decision Boundary & - & - & Top-k & & & & &\\ \midrule

\cite{mackowiak2018cereals} & 2018 & BMVC & Multiple Inferences - Model Disagreement & Vote Entropy & - & - & Top-k & & & & Patch &\\ \midrule

\cite{xu2018quantization} & 2018 & CVPR & Multiple Inferences - Model Ensemble & Variance & Cover-based& Cosine Similarity & Hybrid - Multi-round& & & & &\\ \midrule

\cite{beluch2018power} & 2018 & CVPR & Multiple Inferences - Model Ensemble & \makecell[c]{Entropy, BALD,\\Least Confidence, Variance} & - & - & Top-k & & & & &\\ \midrule

\cite{sourati2018active} & 2018 & DLMIA & Gradient-based Uncertainty & Fisher Information & - & - & Diversity - Solve Programming Problem & & & & &\\ \midrule

\cite{sener2018active} & 2018 & ICLR & - & - & Cover-based& L2 Distance& Diversity - Farthest-first Traversal & & & & &\\ \midrule

\cite{kuo2018costsensitive} & 2018 & MICCAI & Multiple Inferences - Model Disagreement & JS Divergence & - & - & Diversity - Solve Programming Problem & & & & &\\ \midrule

\cite{mahapatra2018efficient} & 2018 & MICCAI & Multiple Inferences - MC Dropout& Variance & - & - & Top-k & & & & & \checkmark \\ \midrule

\cite{haussmann2019deep} & 2019 & IJCAI & - & - & - & - & Learnable - Reinforcement Learning & & & & &\\ \midrule

\cite{zheng2019biomedical} & 2019 & AAAI & - & - & Cover-based& Cosine Similarity & Diversity - Clustering & & \checkmark& & &\\ \midrule

\cite{gissin2019discriminative} & 2019 & arXiv & - & - & Discrepancy-based & H-Divergence & Top-k & & & & &\\ \midrule

\cite{yoo2019learning} & 2019 & CVPR & Performance Estimation - Learnable& Loss & - & - & Top-k & & & & &\\ \midrule

\cite{sinha2019variational} & 2019 & ICCV & - & - & Discrepancy-based & H-Divergence & Top-k & & & & &\\ \midrule



\cite{tran2019bayesian} & 2019 & ICML & Multiple Inferences - MC Dropout& BALD & - & - & Top-k & & & & & \checkmark \\ \midrule

\cite{qi2019labelefficient} & 2019 & JBHI & Single Model & Entropy & - & - & Top-k & Pseudo-label & & & &\\ \midrule

\cite{sadafi2019multiclass} & 2019 & MICCAI & Multiple Inferences - MC Dropout& Average IoU, Class Frequency& - & - & \makecell[c]{Class-balance\\Hybrid - Fusion} & & & & &\\ \midrule

\cite{kirsch2019batchbald} & 2019 & NeurIPS& Multiple Inferences - MC Dropout& BALD & - & - & Top-k & & & & &\\ \midrule

\cite{sourati2019intelligent} & 2019 & TMI & Gradient-based Uncertainty & Fisher Information & - & - & Diversity - Solve Programming Problem & & & & &\\ \midrule

\cite{kasarla2019regionbased} & 2019 & WACV & Single Model & Entropy & - & - & Top-k & & & & Superpixel&\\ \midrule

\cite{zheng2020annotation} & 2020 & AAAI & - & - & Cover-based& Cosine Similarity & Diversity - Clustering & Pseudo-label & & & Slice &\\ \midrule

\cite{shui2020deep} & 2020 & AISTATS& Single Model & Entropy, Least Confidence & Discrepancy-based & Wasserstein Distance & Hybrid - Fusion& & \checkmark& & &\\ \midrule

\cite{siddiqui2020viewal} & 2020 & CVPR & \makecell[c]{Multiple Inferences - MC Dropout\\Multiple Inferences - Data Disagreement}& \makecell[c]{Entropy\\KL Divergence}& - & - & Hybrid - Fusion& & & & Superpixel&\\ \midrule

\cite{zhang2020staterelabeling} & 2020 & CVPR & Single Model & Variance & Discrepancy-based & H-Divergence & Diversity - Farthest-first Traversal & & & & &\\ \midrule

\cite{gao2020consistencybased} & 2020 & ECCV & Multiple Inferences - Data Disagreement & Variance & - & - & Top-k & Consistency& & & &\\ \midrule

\cite{wang2020dual} & 2020 & ECCV & - & - & Discrepancy-based & H-Divergence & Learnable& & & & &\\ \midrule

\cite{agarwal2020contextual} & 2020 & ECCV & - & - & Cover-based& Contextual Diversity & \makecell[c]{Diversity - Farthest-first Traversal\\Learnable - Reinforcement Learning} & & & & &\\ \midrule

\cite{lin2020two} & 2020 & ECCV & - & - & Clustering-based & L2 Distance& Diversity - Clustering & & & & &\\ \midrule

\cite{ash2020deep} & 2020 & ICLR & Gradient-based Uncertainty & Gradient & - & - & Diversity - Clustering & & & & &\\ \midrule

\cite{casanova2020reinforced} & 2020 & ICLR & - & - & - & - & Learnable - Reinforcement Learning & & & & Patch &\\ \midrule

\cite{dai2020suggestive} & 2020 & MICCAI & Gradient-based Uncertainty & Gradient & - & - & \makecell[c]{Latent Space Optimization \&\\Nearest Neighbour Search} & & & & Slice &\\ 
\bottomrule
\end{tabular}}
\end{varwidth}}
\end{table*}

\begin{table*}[!p]
\centering
\setcounter{table}{1}
\rotatebox[origin=c]{90}{%
\tabcolsep=3pt
\begin{varwidth}{\textheight}
\centering
\caption{Methodology summarization of surveyed active learning works.}
\label{tab2}\footnotesize
\resizebox{\textwidth}{!}{%
\begin{tabular}{ccccccccccccc}
\toprule
&\multirow{2}[2]{*}{Year}&\multirow{2}[2]{*}{Venues}&\multicolumn{2}{c}{Uncertainty}&\multicolumn{2}{c}{Representativeness}&\multirow{2}[2]{*}{Sampling Strategy}&\multirow{2}[2]{*}{SemiSL}&\multirow{2}[2]{*}{SelfSL}&\multirow{2}[2]{*}{ADA}&\multirow{2}[2]{*}{Region}&\multirow{2}[2]{*}{Generative}\\
\cmidrule[.06em](l{.3em}r{.3em}){4-5}
\cmidrule[.06em](l{.3em}r{.3em}){6-7}
& & &Method&Basic Metrics&Method&Basic Metrics& & & & & &\\

\midrule

\cite{shen2020deep} & 2020 & MICCAI & \makecell[c]{Multiple Inferences - MC Dropout\\Performance Estimation - Surrogate} & \makecell[c]{Entropy\\IoU of all result} & Cover-based& Cosine Similarity & Hybrid - Multi-round& & & & &\\ \midrule

\cite{liu2020deep} & 2020 & MICCAI & Performance Estimation - Learnable& Loss & - & - & Top-k & & & & &\\ \midrule

\cite{li2020attention} & 2020 & MICCAI & Multiple Inferences - Model Ensemble & Margin & Discrepancy-based & Cosine Similarity & Hybrid - Multi-round& & & & &\\ \midrule

\cite{wang2020deep_RL} & 2020 & MICCAI & - & - & - & - & Learnable - Reinforcement Learning & & & & &\\ \midrule

\cite{hiasa2020automated} & 2020 & TMI & Multiple Inferences - MC Dropout& Variance & Cover-based& Cosine Similarity & Hybrid - Multi-round& & & & Slice, Pixel&\\ \midrule

\cite{huang2020rectifying} & 2020 & TMI & Multiple Inferences - Model Disagreement & Hausdorff Distance & - & - & Top-k & & & & &\\ \midrule

\cite{su2020active} & 2020 & WACV & Single Model & Entropy & Discrepancy-based & H-Divergence & Hybrid - Fusion& & & \checkmark & &\\ \midrule

\cite{choi2021vabal} & 2021 & CVPR & \multicolumn{2}{c}{Probability of Misclassification}& - & - & Class-balance & & & & &\\ \midrule

\cite{fu2021transferable} & 2021 & CVPR & Adversarial Training & Disagreement of Classifiers, margin & Discrepancy-based & H-Divergence & Hybrid - Fusion& & & \checkmark & &\\ \midrule


\cite{kim2021taskaware} & 2021 & CVPR & Performance Estimation - Learnable& Rank of Loss & Discrepancy-based & H-Divergence & Top-k & & & & &\\ \midrule

\cite{yuan2021multiple} & 2021 & CVPR & Adversarial Training & Disagreement of Classifiers & - & - & Top-k & & & & &\\ \midrule

\cite{cai2021revisiting} & 2021 & CVPR & Single Model & BvSB & - & - & Class-balance & & & & Superpixel&\\ \midrule

\cite{caramalau2021sequential} & 2021 & CVPR & Single Model (w/ GNN)& Margin & Cover-based& \makecell[c]{L2 Distance of\\GCN-augmented Features} & \makecell[c]{Top-k\\Diversity - Farthest-first Traversal}& & & & &\\ \midrule

\cite{prabhu2021active} & 2021 & ICCV & Single Model & Entropy & - & - & Diversity - Clustering & & & \checkmark & &\\ \midrule

\cite{ning2021multianchor} & 2021 & ICCV & - & - & Discrepancy-based & L2 Distance& Diversity - Clustering & & & & &\\ \midrule

\cite{huang2021semisupervised} & 2021 & ICCV & Performance Estimation - Surrogate & Temporal Output Discrepancy & - & - & Top-k & Consistency& & & &\\ \midrule

\cite{du2021contrastive} & 2021 & ICCV & - & - & Discrepancy-based & Semantic and distinctive scores& Hybrid - Fusion& & \checkmark& & &\\ \midrule

\cite{shin2021labor} & 2021 & ICCV & Multiple Inferences - Model Disagreement & Inequality & - & - & Diversity - Clustering & Pseudo-label & & \checkmark & Pixel, Point&\\ \midrule


\cite{rangwani2021s3vaada} & 2021 & ICCV & Adversarial Samples & KL Divergence & Cover-based - Submodular & \makecell[c]{KL Divergence\\Bhattacharya Coefficient} & Hybrid - Fusion& & & \checkmark & &\\ \midrule

\cite{choi2021active} & 2021 & ICCV & Uncertainty-aware Models - MDN& Variance & - & - & Top-k & & & & &\\ \midrule


\cite{liu2021influence} & 2021 & ICCV & Gradient-based Uncertainty & Influence & - & - & Top-k & & & & &\\ \midrule

\cite{zhao2021dsal} & 2021 & JBHI & Performance Estimation - Surrogate & Dice & - & - & Top-k & Pseudo-label & & & &\\ \midrule

\cite{zhou2021active} & 2021 & MedIA & \makecell[c]{Single Model\\Multiple Inferences - Data Disagreement} & \makecell[c]{Entropy\\KL Divergence}& - & - & Hybrid - Fusion& & & & &\\ \midrule

\cite{wu2021covidal} & 2021 & MedIA & \makecell[c]{Performance Estimation - Learnable\\Multiple Inferences - Data Disagreement} & \makecell[c]{Loss\\KL Divergence} & - & - & Hybrid - Fusion& & & & &\\ \midrule

\cite{zhou2021qualityaware} & 2021 & MICCAI & Performance Estimation - Learnable& Dice & - & - & Top-k & & & & &\\ \midrule

\cite{xu2021partiallysupervised} & 2021 & MICCAI & Single Model & Distance to Mean Probability& - & - & Top-k & Consistency& & & Patch &\\ \midrule

\cite{wang2021annotationefficient} & 2021 & MICCAI & Multiple Inferences - Model Ensemble & Variance & Discrepancy-based & Cosine Similarity & \makecell[c]{Diveristy - Clustering\\Hybrid - Multi-round}& Consistency& & & &\\ \midrule

\cite{nguyen2022goal} & 2021 & MIDL & Single Model & Entropy & Cover-based& L2 Distance& Diversity - Clustering & Pesudo-label & & & &\\ \midrule

\cite{ash2021gone} & 2021 & NeurIPS& Gradient-based Uncertainty & Fisher Information & - & - & Diversity - Solve Programming Problem & & & \checkmark & &\\ \midrule


\cite{kothawade2021similar} & 2021 & NeurIPS& - & - & Cover-based - Submodular & Gradient& Top-k & & & & &\\ \midrule

\cite{citovsky2021batch} & 2021 & NeurIPS& Single Model & Margin & - & - & Diversity - Clustering & & & & &\\ \midrule

\cite{nath2021diminishing} & 2021 & TMI & Multiple Inferences - Model Ensemble & Entropy & Discrepancy-based & Mutual Information & Hybrid - Fusion& & & & &\\ \midrule

\cite{mahapatra2021interpretabilitydriven} & 2021 & TMI & - & - & Saliency Maps& \makecell[c]{Kurtosis\\Multivariate Radiomics Features\\Deep Saliency Features} & Top-k & & & & &\\ \midrule

\cite{chen2021active} & 2021 & TPAMI & Single Model (in Feature Space) & Entropy & - & - & Top-k & & & & & \checkmark \\
\bottomrule
\end{tabular}}
\end{varwidth}}
\end{table*}

\begin{table*}[!p]
\centering
\setcounter{table}{1}
\rotatebox[origin=c]{90}{%
\tabcolsep=3pt
\begin{varwidth}{\textheight}
\centering
\caption{Methodology summarization of surveyed active learning works.}
\label{tab2}\footnotesize
\resizebox{\textwidth}{!}{%
\begin{tabular}{ccccccccccccc}
\toprule
&\multirow{2}[2]{*}{Year}&\multirow{2}[2]{*}{Venues}&\multicolumn{2}{c}{Uncertainty}&\multicolumn{2}{c}{Representativeness}&\multirow{2}[2]{*}{Sampling Strategy}&\multirow{2}[2]{*}{SemiSL}&\multirow{2}[2]{*}{SelfSL}&\multirow{2}[2]{*}{ADA}&\multirow{2}[2]{*}{Region}&\multirow{2}[2]{*}{Generative}\\
\cmidrule[.06em](l{.3em}r{.3em}){4-5}
\cmidrule[.06em](l{.3em}r{.3em}){6-7}
& & &Method&Basic Metrics&Method&Basic Metrics& & & & & &\\

\midrule

\cite{kothawade2022prism} & 2022 & AAAI & - & - & Cover-based - Submodular & Gradient& Top-k & & & & &\\ \midrule

\cite{xie2022active} & 2022 & AAAI & Single Model & Margin & Density-based& Energy & Hybrid - Multi-round& & & \checkmark & &\\ \midrule

\cite{wang2022boosting} & 2022 & AAAI & Gradient-based Uncertainty & Gradient & - & - & Top-k & & & & &\\ \midrule



\cite{xie2022learning} & 2022 & CVPR & Single Model & Margin, Gradient& - & - & Top-k & & & \checkmark & &\\ \midrule

\cite{zhang2022boostmis} & 2022 & CVPR & \makecell[c]{Single Model\\Adversarial Samples}& \makecell[c]{Entropy\\KL Divergence} & Density-based& Mean Cosine Similarity of KNN & Hybrid - Equal Split& Consistency& & & &\\ \midrule

\cite{zhang2022onebit} & 2022 & CVPR & Single Model & Entropy & - & - & Top-k & & \checkmark& & &\\ \midrule

\cite{parvaneh2022active} & 2022 & CVPR & Multiple Inferences - Data Disagreement & Inequality & - & - & Diversity - Clustering & & & & &\\ \midrule

\cite{xie2022fewer} & 2022 & CVPR & Single Model & Entropy & - & - & Top-k & & & \checkmark & Patch &\\ \midrule

\cite{quan2022which} & 2022 & CVPR & - & - & Cover-based& Cosine Similarity & Top-k & & & & &\\ \midrule

\cite{wu2022entropy} & 2022 & CVPR & Single Model & Entropy & Discrepancy-based & Cosine Similarity & Class-balance & & & & &\\ \midrule

\cite{wang2022unsupervised} & 2022 & ECCV & - & - & Density-based& KNN Density& Diversity - Clustering w/ Regularization & Consistency& \checkmark& & &\\ \midrule

\cite{kothawade2022talisman} & 2022 & ECCV & - & - & Cover-based - Submodular & Cosine Similarity & Top-k & & & & &\\ \midrule

\cite{chen2022when} & 2022 & ECCV & Gradient-based Uncertainty & Gradient & - & - & Top-k & & & & & \checkmark \\ \midrule


\cite{hwang2022combating} & 2022 & ECCV & Single Model & Margin & Discrepancy-based & MMD & Hybrid - Multi-round& Pseudo-label & & \checkmark & &\\ \midrule

\cite{yi2022pt4al} & 2022 & ECCV & Single Model & Least Confidence& Self-supervised Learning & Loss of Pretext Task & Hybrid - Multi-round& & \checkmark& & &\\ \midrule

\cite{wu2022d2ada} & 2022 & ECCV & Single Model & Entropy & Density-based& GMM & Hybrid - Multi-round& & & \checkmark & Superpixel&\\ \midrule

\cite{mahmood2022lowbudget} & 2022 & ICLR & - & - & Discrepancy-based & Wasserstein Distance & Diversity - Solve Programming Problem & & \checkmark& & &\\ \midrule

\cite{hacohen2022active} & 2022 & ICML & - & - & Density-based& \makecell[c]{Inverse Average Distance\\to KNN samples} & Diversity - Clustering & & \checkmark& & &\\ \midrule

\cite{jin2022coldstart} & 2022 & \makecell[c]{Information\\Sciences} & - & - & Clustering-based & Cosine Similarity & Diversity - Clustering & & \checkmark& & &\\ \midrule

\cite{jin2022deep} & 2022 & KBS & - & - & Clustering-based & L2 Distance& Class-balance & & \checkmark& & &\\ \midrule

\cite{jin2022oneshot} & 2022 & KBS & - & - & Clustering-based & L2 Distance& Diversity - Farthest-first Traversal & & \checkmark& & &\\ \midrule

\cite{dai2022suggestive} & 2022 & MedIA & Gradient-based Uncertainty & Gradient & - & - & \makecell[c]{Latent Space Optimization \&\\Nearest Neighbour Search} & & & & Slice &\\ \midrule

\cite{zhou2022volumetric} & 2022 & MedIA & Performance Estimation - Learnable& Dice & - & - & Top-k & & & & &\\ \midrule

\cite{atzeni2022deep} & 2022 & MedIA & Performance Estimation - Surrogate & Dice & - & - & Top-k & & & & &\\ \midrule

\cite{nath2022warm} & 2022 & MICCAI & Multiple Inferences - MC Dropout& Entropy & - & - & Top-k & Pseudo-label & & & &\\ \midrule

\cite{balaram2022consistencybased} & 2022 & MICCAI & Uncertainty-aware Model - EDL & Entropy & - & - & Top-k & \makecell[c]{Pseudo-label \&\\Consistency} & & & &\\ \midrule

\cite{wu2022selflearning} & 2022 & MICCAI & - & - & Cover-based& Cosine Similarity & Diversity - Clustering & & & & Slice &\\ \midrule

\cite{bai2022discrepancybased} & 2022 & MICCAI & Multiple Inferences - Model Disagreement & Entropy-weighted Dice Distance & - & - & Hybrid - Fusion& Pseudo-label & & & &\\ \midrule

\cite{kothawade2022clinical} & 2022 & MICCAIW& - & - & Cover-based - Submodular & Gradient& Top-k & & & & &\\ \midrule

\cite{yehuda2022active} & 2022 & NeurIPS& - & - & Cover-based& L2 Distance& Graph-based Algorithm & & & & &\\ \midrule

\cite{mahapatra2022graph} & 2022 & TMI & - & - & Saliency Maps& Graph-based Methods& Top-k & & & & &\\ \midrule

\cite{li2022pathal} & 2022 & TMI & \multicolumn{2}{c}{Curriculum Learning \& Noisy Sample Detection} & - & - & Top-k & Pseudo-label & & & &\\ \midrule

\cite{bengar2022classbalanced} & 2022 & WACV & Single Model & Entropy & - & - & Class-balance & & & & &\\ \midrule

\cite{xie2023active} & 2023 & CVPR & - & - & Discrepancy-based & Wasserstein Distance & \makecell[c]{Latent Space Optimization \&\\Nearest Neighbour Search} & & \checkmark& & &\\ \midrule

\cite{lyu2023boxlevel} & 2023 & CVPR & Multiple Inferences - Data Disagreement & Cross Entropy, Variance & - & - & Hybrid - Fusion& Pseudo-label & & & Box &\\ 
\bottomrule
\end{tabular}}
\end{varwidth}}
\end{table*}

\begin{table*}[!p]
\centering
\setcounter{table}{1}
\rotatebox[origin=c]{90}{%
\tabcolsep=3pt
\begin{varwidth}{\textheight}
\centering
\caption{Methodology summarization of surveyed active learning works.}
\label{tab2}\footnotesize
\resizebox{\textwidth}{!}{%
\begin{tabular}{ccccccccccccc}
\toprule
&\multirow{2}[2]{*}{Year}&\multirow{2}[2]{*}{Venues}&\multicolumn{2}{c}{Uncertainty}&\multicolumn{2}{c}{Representativeness}&\multirow{2}[2]{*}{Sampling Strategy}&\multirow{2}[2]{*}{SemiSL}&\multirow{2}[2]{*}{SelfSL}&\multirow{2}[2]{*}{ADA}&\multirow{2}[2]{*}{Region}&\multirow{2}[2]{*}{Generative}\\
\cmidrule[.06em](l{.3em}r{.3em}){4-5}
\cmidrule[.06em](l{.3em}r{.3em}){6-7}
& & &Method&Basic Metrics&Method&Basic Metrics& & & & & &\\

\midrule



\cite{jung2023simple} & 2023 & ICLR & Multiple Inferences - Model Ensemble & Entropy, Variance Ratio, BALD, Margin& - & - & Top-k & & & & &\\ \midrule

\cite{xie2022dirichletbased} & 2023 & ICLR & Uncertainty-aware Model - EDL & \makecell[c]{Mutual Information \& Entropy Expectation\\of Dirichlet Distribution} & - & - & Hybrid - Multi-round& & & \checkmark & &\\ \midrule

\cite{kim2023adaptive} & 2023 & ICCV & Single Model & BvSB & - & - & Class-balance & & & & Superpixel&\\\midrule

\cite{park2023active} & 2023 & ICLR & Uncertainty-aware Model - EDL & Mutual Information & - & - & Top-k & & & & &\\ \midrule


\cite{sadafi2023active} & 2023 & ISBI & \makecell[c]{Multiple Inferences - MC Dropout\\Multiple Inferences - Model Disagreement} & \makecell[c]{Variance\\Inequality} & - & - & Hybrid - Fusion& & & & &\\ \midrule

\cite{bai2023slpt} & 2023 & MICCAI & \makecell[c]{Multiple Inferences - Model Disagreement\\Gradient-based Uncertainty} & KL divergence, gradient & Cover-based & L2 Distance & Diversity - Clustering & & & \checkmark & &\\ \midrule

\cite{tang2023pld} & 2023 & MICCAI & Multiple Inferences - Model Disagreement & KL divergence & - & - & Top-k & & & & &\\ \midrule

\cite{qiu2023adaptive} & 2023 & MICCAI & Single Model & Distance to 0.5 & - & - & Top-k & & & & Patch &\\ \midrule

\cite{chen2023making} & 2023 & MIDL & - & - & \multicolumn{2}{c}{Loss of Self-supervised Pretext Tasks} & Top-k & & \checkmark& & &\\ \midrule

\cite{qu2023annotating} & 2023 & NeurIPS & \makecell[c]{Multiple Inferences - Model Disagreement\\} & \makecell[c]{Variance, Entropy,\\Overlap} & - & - & Top-K & & & & &\\ \midrule

\cite{lou2023which} & 2023 & TMI & - & - & Clustering-based & Consistency& Diversity - Clustering & Pseudo-label & & & & \checkmark \\ \midrule

\cite{du2022contrastive} & 2023 & TPAMI & - & - & Discrepancy-based & Semantic and distinctive scores& Hybrid - Fusion& & \checkmark& & &\\ \midrule

\cite{wan2023multiple} & 2023 & TPAMI & Adversarial Training & Disagreement of Classifiers & - & - & Top-k & & & & & \\ 
\bottomrule
\end{tabular}}
\end{varwidth}}
\end{table*}

\subsection{Semi-supervised Learning: Utilizing Unlabeled Data}
\label{semi}

Semi-supervised learning \citep{chen2022semi, han2024deep} aims to boost performance by utilizing unlabeled data upon supervised training.
{The need for tedious human annotation can be further reduced by integrating AL and semi-supervised learning in medical image analysis.
The reason is that AL and semi-supervised learning complement each other.
Specifically, a large pool of unlabeled images should be collected from the hospital information systems to train a DL model for some clinical applications.
With the help of AL, the DL model is trained on an optimal labeled dataset constructed with a certain AL method, which reduces the annotation workload for doctors.
However, massive unlabeled samples sit idle during the model training in the AL cycle. 
By combining the AL with semi-supervised learning, the model can be trained on both labeled and unlabeled samples. \citep{jimenez2023computational}}
This section will introduce the integration of AL and semi-supervised learning from the perspectives of pseudo-labeling and consistency regularization.

\subsubsection{Pseudo-Labeling}

Pseudo-labeling \citep{lee2013pseudo} is one of the most straightforward methods in semi-supervised learning. 
It uses the model's predictions of unlabeled data as pseudo-labels and combines them with labeled data for supervised training. 
Although it's possible to assign pseudo-labels to all unlabeled samples for training, it could introduce noise. 
To mitigate this, \cite{wang2017costeffective} proposed cost-effective active learning (CEAL), integrating pseudo-labeling with uncertainty-based AL. 
Specifically, CEAL sent the most uncertain samples for expert annotation and assigned pseudo-labels to the most confident samples. 
Many subsequent works have built upon the ideas of CEAL.
{
\cite{gorriz2017cost} adopted the CEAL framework in the melanoma segmentation and used the MC dropout for uncertainty estimation.
}
In medical image segmentation, \cite{zhao2021dsal} refined the pseudo-labels with dense conditional random fields.  
{
Additionally, \cite{li2022pathal} proposed a new approach for selecting samples for oracle annotation and pseudo-labeling in the Gleason grading of prostate cancer with histopathology images.
}
They employed curriculum learning to categorize all samples into hard and easy.
Hard samples were all sent for oracle annotation.
For the easy samples, they evaluated the presence of label noise based on the training loss.
Easy samples with low training loss were used for pseudo-labels to assist training, whereas easy samples with high loss were considered noisy and excluded from training.

\subsubsection{Consistency Regularization}

Consistency regularization aims to enforce similar outputs under perturbations of input data or model parameters.
Maximizing consistency serves as an unsupervised loss for unlabeled samples, which helps improve the robustness, reduces overfitting, and improves model performance. 
Many works integrated existing consistency-based semi-supervised methods into the training process of AL.
{In chest X-ray classification, \cite{balaram2022consistencybased} incorporated several semi-supervised methods with AL to further reduce annotation costs, including MeanTeacher \citep{tarvainen2017mean}, VAT \citep{miyato2018virtual} and NoTeacher \citep{unnikrishnan2021semi}.}
\cite{huang2021semisupervised} combined their proposed COD with MeanTeacher \citep{tarvainen2017mean}, demonstrating superior performance. 
\cite{wang2022unsupervised} combined density-based AL with different existing semi-supervised methods. 
Results showed that the proposed method outperforms other active learning methods and excels in semi-supervised learning.

Consistency could be also used for sample selection.
\cite{gao2020consistencybased} introduced a semi-supervised active learning framework.
Consistency here was used for both semi-supervised training and evaluating informativeness.
In this framework, samples are fed into the model multiple times with random augmentations. 
The consistency loss of unlabeled samples was implemented by minimizing the variance between multiple outputs. 
They further selected less consistent samples for annotation. 
Results showed that combining AL with semi-supervised learning significantly improves performance.

Besides, \cite{zhang2022boostmis} combined AL with both pseudo-labeling and consistency regularization. 
The unlabelled images first underwent both strong and weak data augmentations.
When the confidence level of the weakly augmented images exceeded a certain threshold, they used these samples for semi-supervised training.
Specifically, predictions of the weakly augmented images were assigned as pseudo-labels, and the outputs of the strongly augmented images were forced to be consistent with the pseudo-labels.
However, when the confidence level was lower than the threshold, they used these samples for AL.
A balanced uncertainty selector and an adversarial instability selector were used to select samples for oracle annotation. 
They validated the effectiveness of their proposed method in grading metastatic epidural spinal cord compression with MRI images.

\subsection{Self-supervised Learning: Utilizing Pre-trained Model}
\label{self}

Integrating semi-supervised learning with AL has achieved successful applications.
However, its effectiveness is constrained by the dataset size.
This limitation is particularly evident for medical imaging datasets which are relatively small.
{
In clinical practice, plenty of raw medical images are stored in hospital information systems without human annotation. 
Self-supervised learning \citep{krishnan2022self} could be a vital tool for mining information hidden in those raw images.
}
Its idea is to train the model with the supervision of the data itself, thus allowing pre-training on a large unlabeled dataset. 
{Many studies have shown self-supervised pre-trained models could achieve impressive performance by finetuning on a few randomly selected labeled samples in medical image analysis \citep{azizi2021big, tang2022self}.
A natural expectation is to integrate active learning strategies with self-supervised learning, aiming for higher annotation efficiency over mere random sampling \citep{luth2024navigating}. 
}
Besides, these models could also act as a powerful feature extractor, which provides good initialization for AL. 
In this section, we will first introduce how self-supervised models solve the cold-start problem in AL and then explore different ways of integrating AL with self-supervised learning.

\subsubsection{Cold-start Problem in Active Learning}

Current AL methods usually require an initial labeled dataset to train the model for start and ensure reliable informativeness evaluation.
However, when the initial labeled set is small or even absent, the performance of these AL methods drops dramatically, sometimes even worse than random sampling \citep{chen2023making, hacohen2022active, yehuda2022active}.
{Studies also showed that simply integrating self-supervised learning with AL baselines leads to inferior performance than random sampling \citep{bengar2021reducing, xie2022general}. 
This phenomenon is known as the cold-start problem which commonly exists in AL of various domains, including medical image analysis \citep{liu2023colossal}. }
Tackling the cold-start problem is vital for improving the efficacy of AL, especially in the medical domains where annotation costs are extremely high. 
A key solution to the cold-start problem in AL is selecting the optimal set of initial labeled samples, which requires different strategies than the existing AL methods. 

Early attempts focused on utilizing the fully supervised pre-trained models to address the cold-start problem in AL. 
\cite{zhou2017finetuning} and their subsequent work \citep{zhou2021active} used ImageNet pre-trained models to select samples for annotation from completely unlabeled datasets {in medical image analysis}. 
They combined entropy and disagreement as informativeness metrics, where the disagreement was the KL divergence of prediction probabilities between different patches of the same sample. 
They also introduced randomness to balance exploration and exploitation.
{Experiments on two colonoscopy datasets and a CT pulmonary embolism detection dataset showed superior performance than other competitors.}

Self-supervised pre-trained models offer a good initialization for effectively tackling the cold-start problem in AL. 
ALPS \citep{yuan2020coldstart} was the first to introduce the cold-start problem in AL and employed self-supervised pre-trained models to address this issue. 
{
Based on a contrastive learning feature extractor, CALR \citep{jin2022coldstart} employed BIRCH clustering and chose the samples with maximum information density within each cluster for labeling.
Compared to k-Means, BIRCH clustering is less sensitive to outliers and can further identify noisy samples.}
TypiClust \citep{hacohen2022active} theoretically proved that querying typical samples is more beneficial for a low annotation budget. 
Therefore, based on self-supervised features, TypiClust selected samples from high-density areas of each k-Means cluster.
Beyond that, \cite{yehuda2022active} employed a graph-based greedy algorithm to select the optimal initial samples based on self-supervised features.
In CT segmentation, \cite{nath2022warm} proposed ProxyRank which designed new pretext tasks for self-supervised pre-training.
The model was trained to learn the threshold segmentation by an abdominal soft-tissue window.
Results indicated that the proposed method significantly outperforms random sampling in selecting initial samples. 
{
To benchmark the effectiveness of different cold-start AL methods in 3D medical image segmentation, \cite{liu2023colossal} reproduced ALPS, CALR, TypiClust, and ProxyRank on five MSD datasets \citep{antonelli2022medical}.
Results showed that TypiClust stands out from the four competitors.
However, no method consistently outperformed random selection on all five datasets, which calls for further exploration of cold-start AL in medical image analysis.
}

\subsubsection{Combination of Active Learning and Self-supervised Learning}

\noindent \textbf{Features:} 
The simplest way is leveraging the high-quality features of the self-supervised pre-trained models.
Many studies are based on a powerful self-supervised feature extractor \citep{pourahmadi2021simple, jin2022coldstart, hacohen2022active, yehuda2022active}.

\noindent \textbf{Pretext tasks} in self-supervised learning are designed to derive supervision directly from the data itself.
Solving these pretext tasks on large-scale unlabeled data, the model acquires useful feature representations that reflect data characteristics.
Different pretext tasks correspond to different pre-training paradigms, the typical ones including rotation prediction \citep{gidaris2018unsupervised}, contrastive learning \citep{he2020momentum}, and masked image modeling \citep{he2022masked}, etc.
Related works generally employed the loss of pretext task for AL. 
{In \cite{chen2023making}, the loss of contrastive learning was used to tackle cold-start problem for AL in medical image analysis.
They assumed that samples with higher losses are more representative of the data distribution. 
Specifically, they pre-trained on the target dataset with momentum contrastive learning \citep{he2020momentum}, and then used k-Means clustering to partition the unlabeled data into multiple clusters, selecting the samples with the highest contrastive loss within each cluster for annotation.
They then selected samples with the highest contrastive loss in each cluster for annotation.
The proposed method addressed the class imbalance caused by the bias of traditional AL methods, and the failure to detect anomalies when the number of initially labeled datasets was limited.
This method showed superior performance in PathMNIST, OrganMNIST, and BloodMNIST \citep{yang2023medmnist}.}
\cite{yi2022pt4al} found a strong correlation between the loss of pretext tasks and the loss of downstream tasks.
Thus, they initially focused on annotating samples with higher loss of pretext tasks and later shifted to those with lower loss.
Results showed that rotation prediction performed the best among different pretext tasks.

\noindent \textbf{Others:} 
Furthermore, recent works leverage self-supervised learning in other ways for AL. 
\cite{zhang2022onebit} introduced one-bit annotation into AL for classification tasks.
In this setting, oracles only returned whether the prediction was right or wrong rather than its specific class label.
Contrastive learning was adopted to pull the correct predictions closer and push away wrong predictions from their predicted classes. 
Results indicated that the proposed method outperforms other AL methods regarding bit information. 
\cite{du2021contrastive} integrated contrastive learning into AL to address the class distribution mismatch, where unlabeled data includes samples out of the class distribution of the labeled dataset.
In this work, contrastive learning was adopted to filter samples of mismatched classes and highlight sample informativeness by carefully setting negative samples.
Their extended work \cite{du2022contrastive} provided more theoretical analysis and experimental results and also integrated existing label information into the proposed framework.

\subsection{Region-based Active Learning: Smaller Labeling Unit}
\label{region}

Most AL works require the oracle to label the full image in medical image analysis.
However, labeling a full image can introduce redundancy in fine-grained tasks like segmentation or detection, resulting in an inefficient use of the annotation budget.
{For the example of abdomen multi-organ segmentation, large organs that are easy to segment (e.g., liver or spleen) do not need exhaustive annotation. 
Instead, those budgets would be better spent on small organs that are hard to segment, like the esophagus and adrenal glands.}
To address this issue, images could be divided into non-overlap regions for higher annotation efficiency, and experts can opt to annotate specific regions within an image, which is termed ``region-based active learning".
{This section introduces region-based active learning from the perspectives of patches and superpixels, which means that AL methods mentioned in this section selected either patches or superpixels within an image for annotation.}

\subsubsection{Patches}
Patches are most commonly used in region-based active learning, generally represented as square boxes. 
\cite{mackowiak2018cereals} combined uncertainty and annotation cost to select the informative patches for annotation.
{In retinal blood vessels segmentation of fundus images, \cite{xu2021partiallysupervised} selected patches with the highest uncertainty for annotation.
Furthermore, they utilized latent-space mixup to encourage linearization between labeled and unlabeled samples, thus leveraging unlabeled data to improve performance.}
\cite{casanova2020reinforced} employed deep reinforcement learning to automatically select informative patches for annotation. 
{
In grey matter and white matter segmentation of pathology images, \cite{lai2021joint} first split the whole slide image into multiple patches.
With the confidence (i.e., maximum predictive probability) of each patch, a mean filter of size 5x5 is used to aggregate the confidence of the neighboring patches.
As a result, one aggregated metric corresponded to a region of 5x5 patches, and regions with the highest uncertainty were selected for annotation.
Besides, \cite{qiu2023adaptive} adopted an adaptative region selection with non-square patches for whole-slide images.
Instead of sampling square patches, they dynamically determined the size of each non-square patch by carefully locating an informative area on each slide. 
The proposed method demonstrated improvement in annotation efficiency and robustness to AL hyperparameters compared to the square patch baseline.
}

\subsubsection{Superpixels}
Superpixels are also widely used in region-based active learning.
Superpixel-based AL initially pre-segments the images with superpixel generation algorithms based on color and texture \citep{achanta2012slic, van2012seeds}, and then calculates the informativeness of each superpixel.
The informativeness metric of each superpixel is the average of its constituent pixels.
\cite{siddiqui2020viewal} adopted uncertainty and disagreement between different viewpoints to select informative superpixels for annotation. 
{
In OCT segmentation, \cite{kadir2023edgeal} proposed edge-based entropy and divergence to select highly uncertain superpixels for annotation.
Experiments on three datasets were conducted to illustrate the effectiveness of their method.
}
For superpixel selection, \cite{cai2021revisiting} proposed dominant labeling which is the majority class label of all pixels in the superpixel.
They assigned the dominant labeling to every pixel within a superpixel, thus eliminating the need for detailed delineation.
They further introduced a class-balanced sampling strategy to better select superpixels containing minority classes.
Results showed that dominant labeling with superpixels significantly outperforms precise labeling with patches under the same number of labeling clicks. 
As a follow-up work, \cite{kim2023adaptive} proposed to adaptively merge and split spatially adjacent, similar, and complex superpixels, respectively. 
This approach yielded better performance than \cite{cai2021revisiting} with dominant labeling.
{
\cite{li2023hal} utilized the superpixels to estimate the regional consistency which is the difference between the prediction and the dominant class of each superpixel.
Combining other metrics like entropy and diversity, they selected the most uncertain foreground and background superpixel to reduce the annotation cost.
}

\subsection{Generative Model: Data Augmentation and Generative Active Learning}
\label{generative}

In recent years, the advancement of deep generative models enabled high-quality generation and flexible conditional generation. 
For example, a trained model could generate the corresponding lung X-ray scan when conditioned on a lung mask.
By integrating generative models, we can further improve the annotation efficiency of AL. 
In this section, we discuss how AL can be combined with generative models from two aspects: data augmentation and generative active learning.

\subsubsection{Synthetic Samples as Data Augmentation}

The simplest approach considers the synthetic sample produced by generative models as advanced data augmentation. 
These methods utilize label-conditioned generative models. 
As a result, it's guaranteed that all synthetic samples are correctly labeled since specifying the labels is a prerequisite for data generation. 
This method enables us to acquire more labeled samples without any additional annotations.
\cite{tran2019bayesian} argued that most synthetic samples produced by generative models are not highly informative. 
Therefore, they first adopted the BALD uncertainty to select samples for annotation, then trained a VAE-ACGAN on these labeled data to generate more informative synthetic samples. 
\cite{mahapatra2018efficient} used conditional GANs to generate chest X-rays with varying diseases to augment the labeled dataset.
Then, MC Dropout was used to select and annotate highly uncertain samples.
With the help of AL and synthetic samples, they achieved performance near fully supervised using only 35\% of the data.
Training conditional generative models requires a large amount of labeled data, while the labeled dataset in AL is often relatively small. 
To address this issue, \cite{lou2023which} proposed a conditional SinGAN \citep{shaham2019singan} that only requires one pair of images and masks for training.
The SinGAN improved the annotation efficiency for nuclei segmentation. 
\cite{chen2022when} integrated implicit semantic data augmentation (ISDA) \citep{wang2021regularizing} into AL. 
They initially used ISDA to augment unlabeled samples, then selected samples with large diversity between different data augmentations for annotation.
The model is trained on both the original data and its augmentations.
{
\cite{mahapatra2024gandalf} trained a VAE for synthesizing informative and non-redundant samples. 
These samples are generated by first sampling in the latent space of VAE and feeding them to the VAE decoder.
Besides, scores of label preservation and redundancy avoidance were adopted to pick the most informative synthetic samples.
The proposed method was tested in chest X-ray classification and multiple toy datasets from MedMNIST \citep{yang2023medmnist}.
}   

\subsubsection{Generative Active Learning}
Generative active learning selects synthetic samples produced by generative models for oracle annotation, thus without requiring a large unlabeled sample pool. 
The advantage of this approach lies in its ability to continuously search the data manifold through generative models.
It's worth noting that works in this section follow the setting of membership query synthesis, while works in the last section follow the setting of pool-based active learning.
This distinction arises because generative models in the last section were solely utilized to augment existing labeled datasets.
\cite{zhu2017generative} attempted to generate uncertain samples with GAN for expert annotation. 
Unfortunately, the quality of the generated samples was low and included many samples with indistinguishable classes.
Since experts find it difficult to annotate low-quality synthetic samples, alternative methods are needed to annotate these samples. 
\cite{chen2021active} first trained a bidirectional GAN to learn the data manifold. 
They then selected uncertain areas in the feature space and generated images within these regions using bidirectional GAN.
Finally, they used physics-based simulation to provide labels for the generated samples. 
In calcification level prediction in aortic stenosis of CT, they improved annotation efficiency by up to 10 times compared to random generation.

\subsection{Active Domain Adaptation: Tackling Distribution Shift}
\label{ada}

Domain Adaptation (DA) \citep{guan2021domain} has wide applications in medical image analysis. 
It aims to transfer knowledge from the source to the target domain, thus minimizing annotation costs.
Currently, the most common setting of DA is unsupervised domain adaptation (UDA), in which the source domain is labeled while the target domain is unlabeled.
{For the example of abdominal multi-organ segmentation, we can train a domain-adaptive segmentation model with labeled MR images alongside unlabeled CT images to achieve good performance on the CT domain \citep{liu2023structure}.}
However, the performance of UDA still lags behind fully supervised learning in the target domain. 
Selecting and annotating informative samples would be beneficial to bridge this gap. 
This setting is known as active domain adaptation (ADA).
For better queries in ADA, one should consider both uncertainty and representativeness regarding the target domain.
The latter is commonly referred to as domainness or targetness in ADA.
This section reviews the image-wise and region-wise ADA.

\subsubsection{Image-wise Active Domain Adaptation}
\label{imageADA}

{In this section, ADA methods performed an image-level selection, which involves most of the ADA works. }
\cite{su2020active} was the first to introduce the concept of ADA and combined domain adversarial learning with AL. 
Through a domain discriminator and task model, they performed importance sampling to select target domain samples that are uncertain and highly different from the source domain.
\cite{fu2021transferable} combined query-by-committee, uncertainty, and domainness in ADA.
They adopted a domain discriminator to select samples with high domainness and employed Gaussian kernels to filter out anomalous and source-similar samples of the target domain. 
Random sampling was also used to improve diversity.
\cite{prabhu2021active} performed k-Means clustering on target domain samples and selected cluster centers for annotation.
The cluster centers were weighted by uncertainty, thus ensuring that selected samples were uncertain and diverse.
In segmentation tasks, \cite{ning2021multianchor} introduced the idea of anchors in ADA.
They concatenated features of different classes from the source domain images. 
Cluster centers of these concatenations were referred to as anchors.
They then computed the distance between each target sample and its nearest anchor.
Target samples with the highest distance were requested for annotation.
\cite{xie2022active} introduced the concept of energy \citep{lecun2006tutorial} into ADA. 
The energy is inversely proportional to the likelihood of the data distribution.
In this work, the model trained on the source domain was used to calculate the energy of target domain samples. 
Samples with high energy were selected for annotation, which suggested they are representative of the target domain and substantially different from the source data. 
\cite{huang2023divide} selected samples with high uncertainty and prediction inconsistency to their nearest prototypes.
{
In the context of medical image analysis, \cite{chen2023think} tackled domain shifts in the setting of federated active learning. 
They proposed an EDL-based framework with a global model across all clients and local models for each client.
In this work, the EU is related to domain shifts between the global model and local data.
Therefore, the AUs of the global and local models were calibrated by the EU, thus improving performance.
Results on multiple medical imaging datasets showed its effectiveness in reducing annotation costs.
In nasopharyngeal carcinoma tumor segmentation, \cite{wang2023dual} proposed a source-domain and target-domain dual-reference strategy to select informative samples for annotation.
Specifically, the features of the source samples were clustered and the cluster centers were reference samples. 
Target samples with the highest and lowest similarity to the references are selected for annotation, which were treated as domain-invariant and domain-specific samples, respectively.
}

\subsubsection{Region-wise Active Domain Adaptation}
\label{regionADA}

{To better utilize the annotation budget, some ADA works also selected patches or superpixels within an image for annotation.}
\cite{shin2021labor} proposed LabOR, which first used a UDA pre-trained model to generate pseudo-labels for target samples, which was used to train two segmentation heads. 
They maximized the disagreements between the two heads and annotated regions that exhibited the most disagreement.
LabOR achieved performance close to full supervision with only 2.2\% of target domain annotations. 
In \cite{xie2022fewer}, uncertainty and regional impurity were used to select and annotate the most informative patches.
Regional impurity measured the number of unique predicted classes within the neighborhood of a pixel, which presents the edge information.
They used extremely small patches (e.g., size of 3x3) for annotation and achieved performance close to full supervision with only 5\% of the annotation cost. 
\cite{wu2022d2ada} proposed a density-based method to select the most representative superpixels in the target domain for annotation. 
They employed Gaussian mixture models (GMM) as density estimators for superpixels in both the source and target domains, aiming to select those with high density in the target domain and low density in the source domain.

\section{Active Learning for Medical Image Analysis} 
\label{al4mia}
Due to the potential of significantly reducing annotation costs, AL is receiving increasing attention in medical image analysis.
The unique traits of medical imaging require us to design specialized AL methods.
Building on the foundation of the previous two sections, this section will focus on introducing AL works tailored to medical image analysis across different tasks, including classification, segmentation, and reconstruction.

Additionally, in Table \ref{tab_clinical}, we list all the AL works related to medical image analysis in this survey, providing the name of the used dataset, its modality, ROIs, and corresponding clinical and technical tasks.

\begin{table*}[!p]
\centering
\rotatebox[origin=c]{90}{%
\tabcolsep=3pt
\begin{varwidth}{0.85\textheight}
\centering
\caption{Surveyed Works of Active Learning related to Medical Image Analysis. ``-'' stands for such information is not available for not provided by the authors or not in the case.}
\label{tab_clinical}\footnotesize
\resizebox{\textwidth}{!}{%
\begin{tabular}{cccccccccccc}

\toprule
& Year& Venues& Modality & ROIs & Dataset & {Sampling Unit} & { \makecell[c]{Size\\(Train/Val/Test)\\or (Train+Val/Test)}} & {Initial Pool Size} & {Budget per Round} & \multicolumn{1}{c}{Clinical Task} & Technical Task \\
\midrule


\cite{gorriz2017cost} & 2017& arXiv & Dermscopy & Skin & ISIC 2017 & image & 2,000 (1,600/400) & 600 & 35 & Skin Cancer Diagnosis & Classification\\ \midrule 
\multirow{3}{*}{\cite{zhou2017finetuning}} & \multirow{3}{*}{2017} & \multirow{3}{*}{CVPR} & Colonoscopy & Colon& in-house & image & 4,000 (2,000/2,000) & \multirow{3}{*}{-} & \multirow{3}{*}{-} & Image Quality Assessment & Classification \\
& & & Colonoscopy & Colon& in-house & image & 28,250 (16,300/11,950) & & & Polyp Detection & Classification \\
& & & CT & Lung & in-house & PE candidate & 6,255 (3,840/2,415) & & & Pulmonary Embolism Detection& Classification \\ \midrule 

\cite{gal2017deep} & 2017& ICML & Dermscopy & Skin & ISIC 2016 & image & 400 (200/200) & 100 & 20 & Skin Cancer Diagnosis & Classification\\ \midrule 

\multirow{2}{*}{\cite{yang2017suggestive}} & \multirow{2}{*}{2017}& \multirow{2}{*}{MICCAI}& Histopathology & Colon & GlaS & patch & 165 images (80/5/80) & \multirow{2}{*}{8} & \multirow{2}{*}{8} & Gland Segmentation & Segmentation \\
 & & & Ultrasound & Lymoh Node& in-house & image & 74 (37/37) & & & Lymoh Node Segmentation & Segmentation \\ \midrule 

\cite{otalora2017training}& 2017& MICCAIW & Fundus & Eye & e-ophtha & patch & 20,148 (17,520/656/1,972) & 160 & 32 & Exudate Classification & Classification\\ \midrule 

\cite{beluch2018power} & 2018& CVPR & Fundus & Eye& EyePacs & image & 88,702 (67,961/3,000/17,741) & 1,000 & 5,000 & Diabetic Retinopathy Detection& Classification\\ \midrule 

\cite{xu2018quantization} & 2018 & CVPR & Histopathology & Colon& GlaS & patch & 165 images (80/5/80) & 8 & 8 & Gland Segmentation & Segmentation \\ \midrule 

\cite{sourati2018active} & 2018 & DLMIA & \multirow{2}{*}{MRI} & \multirow{2}{*}{Brain} & dHCP & \multirow{2}{*}{patch} & 66 patients & \multirow{2}{*}{3 patinets} & \multirow{2}{*}{50} & \multirow{2}{*}{Brain Extraction} & \multirow{2}{*}{Segmentation} \\ 
\cite{sourati2019intelligent} & 2019 & TMI & & & in-house & & 25 patients & & & & \\ \midrule 

\cite{kuo2018costsensitive} & 2018& MICCAI& CT & Head & in-house & \makecell[c]{region, image,\\volume} & 1,247 volumes (934/313) & 1/32 & double & Intracranial Hemorrhage Detection & Segmentation \\ \midrule 

\multirow{2}{*}{\cite{mahapatra2018efficient}} & \multirow{2}{*}{2018} & \multirow{2}{*}{MICCAI} & \multirow{2}{*}{X-ray}& \multirow{2}{*}{Chest} & \multirow{2}{*}{SCR \& Chestx-ray8} & \multirow{2}{*}{image} & \multirow{2}{*}{647 (247/400)} & \multirow{2}{*}{10\%} & \multirow{2}{*}{5\%} & Lung Segmentation & Segmentation \\
 & & & & & & & & & & Thoracic Disease Diagnosis & Classification\\ \midrule 

\multirow{3}{*}{\cite{zheng2019biomedical}} & \multirow{3}{*}{2019}& \multirow{3}{*}{AAAI} & Histopathology & Colon& GlaS & \multirow{2}{*}{patch} & 1,530 & \multirow{3}{*}{0} & \multirow{2}{*}{30\% \& 50\%} & Gland Segmentation& Segmentation \\
 & & & Electron Microscopy& Fungus & in-house &  & 784 & & & Fungus Segmentation & Segmentation \\ 
 & & & MRI & Heart& HVSMR 2016 & slice & 20 volumes (10/10) & & \makecell[c]{every 2, 10, 20,40, 80 slices} & Whole-heart Segmentation & Segmentation \\\midrule 

\cite{qi2019labelefficient} & 2019& JBHI & Histopathology & Breast & BreaKHis & image & \multicolumn{3}{c}{See paper for details} & Breast Cancer Diagnosis & Classification\\ \midrule 

\cite{sadafi2019multiclass} & 2019& MICCAI& Mircoscopy & Blood& in-house & image & 208 (188/20) & 30 & - & Red Blood Cell Detection & Object Detection\\ \midrule 

\cite{shi2019active} & 2019& MICCAWI& Dermscopy & Skin & ISIC 2017 & image & 4,332 (3,582/150/600) & 10\% & 10\% & Skin Lesion Diagnosis & Classification\\ \midrule 

\multirow{2}{*}{\cite{gu2018reliable}} & \multirow{2}{*}{2019} & \multirow{2}{*}{TBME} & Endomicroscopy & Breast & \cite{gu2017unsupervised} & \multirow{2}{*}{image} & 1,366 (1,042/324) & - & - & Endomicroscopy Mosaic Classification & \multirow{2}{*}{Classification} \\
 & & & Colonoscopy & Colon & \cite{ye2016online} & & 7,847 (3,947/3,900) & - & - & Gastrointestinal Image Classification &  \\ \midrule 
 
\multirow{2}{*}{\cite{zheng2020annotation}} & \multirow{2}{*}{2020}& \multirow{2}{*}{AAAI} & MRI & Heart& HVSMR 2016 & slice & 20 volumes (10/10) & 0 & \makecell[c]{every 5, 10, 20, 40, 80 slices} & Whole-heart Segmentation & Segmentation \\
 & & & Electron Microscopy& Mouse & \cite{lee2015recursive} & slice & 4 volumes (3/1) & 0 & \makecell[c]{every 4, 16, 64 slices} & Neuron Boundary Segmentation& Segmentation \\ \midrule 

\multirow{2}{*}{\cite{lin2020two}} & \multirow{2}{*}{2020} & \multirow{2}{*}{ECCV} & \multirow{2}{*}{Electron Microscopy} & Mouse Synapses  & \multirow{2}{*}{EM-R50} & \multirow{2}{*}{image} & 48.7K (28.7K/20K) & \multirow{2}{*}{-} & \multirow{2}{*}{1,280} & Synapse Detection & Segmentation \\ 
& & & & Mouse Synapses \& Mitochondria & & & 15K (10K/5K) & & & Mitochondria Segmentation & Segmentation \\ 
\midrule 

\multirow{2}{*}{\cite{dai2020suggestive}} & \multirow{2}{*}{2020} & \multirow{2}{*}{MICCAI} & \multirow{2}{*}{MRI} & \multirow{2}{*}{Brain} & \multirow{2}{*}{BraTS 2019} & volume & \multirow{2}{*}{335 patients (260/75)} & 10 & 10 & \multirow{2}{*}{Brain Tumor Segmentation} & \multirow{2}{*}{Segmentation} \\ 
& & & & & & slice & & 500 & 500 & & \\ \midrule 

\multirow{2}{*}{\cite{li2020attention}} & \multirow{2}{*}{2020}& \multirow{2}{*}{MICCAI}& Histopathology & Colon& GlaS & 2D patch & 27,200 & 10\% & 20\% & Gland Segmentation& Segmentation \\
 & & & MRI & Brain& iSeg & 3D patch & 16,380 & - & 50\% & Infant Brain Segmentation & Segmentation \\ \midrule 

\cite{liu2020deep} & 2020 & MICCA & CT & Lung & DeepLesion & volume & 1,281 (1,000/281) & 10\% & 10\% & Pulmonary Nodule Detection& Object Detection\\ \midrule 

\cite{shen2020deep} & 2020& MICCAI& Immunohistochemistry & Breast & in-house & image & 3,441 (2,767/306+368) & 10\% & 10\% & Breast Cancer Region Segmentation & Segmentation \\ \midrule 

\multirow{2}{*}{\cite{wang2020deep_RL}} & \multirow{2}{*}{2020}& \multirow{2}{*}{MICCAI}& CT & Lung & \href{https://tianchi.aliyun.com/competition/entrance/231724/introduction}{Tianchi} & image & 7K (3.5K/3.5K) & 5\% & about 7\% & Lung Diease Detection & Classification\\
 & & & Fundus & Eye& EyePacs & image & 4,460 (2,230/2,230) & 10\% & about 11\% & Diabetic Retinopathy Detection& Classification\\  \midrule 

\cite{hiasa2020automated} & 2020& TMI & CT & Hip \& Thigh & TCIA \& in-house & slice & 20 volumes & 5\% & 5\% & Muscle Segmentation & Segmentation \\ \midrule 

\cite{huang2020rectifying} & 2020& TMI & X-ray& Chest& in-house & region & 10,966 images (7:1:2) & 0\% & 5\% \& 10\% & Rib Fracture Recognition & Object Detection\\ \midrule 

\cite{shen2021labeling} & 2021& ISBI & MRI & Brain & BraTS2018 & slice & 285 volumes (233:52) & 5 volumes & 100 \& 200 & Brain Tumor Segmentation & Segmentation\\ \midrule 

\multirow{2}{*}{\cite{zhao2021dsal}} & \multirow{2}{*}{2021}& \multirow{2}{*}{JBHI} & Dermscopy & Skin & ISIC 2017 & image & 2,000 (1,600/400) & 600 & 100 & Skin Lesion Segmentation & Segmentation \\
& & & X-ray& Hand & RSNA Bone Age Dataset & image & 209 (139/20/50) & 10 & 10 & Finger Bone Segmentation & Segmentation \\ \midrule 

\multirow{2}{*}{\cite{ozdemir2021active}} & \multirow{2}{*}{2021} & \multirow{2}{*}{KBS} & \multirow{2}{*}{MRI} & \multirow{2}{*}{Lower Extremities} & \multirow{2}{*}{in-house} & slice & \multirow{2}{*}{36 volumes (25/2/9)} & 64 & 32 & \multirow{2}{*}{Musculoskeletal Segmentation} & \multirow{2}{*}{Segmentation} \\ 
& & & & & & volume & & 1 & 1 & & \\
\midrule 

\cite{wu2021covidal} & 2021& MedIA & CT & Lung & CC-CCII & volume & 962 (7:1:2) & - & 10 & COVID-19 Diagnosis& Classification\\ \midrule 

\multirow{3}{*}{\cite{zhou2021active}} & \multirow{3}{*}{2021} & \multirow{3}{*}{MedIA} & Colonoscopy & Colon& in-house & image & 4,000 (2,000/2,000) & \multirow{3}{*}{-} & \multirow{3}{*}{-} & Image Quality Assessment & Classification \\
& & & Colonoscopy & Colon& in-house & image & 28,250 (16,300/11,950) & & & Polyp Detection & Classification \\
& & & CT & Lung & in-house & PE candidate & 6,255 (3,840/2,415) & & & Pulmonary Embolism Detection& Classification \\ \midrule  

\multirow{4}{*}{\cite{wang2021annotationefficient}} & \multirow{4}{*}{2021}& \multirow{4}{*}{MICCAI}& \makecell[c]{Mircoscopy\\(Synthetic)} & Bacterial Cells & VGG Cell & \multirow{4}{*}{image} & 150 (50/100) & \multirow{4}{*}{-} & \multirow{4}{*}{10\%} & \multirow{4}{*}{Cell Counting} & \multirow{4}{*}{\makecell[c]{Keypoint\\Localization}} \\
 & & & Histopathology & Human Bone Marrow & MBM & & 29 (15/14) & & & & \\
 & & & Histopathology & Human Adipocyte Cells & ADI & & 150 (50/100) & & & & \\
 & & & - & Various Tissues \& Species & DCC & & 176 (100/76) & & & & \\ \midrule 

\multirow{2}{*}{\cite{xu2021partiallysupervised}} & \multirow{2}{*}{2021}& \multirow{2}{*}{MICCAI}& Fundus & Eye& DRIVE & \multirow{2}{*}{region} & 40 images (20/20) & \multirow{2}{*}{-} & \multirow{2}{*}{\makecell{10\% patches\\in selected images}} & \multirow{2}{*}{Retina Vessel Segmentation} & \multirow{2}{*}{Segmentation} \\
 & & & OCTA& Eye& ROSE-1 & & 117 images & & & &  \\ 
\bottomrule
\end{tabular}}
\end{varwidth}}
\end{table*}

\begin{table*}[!p]
\centering
\setcounter{table}{2}
\rotatebox[origin=c]{90}{%
\tabcolsep=3pt
\begin{varwidth}{0.85\textheight}
\centering
\caption{Methodology summarization of surveyed active learning works.``-'' stands for such information is not available for not provided by the authors or not in the case.}
\label{tab_clinical}\footnotesize
\resizebox{\textwidth}{!}{%
\begin{tabular}{cccccccccccc}

\toprule
  & Year& Venues& Modality & ROIs & Dataset & {Sampling Unit} & { \makecell[c]{Size\\(Train/Val/Test)\\or (Train+Val/Test)}} & {Initial Pool Size} & {Budget per Round} & \multicolumn{1}{c}{Clinical Task} & Technical Task \\
\midrule

\multirow{3}{*}{\cite{zhou2021qualityaware}} & \multirow{3}{*}{2021}& \multirow{3}{*}{MICCAI}& \multirow{3}{*}{CT} & Lung & \multirow{2}{*}{MSD} & \multirow{3}{*}{\makecell[c]{scribble,\\bounding box,\\extreme point}} & 96 volumes (64/32) & \multirow{3}{*}{-} & \multirow{3}{*}{-} & \multirow{2}{*}{Tumor Segmentation} & \multirow{3}{*}{Segmentation} \\
 & & & & Colon & & & 190 volumes (126/64) & & & & \\
 & & & & Kidney & KiTS 19 & & 210 volumes (168/42) & & & Kidney \& Tumor Segmentation & \\ \midrule

\cite{chong2021evaluation} & 2021& MICCAIW & CT & Chest & \citep{rahman2021exploring} & image & 15,153 (8:1:1) & 10\% & 10\% & COVID-19 Diagnosis & Classification\\ \midrule 
 
\multirow{3}{*}{\cite{nguyen2022goal}} & \multirow{3}{*}{2021}& \multirow{3}{*}{MIDL} & \multirow{3}{*}{X-ray}& Chest& in-house & \multirow{3}{*}{image} & 135,309 (131,030/4,279) & \multirow{3}{*}{6,550} & \multirow{3}{*}{$\leq$6,550} & \makecell[c]{Diagnosis of Airspace Opacity\\ \& Lung Lesion} & Classification\\
& & & & Chest & RSNA Pneumonia & & 26,684 (24,015/2,669) & & & Diagnosis of Pneumonia & Classification\\
& & & & Chest & CheXpert & & 228,595 (224,316/4,279) & & & Detection of Pleural Effusion & Classification\\ \midrule 

\multirow{2}{*}{\cite{mahapatra2021interpretabilitydriven}} & \multirow{2}{*}{2021}& \multirow{2}{*}{TMI} & X-ray& Chest& ChestX-ray8 & image & 112,120 (7:1:2) & \multirow{2}{*}{10\%} & \multirow{2}{*}{10\%} & Thoracic Disease Diagnosis& Classification\\
 & & & Histopathology & Colon& GlaS & patch & 165 images & & & Gland Segmentation& Segmentation \\ \midrule 

\multirow{2}{*}{\cite{nath2021diminishing}} & \multirow{2}{*}{2021}& \multirow{2}{*}{TMI} & CT & Pancreas & \multirow{2}{*}{MSD} & \multirow{2}{*}{volume} & 281 (221/30/30) & 20 & 5 & Pancreas \& Tumor Segmentation & Segmentation \\
 & & & MRI & Hippocampus & & & 263 (163/50/50) & 10 & 1 & Hippocampus Segmentation & Segmentation \\ \midrule 

\cite{chen2021active} & 2021& TPAMI & CT & Heart& in-house & volume & 168 (126/42) & - & - & \makecell[c]{Calcification Level Prediction\\in Aortic Stenosis}& Classification\\ \midrule 


\cite{kothawade2022prism} & 2022& AAAI & X-ray & Chest& PneumoniaMNIST & image & 5,856	(4,708/524/624) & - & - & Pneumonia \& Normal Classification& Classification\\ \midrule 

\cite{wang2022boosting} & 2022& AAAI & \makecell[c]{cyro-ET \\(simulated)} & - & SHREC’19 & image & 25,000 (24,000/1,000) & 10\% & 5\% & Subtomogram Classification & Classification\\ \midrule 

\multirow{2}{*}{\cite{quan2022which}} & \multirow{2}{*}{2022}& \multirow{2}{*}{CVPR} & \multirow{2}{*}{X-ray} & Head & \href{https://www.kaggle.com/datasets/jiahongqian/cephalometric-landmarks}{Kaggle} & \multirow{2}{*}{image} & 400 (150/250) & \multirow{2}{*}{-} & \multirow{2}{*}{-} & Cephalometric Landmark Detection & Keypoint Localization \\
 & & & & Hand & \cite{payer2019integrating} & & 909 (609/300) & & & Hand Landmark Detection & Keypoint Localization \\ \midrule 

\cite{zhang2022boostmis} & 2022 & CVPR & MRI & Spine & in-house & image & 7,295 (7:1:2) & 10\% & 5\% & \makecell[c]{Diagnosis of Metastatic Epidural\\Spinal Cord Compression} & Classification\\ \midrule 

\cite{jin2022deep} & 2022& \makecell[c]{Knowledge-based\\Systems} & Dermscopy & Skin & ISIC 2020 & image & 33,126 (8:1:1) & 0 & 10\%, 30\% \& 50\%  & Skin Lesion Classification & Classification\\ \midrule 

\multirow{2}{*}{\cite{jin2022oneshot}} & \multirow{2}{*}{2022}& \multirow{2}{*}{\makecell[c]{Knowledge-based\\Systems}} & Dermscopy & Skin & ISIC 2018 & image & 2,594 (3:1:1) & 0 & 40\%-60\% with step of 2.5\% & Skin Lesion Segmentation & Segmentation \\
 & & & X-ray & Chest & \cite{jaeger2013automatic} & image  & 707 (3:1:1) & 0 & 30\%-50\% with step of 2.5\% & Lung Segmentation & Segmentation \\ \midrule 

\multirow{2}{*}{\cite{atzeni2022deep}} & \multirow{2}{*}{2022}& \multirow{2}{*}{MedIA} & MRI & Brain& SATA & \multirow{2}{*}{boundary pixel} & 35 volumes & \multirow{2}{*}{-} & \multirow{2}{*}{-} & Brain Structure Segmentation & Segmentation \\
 & & & Histology & Brain& in-house & & 15 stacks & & & Brain Structure Segmentation& Segmentation \\ \midrule 

\multirow{2}{*}{\cite{dai2022suggestive}} & \multirow{2}{*}{2022} & \multirow{2}{*}{MedIA} & \multirow{2}{*}{MRI} & \multirow{2}{*}{Brain} & BraTS 2019 & image & 335 patients (260/75) & 0.5\% & 1\% & Brain Tumor Segmentation & \multirow{2}{*}{Segmentation} \\ 
& & & & & MALC & image & 30 patients (20/10) & 6\% & 6\% & Brain Structure Segmentation & \\ \midrule 

 \multirow{4}{*}{\cite{zhou2022volumetric}} & \multirow{4}{*}{2022}& \multirow{4}{*}{MedIA}& \multirow{3}{*}{CT} & Lung & \multirow{2}{*}{MSD} & \multirow{4}{*}{\makecell[c]{scribble,\\bounding box,\\extreme point}} & 96 volumes (64/32) & \multirow{4}{*}{-} & \multirow{4}{*}{-} & \multirow{2}{*}{Tumor Segmentation} & \multirow{4}{*}{Segmentation} \\
 & & & & Colon & & & 190 volumes (126/64) & & & & \\
 & & & & Kidney & KiTS 19 & & 210 volumes (168/42) & & & Kidney \& Tumor Segmentation & \\ 
 & & & Colonoscopy & Colon & CVC-ClinicDB & & 29 sequences (23/3/3) & & & Polyp Segmentation &  \\ \midrule

\multirow{2}{*}{\cite{nath2022warm}} & \multirow{2}{*}{2022}& \multirow{2}{*}{MICCAI}& \multirow{2}{*}{CT} & Liver& \multirow{2}{*}{MSD} & \multirow{2}{*}{volume} & 131 (105/26) & \multirow{2}{*}{0} & 5\% & Liver \& Tumor Segmentation& \multirow{2}{*}{Segmentation} \\
& & & & Hepatic Vessels & & & 303 (242/61) & & 2\% & Hepatic Vessels \& Tumor Segmentation &  \\ \midrule

\cite{bai2022discrepancybased} & 2022 & MICCAI& \makecell[c]{Wireless Capsule\\Endoscopy} & Colon& CAD-CAP & image & 1,812 (4:1) & 0 & 10\% & Polyp Segmentation& Segmentation \\ \midrule 

\multirow{2}{*}{\cite{balaram2022consistencybased}} & \multirow{2}{*}{2022} & \multirow{2}{*}{MICCAI}& \multirow{2}{*}{X-ray}& \multirow{2}{*}{Chest}& \multirow{2}{*}{Chestx-ray8} & \multirow{2}{*}{image} & \multirow{2}{*}{112,120 (7:1:2)} & 2\% & 0.5\% & \multirow{2}{*}{Thoracic Disease Diagnosis}& \multirow{2}{*}{Classification}\\ 
 & & & & & & & & 5\% & 1\% & & \\ 
\midrule 

\multirow{5}{*}{\cite{wu2022selflearning}} & \multirow{5}{*}{2022}& \multirow{5}{*}{MICCAI}& CT & \multirow{5}{*}{Liver}& LiTS & \multirow{5}{*}{slice} & 130 (train) & \multirow{5}{*}{-} & \multirow{5}{*}{-} & \multirow{5}{*}{Liver \& Tumor Segmentation}& \multirow{5}{*}{Segmentation} \\
& & & CT, MRI & & CHAOS & & \makecell[c]{CT: 20(train)\\MRI: 120 (60/60)} & & & &  \\
& & & CT & & Sliver07 & & 20 (test) & & & &  \\
& & & CT & & MSD & & 131 (test) & & & &  \\\midrule 

\multirow{5}{*}{\cite{kothawade2022clinical}} & \multirow{5}{*}{2022}& \multirow{5}{*}{MICCAIW}& X-ray& Chest& PneumoniaMNIST & \multirow{5}{*}{image} & 5,856	(4,708/524/624) & 100 & 10 & Pneumonia \& Normal Classification & \multirow{5}{*}{Classification}\\
& & & Histopathology & Colon & PathMNIST & & 107,180 (89,996/10,004/7,180) & 50 & 500 & Survival Prediction & \\
& & & Mircoscopy & Peripheral Blood& BloodMNIST & & 17,092 (11,959/1,712/3,421) & 228 & 20 & Cell Type Classification & \\
& & & Dermscopy & Skin & ISIC 2018 & & 2,594 & 1947 & 20 & Skin Lesion Diagnosis & \\
& & & Fundus & Eye& APTOS-2019 & & - & 673 & 40 & Diabetic Retinopathy Detection & \\ \midrule 

\cite{aklilu2022alges} & 2022 & MLHC & Laparoscopy & Gallbladder & CholecSeg8k & image & 8,080 (4,640/1,600/1,640) & 10 & 10\% & Segmentation of Laparoscopic Surgical Images & Segmentation \\ \midrule 

\cite{bernhardt2022active} & 2022& \makecell[c]{Nature\\Communications} & X-Ray & Chest & NoisyCXR & image & 26.6K & - & - & Thoracic Disease Diagnosis & Classification\\ \midrule

\cite{li2022pathal} & 2022& TMI & Histopathology & Prostate & PANDA & image & 11,000 & 10\% & 10\% & Gleason Grading of Prostate Cancer& Classification\\ \midrule 

\multirow{4}{*}{\cite{wu2022federated}} & \multirow{4}{*}{2022} & \multirow{4}{*}{TMI} & \multirow{2}{*}{CT} & \multirow{2}{*}{Lung} & \multirow{2}{*}{CC-CCII} & \multirow{2}{*}{slice} & 750 & \multirow{4}{*}{20\%} & \multirow{4}{*}{5\%} & Lung Lesions Segmentation & Segmentation \\ 
 & & & & & & & 108,676 & & & COVID-19 and Penunomia Diagnosis & Classification \\ 
& & & \multirow{2}{*}{Colonoscopy} & \multirow{2}{*}{Colon} & \multirow{2}{*}{HyperKvasir} & \multirow{2}{*}{image} & 1000 & & & Polyp Segmentation & Segmentation \\ 
& & & & & & & 2,717 & & & Colonoscopic Lesion Classification & Classification \\ 

\midrule 

\cite{mahapatra2022graph} & 2022& TMI & X-ray& Chest& ChestXpert & image & 65,240 patients (7:1:2) & 10\% & 10\% & Thoracic Disease Diagnosis& Classification\\ 

\bottomrule
\end{tabular}}
\end{varwidth}}
\end{table*}

\begin{table*}[!p]
\centering
\setcounter{table}{2}
\rotatebox[origin=c]{90}{%
\tabcolsep=3pt
\begin{varwidth}{0.85\textheight}
\centering
\caption{Methodology summarization of surveyed active learning works. ``-'' stands for such information is not available for not provided by the authors or not in the case.}
\label{tab_clinical}\footnotesize
\resizebox{\textwidth}{!}{%
\begin{tabular}{cccccccccccc}

\toprule
  & Year& Venues& Modality & ROIs & Dataset & {Sampling Unit} & { \makecell[c]{Size\\(Train/Val/Test)\\or (Train+Val/Test)}} & {Initial Pool Size} & {Budget per Round} & \multicolumn{1}{c}{Clinical Task} & Technical Task \\
\midrule


\multirow{2}{*}{\cite{khanal2023m}} & \multirow{2}{*}{2023} & \multirow{2}{*}{arXiv} & MRI & Brain & BraTS 2018 & slice & 5,846 (3,673/1,009/1,164) & 200 & 100 & Brain Tumor Analysis & \makecell[c]{Classification \&\\Segmentation} \\
 &  & & X-Ray & Chest & COVID-QU-Ex & image & 5,826 (3,728/932/1,166) & 100 & 100 & COVID-19 Diagnosis & Classification\\ \midrule

\multirow{5}{*}{\cite{chen2023think}} & \multirow{5}{*}{2023} & \multirow{5}{*}{arXiv} & Dermscopy & Skin & Fed-ISIC & \multirow{5}{*}{image} & 21,989 (17,591/4,398) & \multirow{2}{*}{500} & \multirow{2}{*}{500} & Skin Lesion Diagnosis & \multirow{2}{*}{Classification} \\ 
& & & Dermscopy & Histopathology & Fed-Camelyon & & 455,954 (364,761/91,193) & & & \makecell[c]{Detection of Cancer\\Metastasesin Lymph Nodes} & \\
& & & Colonoscopy & Colon & Fed-Polyp & & 2,187 (1,751/436) & 50 & 50 & Polyp Segmentation & \multirow{3}{*}{Segmentation} \\ 
& & & MRI & Prostate & Fed-Prostate & & 1,867 (1,541/326) & \multirow{2}{*}{20} & \multirow{2}{*}{20} & Prostate Segmentation & \\ 
& & & Fundus & Eye & Fed-Fundus & & 1,060 (849/211) & & & Retina Vessel Segmentation & \\ 
\midrule 

\cite{wang2023dual} & 2023 & arXiv & MRI & Nose & in-house & slice & 1,057 patients (7:1:2) & - & 20\% & \makecell[c]{Nasopharyngeal Carcinoma\\Tumor Segmentation} & Segmentation\\ \midrule 

\multirow{2}{*}{\cite{jin2023densitybased}} & \multirow{2}{*}{2023} & \multirow{2}{*}{EAAI} & Dermscopy & Skin & ISIC 2018 & \multirow{2}{*}{image} & 2,594 patients (3:1:1) & 0 & 10\%-30\% with step of 2.5\% & Skin Lesion Segmentation & \multirow{2}{*}{Segmentation} \\
 & & & X-ray& Chest& \cite{jaeger2013automatic} & & 704 patients (3:1:1) & 0 & 30\%-50\% with step of 2.5\% & Lung Segmentation &  \\ \midrule 

\cite{jimenez2023computational} & 2023 & ICCVW & Histopathology & Gland & GlaS & patch & 165 images (85/16/64) & 43\% & 5\% & Gland Segmentation & Segmentation \\ \midrule 

\cite{sadafi2023active} & 2023& ISBI & Histopathology & Breast & CAMELYON17 & WSI & 500 & 0 & 2 & \makecell[c]{Detection of Cancer\\Metastasesin Lymph Nodes} & Classification\\ \midrule 

\multirow{2}{*}{\cite{GAILLOCHET2023102958}} & \multirow{2}{*}{2023} & \multirow{2}{*}{MedIA} & MRI & Prostate & PROMISE 2012 & \multirow{2}{*}{slice} & 1,377 (1,020/109/248) & \multirow{2}{*}{10} & \multirow{2}{*}{10} & Prostate Segmentation & \multirow{2}{*}{Segmentation} \\
 & & & MRI & Hippocampus & MSD & & 9,270 (7,163/350/1,757) & & & Anterior and Posterior Hippocampus Segmentation &\\ \midrule 

\multirow{4}{*}{\cite{li2023hal}} & \multirow{4}{*}{2023} & \multirow{4}{*}{MedIA} & Ultrasound & Breast & in-house & \multirow{4}{*}{superpixel} & 3,200 images (2,600/300/300) & 260 & 130 & Breast Tumor Segmenation & \multirow{4}{*}{Segmentation} \\
 & &  & CT & Liver & in-house & & 8,797 images (8,000/400/397) & 200 & 200 & Liver Segmenation & \\
 & &  & Ultrasound & Breast & BUSI & & 647 images (400/123/124) & 40 & 40 & Breast Tumor Segmenation & \\
 & &  & X-Ray & Chest & CXRSet & & 704 images (500/102/102) & 10 & 10 & Lung Segmenation & \\ \midrule 

\cite{bai2023slpt} & 2023 & MICCAI & CT & Liver & in-house & volumes & 941 (752/189) & 0 & 20 & Liver Tumor Segmentation & Segmentation \\ \midrule 

\multirow{5}{*}{\cite{liu2023colossal}} & \multirow{5}{*}{2023} & \multirow{5}{*}{MICCAI} & MRI & Heart & \multirow{5}{*}{MSD} & \multirow{5}{*}{volume or patch} & 20 (16/4) & \multirow{5}{*}{0} & 3 & Left Atrial Segmentation & \multirow{5}{*}{Segmentation} \\ 
& & & CT & Liver & & & 260 (208/52) & & \multirow{4}{*}{5} & Liver \& Tumor Segmentation & \\ 
& & & MRI & Hippocampus & & & 131 (105/26) & & & Anterior and Posterior Hippocampus Segmentation & \\ 
& & & CT & Pancreas & & & 281 (225/56) & & & Pancreas \& Tumor Segmentation & \\ 
& & & CT & Spleen & & & 260 (208/52) & & & Spleen Segmentation & \\ 
\midrule 

\multirow{3}{*}{\cite{kadir2023edgeal}} & \multirow{3}{*}{2023} & \multirow{3}{*}{MICCAI} & \multirow{3}{*}{OCT} & \multirow{3}{*}{Eye} & Duke & \multirow{3}{*}{superpixel} & 100 scans (6:2:2) & 2\% & 10\% & \multirow{3}{*}{OCT Layer Segmentation} & \multirow{3}{*}{Segmentation} \\ 
& & & & & AROI & & 1136 scans (6:2:2) & - & - & & \\ 
& & & & & UMN & & 725 scans (6:2:2) & - & - & & \\  \midrule 

\multirow{2}{*}{\cite{tang2023pld}} & \multirow{2}{*}{2023} & \multirow{2}{*}{MICCAI} & \multirow{2}{*}{Ultrasound} & \multirow{2}{*}{Carotid} & CUBS & \multirow{2}{*}{image} & 3,220 (2,016/1,204) & 159 & 200 & \multirow{2}{*}{Carotid Intima-Media Segmentation} & \multirow{2}{*}{Segmentation} \\ 
& & & & & in-house & & 350 (test) & - & - & & \\  \midrule 

\cite{qu2023openal}  & 2023 & MICCAI & Histopathology & Colon & NCT-CRC-HE-100K & image & 100,000 & 0 & 5\% & Colorectal Cancer Diagnosis& Classification\\ \midrule 

\cite{qiu2023adaptive} & 2023 & MICCAI & Histopathology & Breast & CAMELYON16 & patch & 398 WSIs (270/128) & - & \makecell[c]{patch size of 4096, 8192, 12288\\patch per WSI of 1, 3, 5} & \makecell[c]{Detection of Cancer\\Metastasesin Lymph Nodes} & Classification\\ \midrule 

\multirow{3}{*}{\cite{chen2023making}} & \multirow{3}{*}{2023} & \multirow{3}{*}{MIDL} & Histopathology & Colon & PathMNIST & \multirow{3}{*}{image} & 107,180 (89,996/10,004/7,180) & \multirow{3}{*}{20} & \multirow{3}{*}{10} & Survival Prediction & \multirow{3}{*}{Classification}\\
 & & & CT & Abdomen & OrganAMNIST & & 58,830	(34,561/6,491/17,778) & & & Classification of Body Organs & \\
 & & & Mircoscopy & Peripheral Blood & BloodMNIST & & 17,092	(11,959/1,712/3,421) & & & Cell Type Classification & \\ \midrule 

\cite{gaillochet2023active} & 2023 & MIDL & MRI & Prostate & PROMISE 2012 & slice & 1,377 (1,020/109/248) & 10 & 10 & Prostate Segmentation & Segmentation\\ \midrule 

\cite{qu2023annotating} & 2023 & NeurIPS & CT & Abdomen & AbdomenAtlas-8K & volume & 8,448 & - & - & Multi-organ Segmentation & Segmentation \\ \midrule 

\cite{luth2024navigating} & 2023 & NeurIPS & Dermscopy & Skin & 25,331 (15,200/3,799/6,332) & image &  & 45, 225, 900 & 45, 225, 900 & Skin Lesion Diagnosis & Classification \\ \midrule 

\multirow{3}{*}{\cite{lou2023which}} & \multirow{3}{*}{2023}& \multirow{3}{*}{TMI} & \multirow{3}{*}{Histopathology} & Seven Organs& TCGA-KUMAR & patch & 30 images (12/4/14) & \multirow{3}{*}{-} & 5\% & \multirow{3}{*}{Nuclei Segmentation} & \multirow{3}{*}{Segmentation} \\
 & & & & Breast & TNBC & & 50 images (30/7/13) & & 5\% & &  \\
 & & & & Seven Organs& MoNuSeg & & 44 image (30/14) & & 7\% &  &  \\ \midrule 

\multirow{2}{*}{\cite{hu2023learning}} & \multirow{2}{*}{2023}& \multirow{2}{*}{TMI} & \multirow{2}{*}{Histopathology} & Colon & NCT-CRC-HE-100K & \multirow{2}{*}{image} & 100,000 (4:1) & 1\%, 2\% & 1\%, 2\% & Colorectal Cancer Diagnosis & \multirow{2}{*}{Classification} \\ 
& & & & Lung \& Colon & LC25000 &  & 25,000 (4:1) & 2\% & 2\% & Pulmonary \& Colorectal Cancer Diagnosis & \\ \midrule 


\multirow{2}{*}{\cite{li2024hybrid}} & \multirow{2}{*}{2024}& \multirow{2}{*}{IJCARS} & MRI & \multirow{2}{*}{Lower Extremities} & \multirow{2}{*}{in-house} & \multirow{2}{*}{volume or slice} & 119 volumes (90/9/20) & 1 volume or 190 slices & 1 volume or 190 slices & \multirow{2}{*}{Musculoskeletal Segmentation} & \multirow{2}{*}{Segmentaion}\\ 
 & & & CT & & & & 30 volumes (25/1/4) & 1 volume or 540 slices & 1 volume or 540 slices & & \\ 
\midrule 
 
\multirow{6}{*}{\cite{mahapatra2024gandalf}} & \multirow{6}{*}{2024} & \multirow{6}{*}{MedIA} & \multirow{2}{*}{X-Ray} & \multirow{2}{*}{Chest} & CheXpert & \multirow{6}{*}{image} & 224,114 (223,414/200/500) & \multirow{6}{*}{10} & \multirow{6}{*}{10} & Detection of Pleural Effusion & \multirow{6}{*}{Classification} \\
 & & & & & ChestXray14 & & 112,120 (test) & & & Thoracic Disease Diagnosis &  \\
 & & & Ultrasound & Breast & BreastMNIST & & 780 (546/78/156) & & & Breast Cancer Diagnosis &  \\
 & & & Dermscopy & Skin & DermaMNIST & & 10,015 (7,007/1,003/2,005) & & & Skin Lesion Diagnosis &  \\
 & & & Fundus & Eye & RetinaMNIST & & 1,600 (1,080/120/400) & & & Diabetic Retinopathy Severity Grading &  \\
 & & & Microscopy & Kidney & TissueMNIST & & 236,386 (165,466/23,640/47,280) & & & Kidney Cortex Disease Classification &  \\

\bottomrule
\end{tabular}}
\end{varwidth}}
\end{table*}

\subsection{Active Learning for Medical Image Classification}
\label{med_classification}

Common clinical tasks like disease diagnosis, cancer staging, and prognostic prediction can be formulated as medical image classification. 
Most AL works in medical imaging classification directly employ general methods, such as using class-balancing sampling in §\ref{class-balance} to mitigate the long-tail effect of medical imaging datasets.
However, specialized design of AL algorithms is required for certain modalities of medical image classification.
For example, the classification of chest X-rays often involves the idea of multi-label.
Besides, classifying pathological whole-slide images typically needs to be formulated as a multiple-instance learning problem. 
This section will introduce AL works specifically targeted at classification problems in chest X-rays and pathological whole-slide images.

\subsubsection{Chest X-ray and Multi-label Classification}

Chest X-ray examinations are crucial for screening and diagnosing lung, cardiovascular, skeletal, and other thoracic diseases. 
Computer-aided diagnosis in this domain has been extensively researched, including AL works aimed at reducing annotation costs for physicians.
\cite{mahapatra2021interpretabilitydriven} introduced saliency maps to select informative samples for annotation. 
To aggregate the per-pixel saliency maps into a single scalar, they explored three different approaches, including computing the kurtosis of the saliency map, utilizing multivariate radiomic features, and combining deep features of autoencoders and clustering. 
Results demonstrated that the aggregation using deep features performs the best. 
\cite{nguyen2022goal} introduced a gist-set to select samples near the decision boundary. 
Besides, uncertain samples with high entropy were sent for annotation, while the confident samples were assigned as pseudo-labels.
Additionally, they adopted momentum updates to enhance the stability of the sample predictions.
{
To handle the annotation noise, \cite{bernhardt2022active} proposed a framework called `active label cleaning'.
This framework ranked samples based on estimated label correctness and labeling difficulty.
Experiments on the chest X-ray dataset showed that the proposed method improves performance by efficiently reducing label noise with fewer expert annotations compared to random selection.
}

However, multiple diseases and abnormalities often coexist simultaneously in diagnosing chest X-rays. 
Therefore, multi-label classification has been introduced, allowing each sample to be categorized into multiple classes \citep{baltruschat2019comparison}. 
Consequently, AL algorithms for chest X-ray classification must adapt to the multi-label setting.
Built upon saliency maps, \cite{mahapatra2022graph} further introduced GNN to model the inter-relationships between different labels.
In this work, each class was treated as a node in a graph, with the relationships between classes represented as edges. 
They employed various techniques to aggregate information between different classes.
{
As a follow-up work, \cite{mahapatra2024gandalf} further introduced graph multiset transformers \citep{baek2020accurate} for more powerful inter-label relationships than GNN.
}

\subsubsection{Pathological Whole-slide Images and Multiple Instance Learning}

Compared to modalities like X-ray, CT, and MRI, pathological whole-slide images (WSIs) provide microscopic details at the cellular level, making them critically important for tasks such as cancer staging and prognostic prediction. 
However, WSIs are very large, with maximum resolutions reaching $100,000\times100,000$ pixels. 
To handle these large images for deep learning, WSIs are usually divided into many small patches. 
Fully supervised methods require patch-level or even cell-level annotations, resulting in high annotation costs. 
AL can effectively improve annotation efficiency. 
For instance, in classifying breast pathological images, \cite{qi2019labelefficient} used entropy as the uncertainty metric.
Uncertain patches were sent for annotation, whereas those with low entropy were given pseudo-labels to assist training.
{
In AL of patch-level histological tissue classification, \cite{hu2023learning} proposed category-wise curriculum querying to dynamically adjust the weight of uncertainty sampling of each class.
They further proposed negative pre-training with wrong predictions to better distinguish the visually similar classes.
To obtain fine-grained cellular annotation from WSI, \cite{van2021biological} proposed a human-augmenting AI-based labeling system with the help of AL.
An active learner was used to select the next best patch for annotation and a classifier was trained for suggesting annotation.
Specifically, Core-Set \citep{sener2018active} was used as the active learner.
Experiments with pathologists demonstrate its ability to reduce workload by around 90\% and slightly improve data quality across various cellular labeling tasks.
}

Nevertheless, pathologists might only provide WSI-level annotations in real-world clinical scenarios. 
Consequently, a prevailing direction in research is to formulate WSI classification as the weakly-supervised multi-instance learning (MIL) \citep{qu2022towards}. 
In this framework, the entire WSI is viewed as a bag, and patches within each WSI are treated as instances within that bag. 
A well-trained MIL learner can automatically identify relevant patches based on WSI-level labels, thus significantly reducing annotation costs. 
For example, a trained MIL classifier can automatically spot related patches by annotating whether or not cancer metastasis is present in a WSI. 
Nonetheless, task-relevant patches are often outnumbered by irrelevant ones, making MIL convergence more challenging.
In MIL-based pathological WSI classification, AL filters out irrelevant patches and selects informative patches for annotation. 
Based on attention-based MIL, \cite{sadafi2023active} adopted MC Dropout to estimate both attention and classification uncertainties of each patch, then sent the most uncertain patches in each WSI for expert annotation. 
\cite{qu2023openal} found that in addition to patches related to the target (e.g., tumors, lymph nodes, and normal cells), WSIs contain many irrelevant patches (e.g., fat, stroma, and debris). 
Therefore, they adopted the open-set AL \citep{ning2022active}, in which the unlabeled pool contained both target and non-target class samples. 
They combined feature distributions with prediction uncertainty to select informative and relevant patches of the target class for annotation.

\subsection{Active Learning for Medical Image Segmentation}
\label{med_segmentation}

Segmentation is one of the most common tasks in medical image analysis, capable of precisely locating anatomical structures or pathological lesions. 
However, training a segmentation model requires pixel-level annotation, which is time-consuming and labor-intensive for doctors.
Therefore, active learning has been widely used in medical image segmentation and has become an important method to reduce annotation costs.
Based on the unique traits of medical imaging, this section will focus on specialized designs in AL for medical image segmentation, including slice-based annotation, one-shot annotation, and annotation cost.

\subsubsection{Slice-based Annotation}
\label{slice}

In 3D modalities like CT and MRI, adjacent 2D slices often exhibit significant semantic redundancy.
Consequently, annotating only the key slices of each sample can reduce annotation costs. 
{
AL works mentioned in this section select 2D slices within a 3D volume for annotation.
}
Representativeness-based methods have been widely applied in this line of work. 
For instance, \cite{zheng2020annotation} utilized autoencoders to learn the semantic features of each slice, then selected and annotated key slices from axial, sagittal, and coronal planes with a strategy similar to RA \citep{zheng2019biomedical}.
Specifically, they initially trained three 2D segmentation networks and one 3D segmentation network, where the inputs for the 2D networks are slices from different planes. 
These segmentation networks were used to generate four sets of pseudo-labels and subsequently to train the final 3D segmentation network. 
Results showed that this slice-based strategy outperforms uniform sampling. 
{
Building upon this method, \cite{peng2022kcb} adopted a similar strategy in 3D knee cartilage and bone segmentation.
}
Besides, \cite{wu2022selflearning} incorporated a self-attention module into the autoencoder to enhance slice-level feature learning. 

Uncertainty methods have also been introduced for selecting key slices. 
\cite{zhou2021qualityaware} introduced a quality assessment module to select slices with the highest predicted average IoU score.
In muscle segmentation of CT images, \cite{hiasa2020automated} selected key slices and key regions. 
This work adopted clustering to select key slices and further selected regions with high uncertainty within each key slice for annotation.

{
In recent years, hybrid strategies combining both uncertainty and representativeness were proposed for slice-based annotation.
In shoulder MRI musculoskeletal segmentation, \cite{ozdemir2021active} adopted the variance of multiple MC dropout runs as the uncertainty metric. 
The posterior probability estimated by infoVAE \citep{zhao2017infovae} was used as the representativeness metric.
\cite{li2024hybrid} proposed a hybrid strategy to select informative slices in musculoskeletal segmentation of lower extremities, where uncertainty was estimated with a Bayesian U-net while the representativeness was based on cosine similarity.
They further adopted mutual information to minimize the sample redundancy following \cite{nath2021diminishing}.
The proposed method achieved impressive performances on both the MRI and CT datasets. 
}

\subsubsection{One-shot Annotation}
\label{oneshot}

Currently, most AL works require multiple rounds of annotation.
However, this setting could be impractical in medical image segmentation.
Multi-round annotation requires physicians to be readily available for each round of labeling, which is unrealistic in practice.
If physicians cannot complete the annotations on time, the AL process must be suspended. 
In contrast, one-shot annotation eliminates the need for multiple interactions with physicians. 
It also allows for selecting valuable samples in a single round, thus reducing time costs.
Both one-shot annotation and cold-start AL aim to select the most optimal initial annotations.
However, the former allows for a higher annotation budget and strictly limits the number of interactions with experts to just one.
Most relevant works combine self-supervised features and specific sampling strategies to achieve one-shot annotation.
For example, RA \citep{zheng2019biomedical} is one of the earliest works in one-shot AL for medical image segmentation.
They applied the VAE feature and a representativeness strategy to select informative samples for annotation in one shot.
RA performed excellently in gland segmentation of pathological images, whole-heart MRI images, and fungal of electron microscopic images. 
{
\cite{wu2022selflearning} proposed a representativeness-based framework for selecting key slices for annotation in one shot.
They adopted self-learning to learn the semantic representation of each slice and used it to propagate the expert annotation to different slices. 
}
\cite{jin2022oneshot} combined features of contrastive learning with farthest-first sampling to achieve one-shot annotation.
The proposed method demonstrated effectiveness on the ISIC 2018 and lung segmentation datasets. 
Additionally, \cite{jin2023densitybased} utilized auto-encoding transformations for self-supervised feature learning.
They selected and annotated samples with high density based on reachable distance.

\subsubsection{Annotation Cost}

Current AL works often assume equal annotation costs for each sample.
Yet, this is not the case in medical image segmentation, where the time to annotate different samples can differ greatly. 
AL techniques can better support physicians by considering annotation costs (e.g., annotation time).
In detecting intracranial hemorrhage of CT scans, \cite{kuo2018costsensitive} combined predictive disagreement with annotation time to select samples for annotation.
Specifically, they adopted the Jensen-Shannon divergence to measure the disagreement between the outputs of multiple models.
Annotation time for each sample was estimated by the length of the segmentation boundary and the number of connected components. 
In this work, AL was framed as a 0-1 knapsack problem, and dynamic programming is used to solve this problem for selecting informative samples.
{
In brain tumor segmentation, \cite{shen2021labeling} derived the annotation cost of a slice based on the distance between the queried slices and the already-labeled slices.
Specifically, lower distance represented lower annotation cost.
The rationale is that the annotation cost of labeling a similar slice would be cheaper than that of the unfamiliar slices.
}
In brain structure segmentation, \cite{atzeni2022deep} further considered the spatial relationships between multiple regions of interest to more accurately estimate the annotation cost.
Moreover, the average Dice coefficient of previous rounds was used to predict the average Dice for current segmentation results.
They selected and annotated regions that can maximize the average Dice.

\subsubsection{{Interactive Segmentation}}
\label{interactive}
{
Despite the success of automatic segmentation in medical imaging, there is still a potential for errors in clinical applications due to domain shifts or unseen ROIs.
Interactive segmentation \citep{budd2021survey, luo2021mideepseg} could produce real-time adjustment of the current segmentations based on the user inputs of clicks, bounding boxes, or scribbles.
As a result, interactive segmentation could rapidly tune the model towards current clinical applications with the guidance of doctors.
For the sake of flexibility, current interactive segmentation methods accept annotations for any position.
However, such a paradigm would be more efficient when the model itself could suggest where to annotate, which is exactly what active learning is good at.
Therefore, combining AL and interactive segmentation would further reduce the annotation cost.
In this section, all the mentioned papers worked interactively with different labeling units.
Before the DL era, \cite{su2015interactive} had already integrated AL in the interactive cell segmentation. 
They selected the most informative superpixels for interactive annotation with expected prediction error.
In MRI fetal brain segmentation, \cite{wang2020uncertainty} proposed an uncertainty-guided framework for interactive refinement.
They developed a novel network architecture to produce multiple segmentation results simultaneously, and the variance between different predictions served as the uncertainty metric.
Slices with the highest uncertainty were fetched for interactive refinement by human experts.
In interactive segmentation of 3D medical images, \citep{zhou2022volumetric} proposed a quality predictor, which produced a predicted IoU score with the current segmentation for each slice.
With the interactive segmentation network, the quality predictor suggested slices with lower scores for expert annotation which could be in the form of scribble, bounding box, or extreme clicking.
In \cite{li2023hal}, the most informative foreground and background superpixels were selected for interactive annotation.
}

\subsection{Active Learning for Medical Image Reconstruction}
\label{med_reconstruction}

AL can also be applied in medical image reconstruction. 
AL methods can help minimize the observations needed for modalities that require a long imaging time. 
This accelerates the imaging process and shortens the waiting period for patients. 
In this section, we'll explore the application of AL in the reconstruction of MRI, CT, and electron microscopy.
{Please refer to Table. \ref{tab_recon} for more detail.}

Deep learning has been applied to accelerate MRI acquisition and reconstruction. 
A common practice is to reduce k-space sampling through a fixed mask and use a deep model to reconstruct the undersampled MRI \citep{qin2018convolutional}. 
To further improve the imaging speed, learnable sampling in AL can be applied to select the next measurement locations in k-space.
For example, \cite{zhang2019reducing} adopted adversarial learning to train an evaluator for selecting the next row in k-space.
\cite{pineda2020active} utilized reinforcement learning to train a dual deep Q-network for active sampling in k-space. 
\cite{bakker2020experimental} adopted policy gradient in reinforcement learning to train a policy network for adaptive sampling in k-space. 
The reward for the policy network was based on the improvement in structural similarity before and after the acquisition. 
Additionally, \cite{bakker2022learning} explored how to jointly optimize the reconstruction and acquisition networks. 

In addition to MRI imaging, AL has been employed in CT reconstruction as illustrated by \cite{wang2022active}. 
They adaptively chose the scanning angles tailored to individual patients, leading to a reduction in both radiation exposure and scanning duration. 
In electron microscopy, \cite{mi2020learning} initially enhanced low-resolution images to high-resolution and then predicted the location of region-of-interest and reconstruction error. 
A weighted DPP based on reconstruction error was applied to select pixels that needed to be rescanned. 
Results showed that weighted DPP maintained both low reconstruction error and spatial diversity.

\begin{table*}[!htb]
\centering
\caption{Summarization of surveyed works of active learning in medical image reconstruction.}
\label{tab_recon}
\scalebox{0.7}{
\begin{tabular}{cccccccccccc}

\toprule
 & Year& Venues& Modality & ROIs & Dataset & \multicolumn{1}{c}{Clinical Task} \\
 \midrule

 \multirow{2}{*}{\cite{jin2019self}} & \multirow{2}{*}{2019}& \multirow{2}{*}{arXiv} & \multirow{2}{*}{MRI} & Heart& Cardiac Atlas Project  & MRI Reconstruction\\
& & & & Knee & fastMRI & MRI Reconstruction\\ \midrule 

\cite{zhang2019reducing} & 2019& CVPR & MRI & Knee & fastMRI & MRI Reconstruction\\ \midrule 

\multirow{2}{*}{\cite{mi2020learning}} & \multirow{2}{*}{2020}& \multirow{2}{*}{MICCAI}& \multirow{2}{*}{Electron Microscopy}& Mouse Cortex& SNEMI3D & \multirow{2}{*}{\makecell[c]{Accelerated Acquisition of\\Electron Microscopy}} \\
 & & & & Human Cerebrum & in-house & \\ \midrule 

\cite{pineda2020active} & 2020 & MICCAI & MRI & Knee & fastMRI & MRI Reconstruction \\ \midrule 

\multirow{2}{*}{\cite{bakker2020experimental}} & \multirow{2}{*}{2020}& \multirow{2}{*}{NeurIPS}& \multirow{2}{*}{MRI} & Knee & fastMRI & MRI Reconstruction\\
& & & & Brain& fastMRI & MRI Reconstruction\\ \midrule 

\multirow{2}{*}{\cite{wang2022active}} & \multirow{2}{*}{2022}& \multirow{2}{*}{arXiv} & \multirow{2}{*}{CT} & Lung & AAPM & CT Reconstruction \\
& & & & Spine& VerSe & CT Reconstruction \\

\bottomrule
\end{tabular}}
\end{table*}

\section{{Performance Evaluation of Active Learning in Medical Image Analysis}}
\label{exps}
{In the field of medical image analysis, there is now an increasing amount of AL works.
Despite its rapid development, AL for medical image analysis still faces several issues that limit its application in real-world clinical tasks.
On one hand, there is a lack of comprehensive evaluation of AL methods on the medical imaging datasets.
Most AL works conducted experiments on standard datasets, such as CIFAR-10, CIFAR-100, or MedMNIST.
However, real-world medical imaging datasets often contain less available data and higher complexity for analysis \citep{varoquaux2022machine}.
Some AL works focused on a specific domain of medical imaging and achieved excellent performance, but their potential to generalize to a wider aspect of applications remains questionable.
Beyond that, different AL methods exhibit inconsistent performance and may not necessarily outperform random sampling.
In AL of classification tasks, \cite{munjal2022robust} highlighted the absence of a consistently outperforming AL method and the fact that random sampling performs relatively well.
In §\ref{self}, we have also mentioned that the AL methods are inferior to random sampling when the annotation budget is low.
As a result, we are uncertain about which AL method works according to our requirements and whether this method could outperform the most straightforward baseline, random sampling. }

{
To clarify the aforementioned issues, we have conducted a comprehensive evaluation of different AL methods on multiple medical imaging datasets.
The adopted three datasets are widely used by the entire medical imaging community.
They also correspond to different modalities, organs, and tasks (e.g., classification and segmentation).
We choose the most representative and popular AL methods for their evaluations of the medical imaging datasets.
Besides, we provided the details of dataset splits, network architecture, and training hyperparameters for better reproducibility.
Codes are also available on our accompanying website \footnote{{https://github.com/LightersWang/Awesome-Active-Learning-for-Medical-Image-Analysis/tree/main/code}}.}

\subsection{{Experimental Setups}}

\begin{table}[t]
\centering
\caption{{Training, validation, and testing splits of each dataset. Unless specified otherwise, the figures presented in this table represent the number of images of each split.}}
\label{tab_splits}
\scalebox{0.8}{
\begin{tabular}{cccc}
\toprule
& NCT-CRC-HE-100K & ISIC 2020 & ACDC \\
\midrule
Training & 90,000 & 20,869 & \begin{tabular}[c]{@{}c@{}}656\\ (slices)\end{tabular} \\
\midrule
Validation & 10,000 & 2,319 & \begin{tabular}[c]{@{}c@{}}10\\ (volumes)\end{tabular} \\
\midrule
Testing & \begin{tabular}[c]{@{}c@{}}7,180\\ (CRC-VAL-HE-7K)\end{tabular} & 9,938 & \begin{tabular}[c]{@{}c@{}}20\\ (volumes)\end{tabular}\\
\bottomrule
\end{tabular}}
\end{table}

\subsubsection{{Datasets}}
{
In this survey, we chose three medical imaging datasets for the performance evaluation of the AL methods, including two classification datasets and one segmentation dataset. 
The descriptions and dataset split are presented as follows, where a summarized table of the dataset split is in Table \ref{tab_splits}.}

{\textbf{NCT-CRC-HE-100K} \citep{kather2019predicting}: This dataset contains 100,000 patches from 86 hematoxylin \& eosin (H\&E) stained histological slides of human colorectal cancer and normal tissue. 
All the patches are $224 \times 224$ at 0.5 microns per pixel.
The patches are grouped into nine classes of different tissues, including adipose (ADI), background (BACK), debris (DEB), lymphocytes (LYM), mucus (MUC), smooth muscle (MUS), normal colon mucosa (NORM), cancer-associated stroma (STR), and colorectal adenocarcinoma epithelium (TUM).
For the dataset split, we divided the dataset into training and validation sets with a ratio of 9:1 and utilized an additional dataset CRC-VAL-HE-7K which is provided by the same authors as the testing set.
CRC-VAL-HE-7K shares the same acquisition protocol and tissue classes with NCT-CRC-HE-100K but contains 7,180 patches from 50 patients other than the patients of NCT-CRC-HE-100K.}

{\textbf{ISIC 2020} \citep{rotemberg2021patient}: ISIC 2020 is composed of 33,126 dermoscopic images from over 2,000 patients. 
Each image is labeled as benign or malignant by either doctors, long-term follow-up, or histopathology.
ISIC 2020 contains 32,542 images of benign lesions and only 584 images of malignant lesions.
We split this dataset of training, validation, and testing sets with a ratio of 6:1:3.}

{\textbf{ACDC} \citep{bernard2018deep}: This dataset contains short-axis cardiac cine-MR images from 100 patients.
In this survey, we only adopted the end-diastolic frame of each patient for evaluating different AL methods, which resulted in a total of 100 scans.
Each scan corresponds to a human-annotated segmentation mask of the left ventricle (LV), myocardium (MYO), and right ventricle (RV).
We followed the split from \cite{luo2022semi}, which contains 70, 10, and 20 scans in the training, validation, and testing sets, respectively.
Due to the large spacing along the z-axis, 2D segmentation is more appropriate compared to 3D segmentation.
Therefore, we trained the segmentation model with 2D slices and evaluated it with 3D volumes following \cite{bai2017semi}.
Therefore, the training set is composed of 656 slices. }

\subsubsection{{Evaluation Metrics}}
{
We employed different evaluation metrics for the task of each dataset.
For the multi-class classification of NCT-CRC-HE-100K, we adopted the accuracy (ACC) to evaluate the classification performance.
Due to the heavy class-imbalance of the binary classification task of ISIC 2020, we adopted the area under the receiver operating characteristic curve (AUC) for evaluation. 
For the segmentation task of ACDC, two well-known metrics of Dice similarity coefficient (DSC) and average surface distance (ASD) are used.
DCS ranges from 0\% (non-overlap) to 100\% (perfect segmentation), and the lower ASD indicates a better alignment between the segmentation prediction and ground truth.
To evaluate the overall performance, we presented the mean DSC and ASD of LV, MYO, and RV.
Evaluation metrics on the testing set were reported as the final results.
}
\begin{table*}[!th]
\caption{{Accuracy of different active learning methods for multi-class pathological tissue classification.  We reported the test performance of the initially labeled dataset and other active learning rounds with mean and standard deviation. The best and second-best results are bolded in red and blue, respectively.}}
\label{nct_exps}
\centering
\subtable[Low budget ($b = 50$)]{
\scalebox{0.57}{
\begin{tabular}{c|cccccc}
\toprule
ACC (\%) & 50 & 100 & 150 & 200 & 250 & 300 \\
\midrule
Random & \multirow{9}{*}{\makecell[c]{60.46}} & 64.39 $\pm$ 0.34 & 66.80 $\pm$ 3.66 & 71.77 $\pm$ 3.35 & 73.55 $\pm$ 2.38 & \textcolor{blue}{\textbf{75.87 $\pm$ 1.21}} \\
Confidence & & 63.61 $\pm$ 2.49 & 66.16 $\pm$ 3.12 & 66.29 $\pm$ 3.84 & 70.65 $\pm$ 2.85 & 71.69 $\pm$ 0.97 \\
Entropy & & 62.65 $\pm$ 0.79 & 65.66 $\pm$ 1.31 & 68.78 $\pm$ 2.34 & 69.12 $\pm$ 2.99 & 68.86 $\pm$ 3.84 \\
Margin & & \textcolor{red}{\textbf{70.65 $\pm$ 2.20}} & \textcolor{blue}{\textbf{71.33 $\pm$ 1.54}} & \textcolor{red}{\textbf{75.93 $\pm$ 3.00}} & \textcolor{blue}{\textbf{74.64 $\pm$ 4.21}} & \textcolor{red}{\textbf{76.61 $\pm$ 2.90}} \\
BALD & & 62.13 $\pm$ 1.09 & 64.17 $\pm$ 4.20 & 69.34 $\pm$ 4.53 & 66.94 $\pm$ 2.50 & 68.49 $\pm$ 2.75 \\
DBAL & & 61.73 $\pm$ 2.43 & 63.76 $\pm$ 5.33 & 67.8 $\pm$ 1.78  & 67.90 $\pm$ 3.06 & 71.48 $\pm$ 3.38 \\
BADGE & & \textcolor{blue}{\textbf{68.52 $\pm$ 6.57}} & \textcolor{red}{\textbf{72.27 $\pm$ 3.45}} & \textcolor{blue}{\textbf{74.69 $\pm$ 0.85}} & \textcolor{red}{\textbf{75.96 $\pm$ 2.65}} & 75.74 $\pm$ 2.22 \\
Core-Set-L2 & & 62.73 $\pm$ 4.33 & 66.26 $\pm$ 2.44 & 64.72 $\pm$ 3.94 & 66.25 $\pm$ 2.98 & 64.66 $\pm$ 1.24 \\
Core-Set-Cosine & & 63.07 $\pm$ 1.84 & 63.12 $\pm$ 4.20 & 69.32 $\pm$ 1.37 & 68.25 $\pm$ 2.30 & 68.21 $\pm$ 3.48 \\
\midrule
Fully supervised & \multicolumn{6}{c}{93.36} \\
\bottomrule
\end{tabular}
}}
\subtable[High budget ($b = 1000$)]{
\scalebox{0.57}{
\begin{tabular}{c|cccccc}
\toprule
ACC (\%) & 1000 & 2000   & 3000   & 4000  & 5000   & 6000   \\
\midrule
Random & \multirow{9}{*}{\makecell[c]{83.38}} & 87.16 $\pm$ 1.43 & 87.68 $\pm$ 1.15 & 89.00 $\pm$ 0.91 & 89.04 $\pm$ 2.41 & \textcolor{red}{\textbf{89.09 $\pm$ 1.08}} \\
Confidence & & 87.93 $\pm$ 0.76 & 87.46 $\pm$ 0.96 & 88.45 $\pm$ 0.64 & \textcolor{red}{\textbf{90.01 $\pm$ 1.22}} & 87.93 $\pm$ 3.04 \\
Entropy & & \textcolor{blue}{\textbf{89.39 $\pm$ 0.23}} & 86.54 $\pm$ 2.33 & 87.70 $\pm$ 2.25 & 88.04 $\pm$ 2.89 & \textcolor{blue} {\textbf{89.06 $\pm$ 1.57}} \\
Margin &  & 88.60 $\pm$ 0.62 & 88.05 $\pm$ 1.83 & 88.13 $\pm$ 1.74 & \textcolor{blue}{\textbf{89.90 $\pm$ 0.63}} & 86.89 $\pm$ 2.99 \\
BALD & & 87.01 $\pm$ 0.99 & 85.85 $\pm$ 1.96 & \textcolor{red}{\textbf{90.11 $\pm$ 0.53}} & 89.19 $\pm$ 2.46 & 88.24 $\pm$ 2.21 \\
DBAL & & \textcolor{red}{\textbf{89.55 $\pm$ 0.75}} & 86.56 $\pm$ 1.86 & 86.92 $\pm$ 3.62 & 88.76 $\pm$ 1.97 & 88.44 $\pm$ 0.53 \\
BADGE & & 87.31 $\pm$ 0.51 & 87.56 $\pm$ 0.90 & 88.66 $\pm$ 2.33 & 88.39 $\pm$ 0.97 & 87.17 $\pm$ 1.30 \\
Core-Set-L2 & & 86.82 $\pm$ 0.37 & \textcolor{red}{\textbf{88.99 $\pm$ 1.74}} & \textcolor{blue}{\textbf{89.92 $\pm$ 0.98}} & 89.34 $\pm$ 1.87 & 88.64 $\pm$ 3.10 \\
Core-Set-Cosine &   & 87.40 $\pm$ 0.49 & \textcolor{blue}{\textbf{88.09 $\pm$ 2.35}} & 89.51 $\pm$ 1.20 & 89.80 $\pm$ 3.01 & 87.19 $\pm$ 2.78 \\
\midrule
Fully supervised & \multicolumn{6}{c}{93.36} \\
\bottomrule
\end{tabular}
}}
\end{table*}

\subsubsection{{Active Learning Settings}}

{In this study, we performed $T = 5$ rounds of annotation.
To investigate how the number of queried samples in each round (i.e., annotation budget) affects the performance of different AL methods, we set different levels of annotation budgets $b$ for each dataset.
Following \cite{luth2024navigating}, high ($b = 1000$) and low ($b = 50$) budgets were used for classification tasks of NCT-CRC-HE-100K and ISIC 2020.
Following \cite{GAILLOCHET2023102958}, we adopted a budget of 10 slices ($b = 10$) for ACDC segmentation.
Considering the sizes of the involved datasets, the low budgets for classification ($b = 50$) and segmentation ($b = 10$) provide a chance to look into the performance of different AL methods in the low data regime. 
Before the active learning process started, we randomly selected the initial labeled pool to train an initial model.
The size of the initial pool is equal to the annotation budget.
The model training and sample selection are seeded.
We ran each active learning method of a specific budget and dataset five times with different random seeds and reported the mean and standard deviation as the result.}

\subsubsection{{Comparison Methods}}
{
For fairness and reproducibility, we conducted the evaluation using the following methods: 
\textbf{Random}: the baseline of active learning, which randomly draws the unlabeled samples.
\textbf{Confidence, Entropy, and Margin} \cite{lewis1994heterogeneous, joshi2009multiclass, roth2006marginbased}: these methods are all classic uncertainty-based AL methods, which calculated the confidence, entropy, and margin with the prediction probability as the uncertainty scores. 
Lower confidence, higher entropy, and lower margin indicate higher uncertainty.
\textbf{DBAL} \cite{gal2017deep}: This method integrated entropy and MC dropout for better uncertainty estimation. 
During sample selection, the model runs multiple times with all dropout layers activated, and the average probability of all MC dropout runs is used for calculating entropy.
\textbf{BALD} \cite{gal2017deep}: This method calculated BALD as the uncertainty score, which aims to maximize the mutual information between predictions and model parameters.
MC dropout was also used in this method.
\textbf{Core-Set} \cite{sener2018active}: This method performed cover-based sampling using the feature embedding of each sample.
For the balance of computation time and performance, we used the k-Center-Greedy for sample selection.
To investigate how the distance metric affected the AL performance, we proposed a variant of Core-Set named ``Core-Set-Cosine'' which replaced the original L2 distance with the cosine distance.
The original Core-Set was referred to as ``Core-Set-L2'' to avoid confusion.
\textbf{BADGE} \cite{ash2020deep}: This method applied gradient as uncertainty estimation and utilized k-Means++ to improve diversity.
Specifically, the gradient of cross-entropy loss was used in the classification task while the segmentation task used the gradient of the sum of the Dice loss and cross-entropy loss.}

{
It should be noted that uncertainty-based methods in segmentation are slightly different from those of the classification. 
Specifically, we first produced the pixel-wise scores and then utilized the averaged scores for sample selection in the segmentation task.}

\begin{table*}[!th]
\caption{{AUC of different active learning methods for the binary skin lesion classification. We reported the test performance on ISIC 2020 of the initially labeled dataset and other active learning rounds with mean and standard deviation. The best and second-best results are bolded in red and blue, respectively.}}
\label{isic_exps}
\centering
\subtable[Low budget ($b = 50$)]{
\scalebox{0.5}{
\begin{tabular}{c|cccccc}
\toprule
AUC & 50 & 100 & 150 & 200 & 250 & 300 \\ \midrule
Random & & 0.5609 ± 0.0325 & 0.5408 ± 0.0228 & 0.5807 ± 0.0391 & 0.5679 ± 0.0632 & 0.6143 ± 0.0610 \\
Confidence & & 0.5432 ± 0.0233 & 0.5713 ± 0.0472 & 0.5622 ± 0.0302 & 0.5584 ± 0.0171 & 0.5649 ± 0.0794 \\
Entropy & & 0.5375 ± 0.0182 & 0.5432 ± 0.0209 & 0.5919 ± 0.0413 & 0.5537 ± 0.0383 & 0.5543 ± 0.0220 \\
BALD & & {\color[HTML]{FF0000} \textbf{0.5865 ± 0.0235}} & 0.5828 ± 0.0499 & 0.5818 ± 0.0800 & 0.5977 ± 0.0273 & 0.5682 ± 0.0313 \\
DBAL & & 0.5322 ± 0.0300 & 0.5344 ± 0.0226 & 0.5601 ± 0.0341 & 0.5879 ± 0.0274 & 0.5442 ± 0.0350 \\
BADGE & & 0.5601 ± 0.0373 & 0.5648 ± 0.0333 & {\color[HTML]{FF0000} \textbf{0.6219 ± 0.0757}} & {\color[HTML]{FF0000} \textbf{0.6246 ± 0.0894}} & 0.6172 ± 0.0890 \\
Core-Set-L2 & & {\color[HTML]{0000FF} \textbf{0.5834 ± 0.0344}} & {\color[HTML]{FF0000} \textbf{0.6761 ± 0.0382}} & 0.5956 ± 0.0610 & 0.5934 ± 0.0464 & {\color[HTML]{0000FF} \textbf{0.6313 ± 0.0867}} \\
Core-Set-Cosine & \multirow{-8}{*}{0.574} & 0.5459 ± 0.0367 & {\color[HTML]{0000FF} \textbf{0.6361 ± 0.0621}} & {\color[HTML]{0000FF} \textbf{0.6032 ± 0.0402}} & {\color[HTML]{0000FF} \textbf{0.6127 ± 0.0586}} & {\color[HTML]{FF0000} \textbf{0.6317 ± 0.0797}} \\
\midrule
Fully supervised & \multicolumn{6}{c}{0.8424} \\ 
\bottomrule
\end{tabular}
}}
\subtable[High budget ($b = 1000$)]{
\scalebox{0.5}{
\begin{tabular}{c|cccccc}
\toprule
AUC & 1000 & 2000 & 3000 & 4000 & 5000 & 6000 \\
\midrule
Random & & 0.6898 ± 0.0216 & 0.7258 ± 0.0249 & 0.7556 ± 0.0236 & 0.7551 ± 0.0157 & 0.7718 ± 0.0124 \\
Confidence & & 0.7058 ± 0.0120 & 0.7525 ± 0.0069 & {\color[HTML]{0000FF} \textbf{0.7760 ± 0.0171}} & 0.7957 ± 0.0184 & {\color[HTML]{FF0000} \textbf{0.7949 ± 0.0113}} \\
Entropy & & 0.7063 ± 0.0144 & 0.7478 ± 0.0090 & 0.7692 ± 0.0117 & 0.7838 ± 0.0062 & 0.7875 ± 0.0101 \\
BALD & & 0.6851 ± 0.0156 & 0.7489 ± 0.0144 & 0.7683 ± 0.0097 & {\color[HTML]{0000FF} \textbf{0.7947 ± 0.0098}} & 0.7837 ± 0.0073 \\
DBAL & & 0.7120 ± 0.0195 & {\color[HTML]{FF0000} \textbf{0.7613 ± 0.0053}} & 0.7616 ± 0.0261 & 0.7896 ± 0.0120 & 0.7883 ± 0.0127 \\
BADGE & & 0.7074 ± 0.0264 & 0.7456 ± 0.0103 & 0.7734 ± 0.0150 & {\color[HTML]{FF0000} \textbf{0.7983 ± 0.0100}} & {\color[HTML]{0000FF} \textbf{0.7917 ± 0.0064}} \\
Core-Set-L2 & & {\color[HTML]{FF0000} \textbf{0.7304 ± 0.0303}} & 0.7511 ± 0.0156 & 0.7499 ± 0.0106 & 0.7588 ± 0.0056 & 0.7575 ± 0.0067 \\
Core-Set-Cosine & \multirow{-8}{*}{0.644} & {\color[HTML]{0000FF} \textbf{0.7288 ± 0.0236}} & {\color[HTML]{0000FF} \textbf{0.7593 ± 0.0177}} & {\color[HTML]{FF0000} \textbf{0.7807 ± 0.0182}} & 0.7925 ± 0.0135 & 0.7917 ± 0.0109 \\
\midrule
Fully supervised & \multicolumn{6}{c}{0.8424} \\ 
\bottomrule
\end{tabular}
}}
\end{table*}

\subsubsection{{Implementation Details}}
{
\textbf{Classification:} For all the classification task, we used ResNet-18 \citep{he2016deep} as the backbone and the loss function is cross-entropy.
We trained the model using stochastic gradient descent with momentum for 100 epochs with a batch size of 128.
The learning rate and momentum were set as 0.01 and 0.9, respectively.
Also, the cosine learning rate decay was adopted for smoother convergence.
Data augmentations of the input image are different in the two classification datasets.
In NCT-CRC-HE-100K, we only used the random horizontal flip.
For ISIC 2020, we followed the data augmentation from \cite{zhuang2018skin} which includes random crop, flip, rotation, affine transforms, and color jittering.}

{\textbf{Segmentation:} We used a 5-level U-Net \citep{ronneberger2015u} for segmentation.
Each level of the encoder or decoder contains two blocks.
Each block consists of a 2D convolution, dropout layer with a probability of 0.1, batch normalization, and leaky ReLU activation.
The segmentation loss is the combination of cross-entropy loss and Dice loss.
We trained the model using the Adam optimizer \citep{kingma2014adam} for 4,000 iterations with a batch size of 32.
The learning rate is 0.001 while decaying along the training iterations with the polynomial schedule.
Data augmentation includes random flip, rotation of 90 degrees, and rotation of arbitrary degrees.} 

{Experiments were conducted on NVIDIA GeForce RTX 3090 and 4090 GPUs and the CUDA version is 11.3. 
Codes are implemented using Python (version 3.8.10) and the PyTorch framework (version 1.11.0).}

\subsection{{Experimental Results and Performance Analysis}}

\subsubsection{{Active Learning Results for Pathological Tissue Classification}}
{
We first evaluated the active learning performance on the pathological tissue classification task. The results of the testing accuracy are shown in Table \ref{nct_exps}. 
Margin performed well in the low-budget scenario. 
The reason for that is this method exploits the information of the wrong predictions of a similar class, which is in line with the finding in \cite{hu2023learning}. 
BADGE performs well in the low-budget scenarios largely due to the k-Means++ clustering. 
However, its performance drops in the high-budget scenario, which may indicate that the gradient embeddings are less suitable in AL when there is a distribution shift between the training and testing sets. 
The results of this section call for a more in-depth investigation of the generalizability to distribution shift of the AL methods.
}

\subsubsection{{Active Learning Results for Skin Lesion Classification}}
{
We have also conducted thorough evaluations on the ISIC 2020 dataset, which corresponds to a binary classification problem with severe class-imbalance. 
The AUC of the test split is shown in Table \ref{isic_exps}. 
It should be noted that Confidence and Margin are equivalent in the binary setting, so we only report the results of the former. 
In the low-budget scenario, Core-Set and its variant achieved better performances compared to the uncertainty-based methods. 
It indicates that representativeness-based methods or methods with improved diversity are more favored than uncertainty-based ones when the budget is low and the task is extremely difficult. 
Among all the uncertainty-based methods, BADGE stands out in certain rounds for its clustering operations that enhance diversity. 
For the high budget, the performance of the Core-Set variants is still competitive.
However, the performance of the uncertainty-based methods improved. 
Results here demonstrated how the annotation budget affects the performance of the uncertainty-based and representativeness-based methods, in which the former fits a higher budget while the latter fits a lower budget.
}

\begin{table*}[!th]
\caption{{Mean DSC and ASD of different active learning methods for cardiac MRI segmentation. We reported the test metrics on ACDC of the initially labeled dataset and other active learning rounds with mean and standard deviation. The best and second-best results are bolded in red and blue, respectively.}}
\label{acdc_exps}
\centering
\subtable[Mean DSC]{
\scalebox{0.55}{
\begin{tabular}{c|cccccc}
\toprule
DSC & 10 & 20 & 30 & 40 & 50 & 60 \\
\midrule
Random &   & 0.7872 ± 0.0552 & 0.8285 ± 0.0379 & {\color[HTML]{FF0000} \textbf{0.8657 ± 0.0063}} & 0.8656 ± 0.0089 & {\color[HTML]{0000FF} \textbf{0.8721 ± 0.0047}} \\
Confidence  &   & 0.6805 ± 0.0205 & 0.7499 ± 0.0336 & 0.8347 ± 0.0203 & {\color[HTML]{0000FF} \textbf{0.8696 ± 0.0057}} & 0.8717 ± 0.0065 \\
Entropy &   & 0.6888 ± 0.0088 & 0.7712 ± 0.0393 & 0.8538 ± 0.0117 & 0.8639 ± 0.0062 & 0.8714 ± 0.0040 \\
Margin &   & 0.6841 ± 0.0124 & 0.7360 ± 0.0338 & 0.8214 ± 0.0535 & 0.8640 ± 0.0096 & {\color[HTML]{FF0000} \textbf{0.8734 ± 0.0053}} \\
BADGE  &   & {\color[HTML]{FF0000} \textbf{0.8192 ± 0.0144}} & {\color[HTML]{0000FF} \textbf{0.8486 ± 0.0151}} & {\color[HTML]{0000FF} \textbf{0.8632 ± 0.0060}} & {\color[HTML]{FF0000} \textbf{0.8731 ± 0.0032}} & 0.8698 ± 0.0080 \\
Core-Set-L2 & \multirow{-6}{*}{0.4966} & {\color[HTML]{0000FF} \textbf{0.8112 ± 0.0210}} & {\color[HTML]{FF0000} \textbf{0.8491 ± 0.0144}} & 0.8570 ± 0.0074 & 0.8572 ± 0.0057 & 0.8660 ± 0.0073 \\
Core-Set-Cosine & & 0.7344 ± 0.0234 & 0.8248 ± 0.0109 & 0.8458 ± 0.0225 & 0.8549 ± 0.0204 & 0.8678 ± 0.0089 \\
\midrule
Fully supervised & \multicolumn{6}{c}{0.9045} \\
\bottomrule
\end{tabular}
}}
\subtable[Mean ASD]{
\scalebox{0.55}{
\begin{tabular}{c|cccccc}
\toprule
ASD (mm) & 10 & 20 & 30 & 40 & 50 & 60 \\
\midrule
Random &  & 4.71 ± 1.13 & {\color[HTML]{0000FF} \textbf{2.82 ± 0.37}} & 3.13 ± 0.48 & {\color[HTML]{FF0000} \textbf{2.05 ± 0.70}} & {\color[HTML]{0000FF} \textbf{2.21 ± 0.73}} \\
Confidence  &  & 6.63 ± 2.05 & 3.60 ± 1.39 & 2.86 ± 0.39 & {\color[HTML]{0000FF} \textbf{2.21 ± 0.98}} & 2.22 ± 0.62 \\
Entropy &  & 7.80 ± 1.70 & 3.27 ± 1.22 & 3.00 ± 1.08 & 2.24 ± 1.23 & 2.67 ± 0.84 \\
Margin &  & 6.64 ± 3.17 & 4.65 ± 1.94 & {\color[HTML]{FF0000} \textbf{2.66 ± 0.76}} & 3.08 ± 1.02 & 3.22 ± 1.41 \\
BADGE  &  & {\color[HTML]{0000FF} \textbf{4.22 ± 1.10}} & 4.14 ± 0.32 & 3.10 ± 0.70 & 2.62 ± 0.81 & 2.56 ± 0.97 \\
Core-Set-L2 & \multirow{-6}{*}{11.36} & {\color[HTML]{FF0000} \textbf{3.22 ± 0.69}} & 2.84 ± 0.66 & {\color[HTML]{0000FF} \textbf{2.71 ± 0.76}} & 3.40 ± 1.31 & 2.86 ± 0.76 \\
Core-Set-Cosine & & 4.75 ± 1.57 & {\color[HTML]{FF0000} \textbf{2.74 ± 0.52}} & 3.50 ± 1.36 & 2.33 ± 0.88 & {\color[HTML]{FF0000} \textbf{1.85 ± 0.31}}  \\
\midrule
Fully supervised & \multicolumn{6}{c}{2.09} \\
\bottomrule
\end{tabular}
}}
\end{table*}

\subsubsection{{Active Learning Results for MRI Cardiac Segmentation}}
{
For segmentation, we evaluated different AL methods on the ACDC dataset. 
We reported the mean DSC and ASD of the segmentation results in Table \ref{acdc_exps}. 
BADGE achieved the best or second-best performance in mean DSC for multiple rounds. 
Core-Set performed well on both the mean DSC and ASD in the early rounds of the lower budget scenario.
Both two methods improve sampling diversity to some extent. 
However, for the later rounds, the performance of the uncertainty-based methods and random sampling improves in mean DSC and ASD. 
This result aligns with the findings in the previous section.
}

\subsubsection{{Effectiveness on Different Distances between Images}}
{
Distance measurement plays an important role in AL which may significantly impact the performance of AL algorithms.
In this section, we evaluated the performance of the two most popular distances in AL, which are L2 and cosine distances.
These two distances are based on the feature embedding.
Assume $x$ stands for the sample itself and its corresponding feature embedding is $\mathbf{z} = \left[z_1, z_2, \ldots, z_d \right]$, $d$ is the feature dimension.
Based on the feature embedding, the L2 distance between two images $x^a$ and $x^b$ is as follow:
\begin{equation}
    \text{L2} (x^a, x^b) = \text{L2} (\mathbf{z}^a, \mathbf{z}^b) = \sqrt{\sum_{i=1}^{d} (z^a_i - z^b_i)^2}
\end{equation}
while the cosine distance is:
\begin{small}
\begin{equation}
    \text{Cosine}(x^a, x^b) = 1 - \frac{\mathbf{z}^a \cdot \mathbf{z}^b}{\|\mathbf{z}^a\| \cdot \|\mathbf{z}^b\|} = 1 - \frac{\sum_{i=1}^{d} z^a_i z^b_i}{\sqrt{\sum_{i=1}^{d} {z^a_i}^2} \sqrt{\sum_{i=1}^{d} {z^b_i}^2}}
\end{equation}
\end{small}
To conduct experiments, we replaced the L2 distance in Core-Set with the cosine distance.}

\begin{figure*}[!t]
    \centering
    \subfigure[NCT-CRC-HE-100K ($b = 1000$)]
    {
      \begin{minipage}[b]{0.3\linewidth}
      \includegraphics[width=\textwidth]{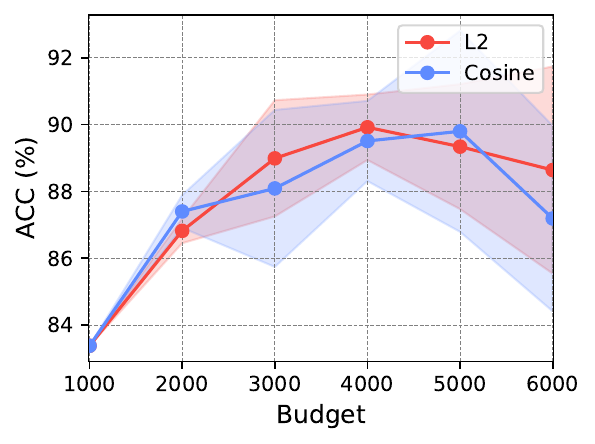}
      \end{minipage}
    }
    \subfigure[NCT-CRC-HE-100K ($b = 50$)]
    {
      \begin{minipage}[b]{0.3\linewidth}
      \includegraphics[width=\textwidth]{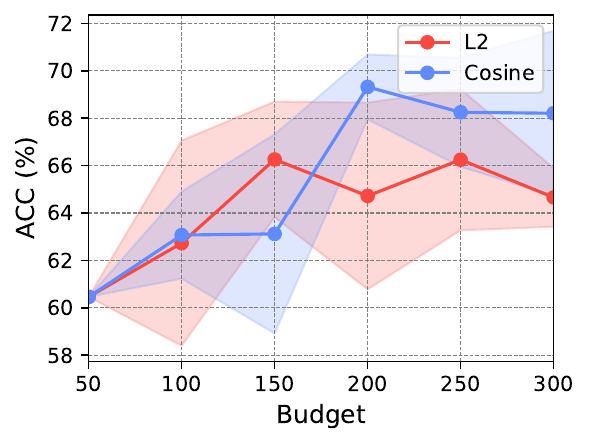}
      \end{minipage}
    }

    \subfigure[ISIC 2020 ($b = 1000$)]
    {
      \begin{minipage}[b]{0.3\linewidth}
      \includegraphics[width=\textwidth]{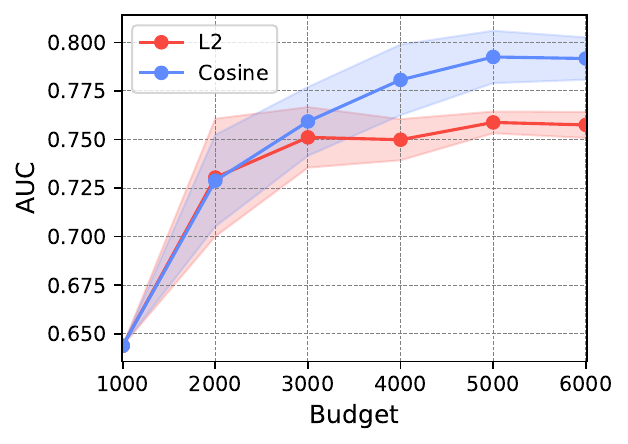}
      \end{minipage}
    }
    \subfigure[ISIC 2020 ($b = 50$)]
    {
      \begin{minipage}[b]{0.3\linewidth}
      \includegraphics[width=\textwidth]{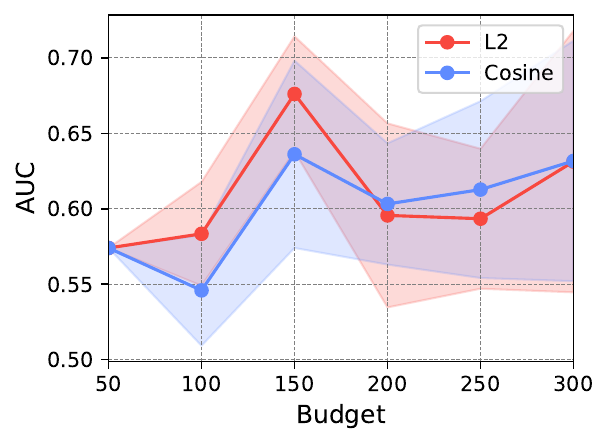}
      \end{minipage}
    }

    \subfigure[ACDC (Mean DSC)]
    {
      \begin{minipage}[b]{0.3\linewidth}
      \includegraphics[width=\textwidth]{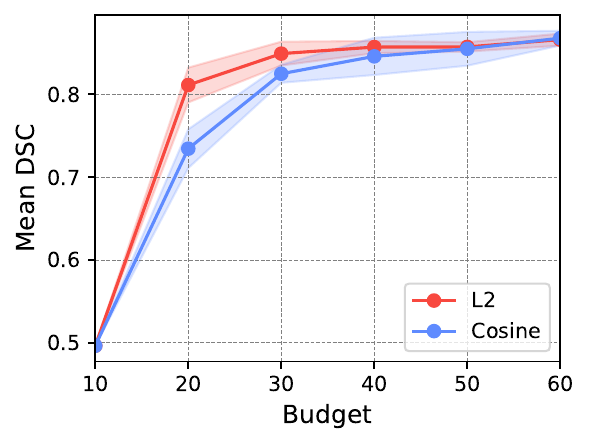}
      \end{minipage}
    }
    \subfigure[ACDC (Mean ASD)]
    {
      \begin{minipage}[b]{0.3\linewidth}
      \includegraphics[width=\textwidth]{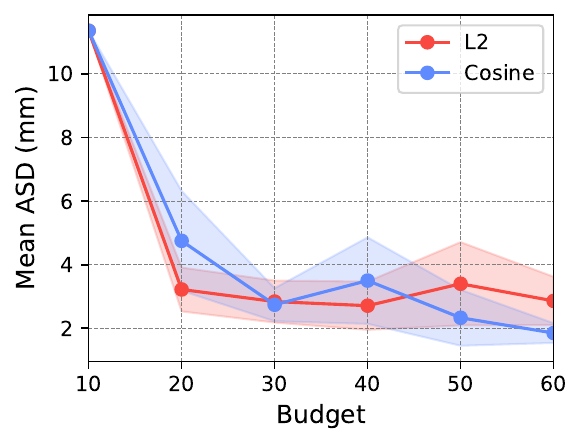}
      \end{minipage}
    }
    
    \caption{{Performance comparisons between the L2 and cosine distances on all three datasets. We used Core-Set as the base AL method.}}
    \label{fig_distances}
\end{figure*}

{
Performance comparisons between Core-Set-L2 and Core-Set-Cosine are illustrated in Fig.\ref{fig_distances}. 
Results on the NCT-CRC-HE-100K dataset showed no significant difference between the L2 and cosine distances across all budget levels. 
In ISIC 2020, the L2 distance tends to be better in the early rounds, indicating its ability to rapidly start the model, while Core-Set-Cosine significantly outperformed the Core-Set-L2 when the budget was high. 
On the ACDC dataset, Core-Set-L2 also outperformed Core-Set-Cosine in the early rounds but their performance after selecting more samples is similar. 
These results indicate that distance metrics play an important role in the performance of AL methods used in medical image analysis, and they should be carefully chosen according to the target tasks and the budgets. 
Generally speaking, L2 distance is more suitable for the low-budget scenario while the cosine distance might be a better choice when the budget is high.
}

\section{Challenges and Future Perspectives} 
\label{challenge}
Currently, annotation scarcity is a significant bottleneck hindering the development of medical image analysis. 
AL improves annotation efficiency by selectively querying the most informative samples for annotation. 
This survey reviews the recent developments in deep active learning, focusing on the evaluation of informativeness, sampling strategies, integration with other label-efficient techniques, and the application of AL in medical image analysis. 
In this section, we will discuss the existing challenges faced by AL in medical image analysis and its future perspectives.

\subsection{Towards Active Learning with Better Uncertainty}

In AL, uncertainty plays a pivotal role.
However, it would be beneficial if the uncertainty more directly highlighted the model's mistakes. 
We can enhance the model's performance by querying samples with inaccurate predictions.

Recently, many works have adopted learnable performance estimation for quality control of deep model outputs. 
For instance, the recently proposed segment anything model (SAM) \citep{kirillov2023segment} provides IoU estimates for each mask to evaluate its quality.
In medical image analysis, automated quality control is critical to ensure the reliability and safety of the deep model outputs \citep{kohlberger2012evaluating}. 
For example, \cite{wang2020deep_qc} employed deep generative models for learnable quality control in cardiac MRI segmentation, where the predicted Dice scores showed a strong linear relationship with the real ones. 
Additionally, \cite{billot2023robust} used an additional neural network to predict the Dice coefficient of brain tissue segmentation results. 
Overall, learnable performance estimation can accurately predict the quality of model outputs. 
Hence, delving deeper into their potential for uncertainty-based AL is crucial to effectively tackle the issue of over-confidence.

Moreover, improving the probability calibration of model prediction is a promising way to mitigate the over-confidence issue.
Calibration \citep{guo2017calibration, mehrtash2020confidence} reflects the consistency between model prediction probabilities and the ground truth. 
A well-calibrated model should display a strong correlation between confidence and accuracy. 
For instance, if a perfect-calibrated polyp classifier gives an average confidence score of 0.9 on a dataset, it means that 90\% of those samples should indeed have polyps. 
In reality, deep models generally suffer from the issue of over-confidence, which essentially means that they are not well-calibrated. 
Currently, only a few uncertainty-based AL works have considered probability calibration. 
For instance, \cite{beluch2018power} found that the model ensemble has better calibration than MC Dropout.
\cite{xie2022dirichletbased} mitigated miscalibration by considering all possible prediction outcomes in the Dirichlet distribution. 
However, these methods are limited to proposing a better uncertainty metric and validating the calibration quality post-hoc. 
Existing calibration methods \citep{guo2017calibration, ding2021local} directly adjusted the distribution of prediction probabilities. 
However, these methods require an additional labeled dataset, thus limiting their practical applicability. 
Therefore, integrating probability calibration into uncertainty-based AL represents a valuable research direction worth exploring.

{
Among all the mentioned methods in §\ref{uncertainty}, adversarial-based uncertainty currently has limited applications in AL of medical image analysis. 
Since the adversarial samples tend to be close to the classification boundary, they can be regarded as uncertain samples, and selecting them for training can potentially improve the trained model's robustness. 
Exploring such ideas in medical image analysis, especially in the federated learning scenario, could be an interesting topic for future work.
}

\subsection{Towards Active Learning with Better Representativeness}

Representativeness-based AL effectively utilizes feature representations and data distributions for sample selection.
Cover-based and discrepancy-based AL methods implicitly capture the data distribution, whereas density-based AL explicitly estimates it. 
However, the latter requires supplementary strategies to ensure diversity.
{
For discrepancy-based AL, we can opt for a better metric of the distance between two probability distributions \citep{zhao2022comparing}. 
Besides, discrepancy-based AL has limited applications in medical image analysis currently. 
Finding a proper metric for medical images considering their special characteristics could be a promising direction for the future development of AL in the medical imaging domain.
}

As the core of density-based AL, density estimation in high-dimensional spaces has always been challenging. 
Popular density estimation methods, such as kernel density estimation and GMM, can encounter challenges when applied in high-dimensional spaces.
In future research, we can consider introducing density estimators tailored to high-dimensional spaces.
{Advanced tools like normalizing flow \citep{papamakarios2021normalizing} could be an appropriate choice in density estimation in high-dimensional spaces.}

\subsection{Towards Active Learning with Weak Annotation}

In §\ref{region}, we discuss region-based active learning, which only requires region-level annotation of a sample.
However, annotating all pixels within the region is still needed.
Several existing works have incorporated weak annotations with AL to simplify the task for annotators. 
In object detection tasks, \cite{vo2022active} trained deep models with image-level annotation. 
They selected samples with box-in-box prediction results and annotated them with bounding boxes.
Moreover, \cite{lyu2023boxlevel} adopted disagreement to choose which objects are worth annotating. 
Rather than annotating all objects within the image, they only required box-level annotations for a subset of objects. 
In AL of instance segmentation, \cite{tang2022active} only required annotations for each object's class label and bounding box, without the annotation of fine-grained segmentation masks. 
In future research, AL based on weak annotations is a direction worthy of in-depth exploration.

\subsection{Towards Active Learning with Better Generative Models}

In §\ref{generative}, we summarize the applications of generative models in AL. 
However, existing works have mainly focused on using GANs as sample generators.
Recently, diffusion models \citep{kazerouni2023diffusion} have advanced in achieving state-of-the-art generative quality. 
Furthermore, text-to-image diffusion models, represented by Stable Diffusion \citep{rombach2022high}, have revolutionized the image generation domain. 
Their high-quality, text-guided generation results enable a more flexible image generation. 
{With the use of ControlNet \cite{zhang2023adding}, the diffusion models could learn to follow a more detailed condition like a sketch or segmentation mask. }
Exploring the potential of diffusion models in deep AL is a promising avenue for future research.

\subsection{Towards Active Learning with Foundation Models}

With the rise of visual foundational models, such as contrastive language-image pre-training (CLIP) \citep{radford2021learning} and SAM \citep{kirillov2023segment}, and large language models (LLMs) like GPT-4 \citep{openai2023gpt4}, deep learning in medical image analysis and computer vision is undergoing a paradigm shift. 
These foundational models \citep{bommasani2021opportunities} offer new opportunities for the development of AL.

AL is closely related to the training paradigms in deep learning of computer vision and medical image analysis. 
From the initial approach of train-from-scratch to the ``pre-train-finetune" strategy using supervised or self-supervised pre-trained models, these paradigms usually require fine-tuning the entire network. 
Foundation models contain a wealth of knowledge.
When combined with recently emerging parameter-efficient fine tuning (PEFT) or prompt tuning techniques \citep{hu2021lora, jia2022visual}, we can tune only a minimal subset of model weights (e.g., 5\%) for rapid transfer to downstream tasks. 
As the number of fine-tuned parameters decreases, AL has the potential to further reduce the number of required annotated samples. 
{
\cite{bai2023slpt} integrated prompt tuning with AL in liver tumor segmentation.
A segmentation model trained on publicly available datasets was transferred to in-house datasets via a novel prompt updater.
With a mixed AL strategy of uncertainty and diversity, the proposed method reached the comparable performance of fully supervised tuning using around 5\% of samples and 6\% of tunable parameters.  
}
Therefore, it is essential to investigate the applicability of existing AL under PEFT or prompt tuning and explore the most suitable AL strategies for PEFT.

In natural language processing, LLMs have already taken a dominant role. 
Since most researchers cannot tune the LLMs, they rely on in-context learning, which provides LLMs with limited examples to transfer to downstream tasks. 
We believe that visual in-context learning will play a vital role in future research.
Therefore, selecting the most suitable prompts for visual in-context learning will become an important research direction of AL.

\section{Conclusion} 
\label{conclusion}
Active learning is important to deep learning in medical image analysis since it effectively reduces the annotation costs incurred by human experts.
This survey comprehensively reviews the core methods in deep active learning, its integration with different label-efficient techniques, and active learning works tailored to medical image analysis. 
We further discuss its current challenges and future perspectives. 
In summary, we believe that deep active learning and its application in medical image analysis hold important academic value and clinical potential, with ample room for further development. 

\section*{Acknowledgments}
This study was supported by the National Natural Science Foundation of China (Grant 82372097 and 82072021) and the Science and Technology Innovation Plan of Shanghai Science and Technology Commission (Grant 23S41900400).

\biboptions{authoryear}
\bibliography{refs}

\end{document}